\documentclass[compsoc, conference, a4paper, 10pt, times]{IEEEtran}

\usepackage{cite}
\usepackage{amsmath,amssymb,amsfonts, amsthm}
\usepackage{algorithmic}
\usepackage{graphicx}
\usepackage{textcomp}
\usepackage{xcolor}
\def\BibTeX{{\rm B\kern-.05em{\sc i\kern-.025em b}\kern-.08em
    T\kern-.1667em\lower.7ex\hbox{E}\kern-.125emX}}

\usepackage{todonotes}
\usepackage{tikz}

\newlength{\bibitemsep}
\newlength{\bibparskip}
\let\oldthebibliography\thebibliography
\renewcommand\thebibliography[1]{%
  \oldthebibliography{#1}%
  \setlength{\parskip}{\bibitemsep}%
  \setlength{\itemsep}{\bibparskip}%
}

\usepackage[normalem]{ulem}

\newcommand{\myparagraph}[1]{\vspace{1ex}\noindent{\bf #1}}
\makeatletter

\renewcommand*{\@fnsymbol}[1]{\ensuremath{\ifcase#1\or *\or \dagger\or \ddagger\or
   \mathsection\or \mathparagraph\or \|\or **\or \dagger\dagger
   \or \ddagger\ddagger \else\@ctrerr\fi}}
   
\DeclareRobustCommand\onedot{\futurelet\@let@token\@onedot}
\def\@onedot{\ifx\@let@token.\else.\null\fi\xspace}
\def\eg{\emph{e.g}\onedot} \def\Eg{\emph{E.g}\onedot}
\def\ie{\emph{i.e}\onedot}

\def\wrt{w.r.t\onedot} 
\def\etal{\emph{et al}\onedot}
\usepackage{pifont}
\makeatother
\usepackage{amsmath}
\usepackage{makecell}
\DeclareMathOperator*{\argmin}{arg\,min}
\DeclareMathOperator{\ReLU}{ReLU}
\usepackage{hhline}
\usepackage{etoolbox}
\expandafter\patchcmd\csname citeauthor \endcsname
  {\begingroup}{\begingroup\aftergroup\@}{}{}

\usepackage{amsmath,amssymb,amsfonts}
\usepackage{mathtools}
\usepackage{booktabs}
\usepackage[algoruled,nofillcomment,algo2e]{algorithm2e}
\usepackage{amsthm}
\usepackage{caption} %
\usepackage{subcaption} %
\usepackage{bbm} %

\usepackage{hyperref}

\usepackage{cleveref}
\usepackage{xspace} %
\usepackage{multirow}

\usepackage{cuted}

\usepackage{lastpage} %

\theoremstyle{definition}

\def\name{\textsf{trap weights}\xspace}
\def\namenoformat{{trap weights}\xspace}

\def\recall{\emph{extraction-recall}\xspace}
\def\Precision{\emph{Extraction-Precision}\xspace}
\def\precision{\emph{extraction-precision}\xspace}

\def\act{\emph{active neurons}\xspace}

\newcommand{\zero}{\emph{zero}}
\newcommand{\one}{\emph{one}}

\newcommand{\user}{user\xspace}
\newcommand{\users}{users\xspace}

\newcommand{\cp}{central party\xspace}

\newcommand{\attacker}{attacker\xspace}
\newcommand{\attackers}{attackers\xspace}
\newcommand{\Attacker}{Attacker\xspace}

\newcommand{\WAll}{\mathcal{W}} %
\newcommand{\f}{f_\WAll} %
\newcommand{\li}{l_i} %
\newcommand{\Grad}{G} %
\newcommand{\cla}{k} %
\newcommand{\Loss}{\mathcal{L}} %
\newcommand{\att}[1]{^{[#1]}} %
\newcommand{\Ltwo}{\ell_2} %
\newcommand{\totalusers}{\mathsf{N}}
\newcommand{\selectusers}{\mathsf{M}}
\newcommand{\posind}{\mathtt{P}}
\newcommand{\negind}{\mathtt{N}}

\begin{document}

\title{When the Curious Abandon Honesty:\\ Federated Learning Is Not Private$^\diamond$}

\author{
\IEEEauthorblockN{%
    Franziska Boenisch\IEEEauthorrefmark{1},
    Adam Dziedzic\IEEEauthorrefmark{1}\IEEEauthorrefmark{2}\textsuperscript{\textsection},
    Roei Schuster\IEEEauthorrefmark{1}\textsuperscript{\textsection},\\
    Ali Shahin Shamsabadi\IEEEauthorrefmark{1}\IEEEauthorrefmark{3}\textsuperscript{\textsection},
    Ilia Shumailov\IEEEauthorrefmark{1}\textsuperscript{\textsection}, and
    Nicolas Papernot\IEEEauthorrefmark{1}\IEEEauthorrefmark{2}
  }%
  \IEEEauthorblockA{\IEEEauthorrefmark{1}Vector Institute \IEEEauthorrefmark{2}University of Toronto \IEEEauthorrefmark{3}The Alan Turing Institute}%
}

\maketitle

\begingroup\renewcommand\thefootnote{\textsection}
\footnotetext{Equal contribution. \newline \indent $^\diamond$. Accepted at the 8th IEEE European Symposium on Security and Privacy (IEEE Euro S\&P).}
\endgroup

\begin{abstract}
In federated learning (FL), data does not leave personal devices when they are jointly training a machine learning model. 
Instead, these devices share gradients, parameters, or other model updates, with a central party (\eg,~a company) coordinating the training. 
Because data never ``leaves'' personal devices, FL is often presented as privacy-preserving. 
Yet, recently it was shown that this protection is but a thin facade, as even a passive, honest-but-curious \attacker 
observing gradients can reconstruct data of individual \users contributing to the protocol. 

In this work, we show a novel data reconstruction attack which allows an active and dishonest \cp to efficiently extract user data from the received gradients.
While prior work on data reconstruction in FL relies on solving computationally expensive optimization problems or on making easily detectable
modifications to the shared model's architecture or parameters, in our attack 
the \cp makes inconspicuous changes to the shared model's weights before sending them out to the \users. 
We call the modified weights of our attack \emph{\namenoformat}.

Our active \attacker is able to recover \user data \textit{perfectly}, \ie, with zero error, even when this data stems from the same class. 
Recovery comes with near-zero costs: the attack requires no complex optimization objectives. 
Instead, our \attacker exploits inherent data leakage from model gradients and simply amplifies this effect by maliciously altering the weights of the shared model through the \namenoformat.
These specificities enable our attack to scale to fully-connected and convolutional deep neural networks trained with large mini-batches of data.
For example, for the high-dimensional vision dataset ImageNet, we perfectly reconstruct more than 50\% of the training data points from mini-batches as large as 100 data points.
In textual tasks, such as IMDB sentiment analysis, more than 65\% of data points from mini-batches containing 100 data points can be perfectly reconstructed.

\end{abstract}

\section{Introduction}
\label{sec:introduction}

With machine learning (ML) being increasingly applied to sensitive data in critical use-cases such as health care~\cite{Pryss.2015Mobile, Hakak.2020Framework},  smart metering~\cite{Wang.2021Electricity, Gholizadeh.2021Distributed}, or the internet of things~\cite{Khan.2021Federated, Nguyen.2021Federated}, there is a growing need for privacy-preserving training schemes that do not leak sensitive information. 
Federated learning (FL) is a widely popular distributed learning protocol~\cite{McMahan.2017Communication} where \user data can be utilized for jointly training an ML model without the data ever leaving the \users' device.
Instead, the device  computes and sends model updates to a \cp which aggregates them to produce a shared model. 
\emph{Assuming} the model updates do not reveal the \user data, FL would, thereby, preserve a notion of privacy.

This assumption has been repeatedly contested by prior work. It has been shown how the model updates sent to the \cp not only leak training data  membership~\cite{Melis.2019Exploiting} (\ie~allow the \attacker to tell if a given data point was used in training) but also properties of the training data~\cite{Ganju.2018Property,Melis.2019Exploiting}. Inspecting model updates allows \attackers to even (partially) \emph{reconstruct}~\cite{Geiping.2020Inverting,Wang.2019Beyond,Yin.2021See,Zhu.2020Deep,Zhao.2020iDLG,fowl2021robbing} \users' training data. 
Ultimately, FL in its naive implementation offers little to no \emph{guarantees} regarding potential leakage of \user data to other \users or to the \cp.

\begin{figure}[t]
\begin{flushright}

\includegraphics[width=0.46\textwidth]{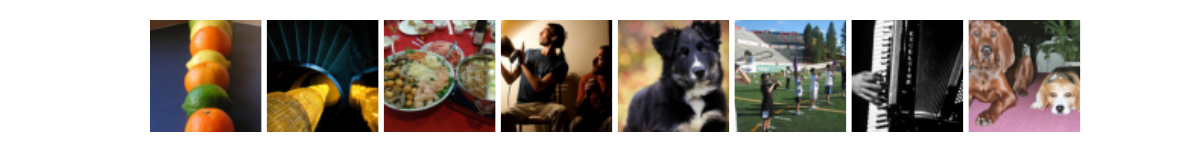}
\includegraphics[width=0.46\textwidth]{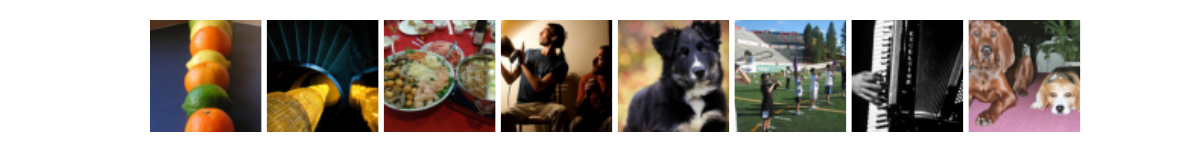}
\includegraphics[width=0.46\textwidth]{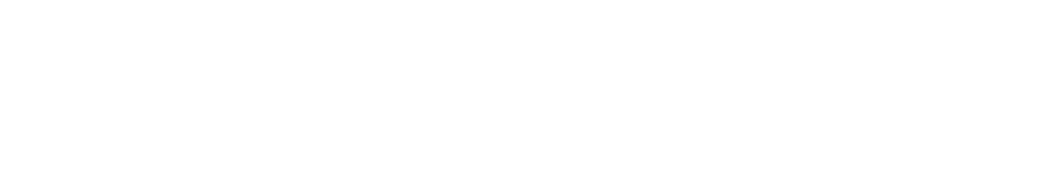}

\begin{tikzpicture}[remember picture, overlay]

\newcommand\xvalue{-7.8}
\node at (\xvalue, 3.25) {\footnotesize Original};
\node at (\xvalue, 2.25) {\footnotesize Extracted};
\node at (\xvalue, 1.4) {\footnotesize Original};
\node at (\xvalue, 0.4) {\footnotesize Extracted};

\node[text width=6.5cm,align=left,anchor=north, execute at begin node=\setlength{\baselineskip}{2ex}] at (-3.8, 1.8) {\normalsize ive read a few of the reviews and im kinda sad that a lot of the story seems [UNK] ...};
\node[text width=6.5cm,align=left,anchor=north, execute at begin node=\setlength{\baselineskip}{2ex}] at (-3.8, 0.8) {\normalsize ive read a few of the reviews and im kinda sad that a lot of the story seems [UNK] ...};

\end{tikzpicture}
\end{flushright}
\caption{\textbf{Original and Reconstructed Data.} Original data and data points extracted from model gradients with our \namenoformat attack. Extraction is \textbf{perfect} \ie~reconstruction error is zero.}
\label{fig:teaser_reconstruction}
\end{figure}
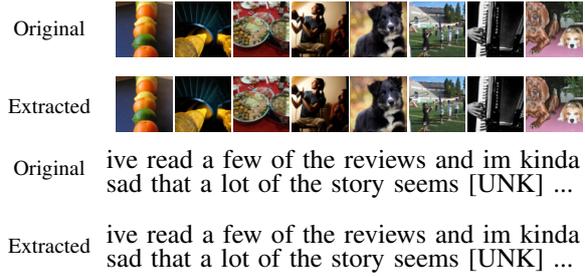

Yet, existing data reconstruction attacks either are computationally expensive and yield low-fidelity extraction~\cite{Zhu.2020Deep,Zhao.2020iDLG}, are limited to small mini-batch sizes~\cite{Geiping.2020Inverting}, or require modifications of the model architecture that are trivially detected~\cite{fowl2021robbing}.
Another other line of concurrent work proposes modifications to the model parameters that are still easily noticeable:
The attack introduced by Pasquini~\etal~\cite{pasquini2021eluding} sets a noticeable portion of parameters to zero or negative values. Similarly, the attacks by Wen~\etal~\cite{wen2022fishing} zero out many parameters of the last fully connected classification layer. 
In this work, we perform data extraction from large mini-batches of local data based on inconspicuous manipulations of the shared model weights. %
We start by showing scenarios where the gradients sent to the \cp include full, memorized training data points.
We then proceed to show that a malicious \cp can significantly amplify this leakage by simply adversarially setting the model's weights with our \emph{\namenoformat} method, prior to dispatching the weights to \users.

Our \namenoformat mainly rely on re-scaling components in the model's weights matrix and can be applied to unmodified model architectures, which makes the attack more stealthy.
By adversarially initializing the shared model with our \namenoformat, the \cp can ensure they are able to \emph{perfectly} extract a significant portion of the \users' training data, as depicted in \Cref{fig:teaser_reconstruction}. 
This even holds when the gradients are computed over large training data mini-batches containing only data from the same class, a scenario in which previous \textit{optimization-based attacks} usually fail to obtain high-fidelity reconstructions~\cite{wainakh2021federated}. 
Since in FL, the \cp holds full control over the shared model weights that are sent out to \users, our attack integrates naturally in the FL protocol.
Furthermore, our attack is highly computationally efficient since it extracts individual training inputs by simply projecting the appropriate portions of the \users' gradients onto the input domain.
Finally, we show both in theory and in practice that our attack is equally successful when \users perform multiple rounds of local training (Fed-Avg~\cite{McMahan.2017Communication}) and send the model updates instead of the gradients to the \cp.

In summary, we make the following contributions:
\begin{itemize}
    \item We observe that in neural networks starting with a fully-connected layer, even gradients of large training data mini-batches contain \emph{individual} training data points. 
    In other words, in FL, mini-batch training data points are often directly sent from \users to the \cp, such that even an honest-but-curious central party has access to them.
    \item We show that a dishonest and active \cp can amplify the leakage of individual training data points and extend it to other model architectures by adversarially initializing the weight of the shared model.
    \item In this setting, we perform data reconstruction on image and text data. Our attack is able to perform an extremely computationally efficient extraction of individual training data points in only a single-step computation over the received model updates with the attack setup depicted in \Cref{fig:privacy_attack_points}.
    For complex image datasets such as ImageNet~\cite{deng2009imagenet}, the attack yields perfect reconstruction of more than $50$\% of the training data points, even for large training data mini-batches that contain as many as $100$ data points. For textual tasks such as IMDB sentiment analysis~\cite{imdb}, it perfectly extracts more than 65\% of the data points for mini-batches with $100$ data points.
\end{itemize}

\newcommand{\oa}{\includegraphics[scale=0.4]{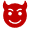}}%

\begin{figure}[t]
\centering

\includegraphics[width=0.25\textwidth]{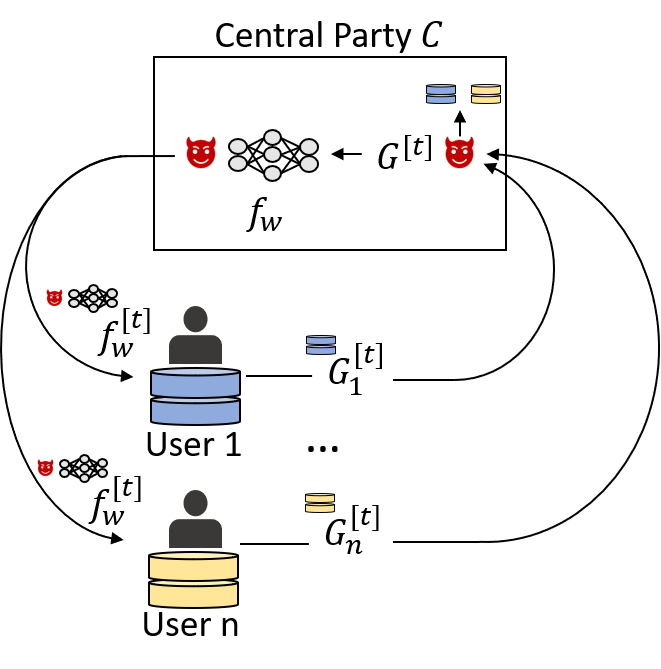}

\caption{\textbf{Course of our Attack.} Our attack (\oa) targets two points in the FL protocol: At iteration $t$, the \cp actively manipulates the weights $\WAll$ of the shared model $\f$ before the model is sent out to the \users. This causes the gradients $\Grad_i\att{t}$ of \user~$i$ to contain individual training data points which the \cp can then extract before calculating the averaged gradients $\Grad\att{t}$ and applying them to $\f$.}
\label{fig:privacy_attack_points}
\end{figure}

\setcounter{footnote}{0}

\section{Background: Neural Networks, Federated Learning, and Differential Privacy}
\label{sec:background}

\myparagraph{Neural Networks}. Let $\f:\mathbb{R}^m\rightarrow \{1,\cdots,\cla\}$ be a $\cla$-class classifier defined as a set of $l$ layers parameterized by trainable \emph{weights} $\WAll$.
Each  layer consists of a linear operation paired with a non-linear activation function (e.g. ReLU). In this work, we consider two popular layer types: fully-connected and convolutional layers.

The goal of the model $\f$ is to map an input $x_i \in X$ to its desired ground-truth $y_i \in Y$. Therefore, the model weights $\WAll$ are adapted in a training process, most commonly with the mini-batch \emph{Stochastic Gradient Descent} (SGD). 
To adjust the initial $\WAll$, mini-batch SGD  repeats the following sequence of steps: (1) sample a mini-batch of size $B$  from the training data $\{(X, Y)_b\}_{b=1}^B$, (2) take a forward pass through the model to obtain its predictions on the mini-batch, (3) compute the difference between predictions and ground-truth labels, called the \emph{loss} $\Loss$, (4) compute the gradient of $\Loss$ w.r.t. the weights, called the \emph{weight gradient} $\Grad$, and update the weights accordingly. 

To bootstrap mini-batch SGD, weights need to be initialized by sampling from a random distribution; popular distributions  include the zero-mean Gaussian~\cite{Giryes.2016Deep},  \emph{Xavier}~\cite{Glorot.2011Deep} or \emph{He}~\cite{He.2015Delving} distributions. The choice of distribution has a large effect on learning success~\cite{Xie.2017All}. In fact, when weights are maliciously initialized, the final model's utility might be degraded~\cite{Grosse.2019Adversarial}.

\myparagraph{Federated Learning}. FL~\cite{McMahan.2017Communication} is a communication protocol for training a shared ML model $\f(\cdot)$ on decentralized data {${\{(X_i,Y_i)\}}_{i=1}^\totalusers$} owned by $\totalusers$ different \users $\{u_i\}_{i=1}^\totalusers$. Since collecting and managing all the data centrally might be costly, time consuming, and stand in conflict with the confidentiality of these respective \users' data, FL enables each \user to keep their data locally.
A \cp coordinates the training of the shared model by iteratively aggregating gradients computed locally by users.

More formally, let $t \in \{1, \cdots T\}$ be the current iteration of the FL protocol.
At iteration $t=0$, the model $f(\cdot)$ is initialized (at random) by the \cp denoted as~$C$.
Let  $\f\att{t}(\cdot)$ be the model with its weights $\WAll\att{t}$ at iteration~$t$.
At every iteration $t$, $\selectusers$ out of the $\totalusers$ ($\selectusers \ll \totalusers$) \users are selected to contribute to the learning.
Then, each of the selected $\selectusers$ \users $u_i$ obtains $\f\att{t}(\cdot)$ from~$C$ and calculates the gradients $\Grad_i\att{t}$ for $\f\att{t}(\cdot)$ based on one mini-batch $b$ sampled from their local dataset ${(X_i,Y_i)}_b$.
In other words, the \user computes the gradient $\Grad_i\att{t} = \nabla_{\WAll}\Loss({(X_i,Y_i)_b})$.
Each $u_i$ uploads their gradients to $C$,
who then averages all of these gradients to update the shared model's parameters:
\begin{equation}
   \Grad\att{t}=\frac{1}{\selectusers}\sum_{i=1}^{\selectusers} \Grad\att{t}_i, \quad \WAll\att{t+1}=\WAll\att{t}-\eta G\att{t}.
\end{equation}
FL, thereby, represents a decentralization of the mini-batch SGD (i.e. distributed training from mini-batches of \user data).

\section{Existing Data Reconstruction Attacks}
\label{sec:RelatedWork}
This section introduces prior work on passive and active data reconstruction attacks in FL and discusses the limitations of attacks based on iterative optimization.

\subsection{Passive Attackers}
\label{sub:prior_work}
Passive attackers performing data reconstruction attacks in FL can simply observe the received gradients but not maliciously manipulate the protocol. 
Phong~\etal~\cite{Phong.2017Privacy} were the first to show how gradients leak information that can be used to recover training data from single neurons or linear layers. Recent work~\cite{Hitaj.2017Deep,Phong.2017Privacy,Wang.2019Beyond,Zhu.2020Deep,Zhao.2020iDLG,Geiping.2020Inverting,Yin.2021See,sun2020provable} proposed that the \cp or \users involved in FL training launch data reconstruction attacks based on either training a Generative Adversarial Network \cite{goodfellow2014generative} (GAN) or solving a second order optimization problem.

\myparagraph{Optimization-based Instance Reconstruction Attacks}. Several attacks aim to reconstruct individual \user data points while also relaxing the assumption that data labels are available to the \attacker. Zhu~\etal~\cite{Zhu.2020Deep} proposed Deep Leakage from Gradients (DLG), where a data reconstruction attack is formulated as a joint optimization problem on the labels and input data, see \Cref{algo:SOA_opt_based_data_reconstruction} in Appendix~\ref{app:passive_reconstruction}.
iDLG~\cite{Zhao.2020iDLG} sped up the convergence rate of DLG~\cite{Zhu.2020Deep} by analytically computing the labels based on the \users' gradients of the last layer.
These works, and other optimization-based ones~\cite{Geiping.2020Inverting}, are limited to a setting where mini-batches only contain a single example, i.e., $B=1$.
GradInversion~\cite{Yin.2021See} regularizes DLG's objective to improve the extraction fidelity, attaining some success in extraction for mini-batches of size $B>1$.
In Section~\ref{sub:comparison_previous_work} we compare performance of our approach against a state-of-the-art optimization-based attack. 
Our attack is superior in extracting individual training data even for large mini-batch sizes of $B\geq100$, and being far more computationally efficient (even for passive adversaries in the honest-but-curious model).
We present a more thorough overview on passive data reconstruction attacks in Appendix~\ref{app:passive_reconstruction}.

\myparagraph{Limitations of Optimization-Based Attacks.}
We hereby provide a brief exposition to Zhu~\etal~\cite{Zhu.2020Deep}'s DLG, as a representative case study of an optimization-based attack.
Their approach, characteristic of optimization-based data reconstruction attacks, is given in Algorithm~\ref{algo:SOA_opt_based_data_reconstruction} in Appendix~\ref{app:passive_reconstruction}. It firstly randomly initializes a ``dummy data point and corresponding label''  $(\hat{\mathbf{x}},\hat{y})$ and computes the resulting ``dummy gradients'' as $\hat{G}={\nabla_{\WAll^t}\mathcal{L}(f_{\WAll^t}(\hat{\mathbf{x}}),\hat{y})}$. Then, they iteratively optimize the dummy data to produce gradients that are close to the original gradients $G^t_i$ by solving: 
\begin{align}
    \label{eq:DGL}
    \mathbf{x}^*_i =\argmin_{\hat{\mathbf{x}}} \| \Grad_i\att{t} - \hat{G} \|^2 \\
    \label{eq:DGL2}
    \quad y^*_i =\argmin_{\hat{y}} \| \Grad_i\att{t} - \hat{G} \|^2.
\end{align}
\setlength{\textfloatsep}{0.5cm}

DLG often fails to reconstruct high-fidelity data points and discover the ground-truth labels consistently because of a lack of convergence in the optimization. 
While other methods offer improvements (\eg iDLG~\cite{Zhao.2020iDLG} sped up the convergence by simplifying the objectives in Equations~\ref{eq:DGL} and~\ref{eq:DGL2} from both data and label reconstruction to only data reconstruction; and GradInversion~\cite{Yin.2021See} adds useful regularization), they suffer from the same pathology.

We identify several reasons for this. First, the gradient of the loss is non-injective \ie~is not invertible everywhere: different mini-batches may yield nearly identical gradients~\cite{Shumailov.2021Manipulating}. This holds whether the \user samples mini-batches that contain multiple data points or a single data point only, \ie $B=1$. Second, optimization-based attacks converge to different minima due to the underlying randomness (see step 1 in \Cref{algo:SOA_opt_based_data_reconstruction} in the Appendix). These minima correspond to different possible reconstructions of the input that often differ from the original training points~\cite{Yin.2021See}. Third, optimization-based attacks are computationally expensive: they either need to train a GAN or solve a second-order gradient optimization problem. Instead, our attack extracts \textit{exact} data points from the gradients without any optimization or GAN training.

\subsection{Active Attackers}
In the work most similar to ours, \cite{fowl2021robbing} considers a threat model with an active and dishonest central party, similar to our setup. 
This attack relies on the existence of a fully-connected layer early within the network (otherwise, the attack adds it). Since this layer's weights have to contain many weight rows with the exact same weight values, this layer is inherently detectable.\footnote{This is inherent to the attack because the method relies on each row computing the exact same function on the data and binning its result by varying only the bias term, such that it becomes likely that a bin contains only one input. Conversely, our \namenoformat are initialized such that it is likely that an output neuron is only activated for a single input in a mini-batch while avoiding imposing a highly regular structure on the weight matrix.}
Additionally, they do not discuss passive analytical-extraction attacks.
Finally, our work generalizes their setup and performs successful extraction also for textual data.

In follow-up work, \cite{wen2022fishing} proposes an attack that requires modifications to the model parameters (specifically, to the last fully-connected classification layer) sent to a \user but without changing the model architecture. The attack extracts single data points by increasing the gradient contribution of a target data point and decreasing the gradient contribution of other data points. The final goal of an attacker is to reduce an aggregated gradient to an update calculated on a single sample. The attack is easily detectable since it requires many parameters in the last layer to be zeroed out. Moreover, our attack extracts individual data points in a single training round while their approach requires a collection of many updates from an individual user.
In the cross-device FL setting where participants get randomly sampled from millions of users, it is possible that single users participate fewer times than required by the attack.

In another active attack proposed by~\cite{pasquini2021eluding}, a server sends distinct malicious parameters to individual users. 
The main purpose of the attack is to circumvent the protection of Secure Aggregation (SA) in FL and enable the \cp to learn individual model updates from a target user.
However, the work does not propose individual \user-data point extraction, as enabled by out \namenoformat.
We argue that by including our \namenoformat into their attack and sending our \namenoformat to the target \user, they could efficiently extract this target \user's private data.

\section{Threat Model and Assumptions}
This section presents our threat model in terms of the assumed attacker, the FL deployment, and the assumptions required for our attack to succeed.

\subsection{The Attacker}
\label{sec:attacker_model}
Our \attacker aims at extracting individual training data points from a chosen subset of the participating \users. 
Therefore, the attacker's primary vantage point is the \cp who is in charge of orchestrating the FL protocol. The assumption here is that, for example, the company orchestrating the FL protocol or potential rogue employees, are untrusted. This is the same attacker that FL is meant to defend against by leaving data on the users' devices.
For brevity, in the following, we will refer to the \cp as the \attacker, even though the \attacker can be a third party controlling the \cp to deploy our \namenoformat-attack.

In FL, the \cp initiates the FL protocol and chooses the task to train the shared ML model for.
Therefore, the \cp is aware of the type, domain, and dimensionality of data held by the users.
It instantiates the shared model appropriately to learn from this data.
Furthermore, in the standard FL scenario considered in this work, the \cp holds full control over the shared model weights and can read \users' gradient updates that are sent back.
Finally, our \cp is in charge of sampling the \users who contribute their gradients in a given round---following standard deployments of the protocol~\cite{bonawitz2019towards}.
This allows the \cp to even run targeted attacks against specific \users. 

\subsection{Assumptions and FL Setup}
\label{sub:FL_deployment}
Following prior work~\cite{bonawitz2019towards}, we consider an FL protocol where users calculate the model gradients locally on one (potentially large) mini-batch of their training data and share the resulting gradients directly with the \cp. 
We assume that the data features are scaled in the range $[0,1]$, which is a standard pre-processing step in ML.
When users have abundant amounts of data, they can perform local gradient calculation and averaging over more than one mini-batch, see evaluation in \Cref{sub:active_attack_evaluation}.

We, furthermore, assume that the \attacker is in possession of a small amount (\eg, one mini-batch) of data from the \users' private data domain.
This is no strong assumption given that the \cp chooses the ML task and has to instantiate the ML model appropriately.

The weight-manipulation attacks we study in this paper are not agnostic of the model architecture. 
In designing the attack, we focus on victim models that contain a ReLU-based fully-connected layer, and we experiment with several different types of such networks.
This is not a material limitation: the approach of manipulating shared model weights to promote leakage is very flexible, and can be extended to cover many more architectures as needed using similar techniques.
For example in \Cref{sub:adv-forwarding-cnns} and Appendix~\ref{sec:adv_forwarding}, we show how to extend the attack to work on networks that contain convolutional layers and in \Cref{sec:adv-initialize-evaluation}, also experimentally evaluate extraction under the presence of a token-embedding layer.

\subsection{Course of Attack}
\label{sub:attack_course}

The course of our attack is illustrated in Figure~\ref{fig:privacy_attack_points}.
In a given round of the protocol, the \cp maliciously manipulates the shared model with our \namenoformat.
Note that the \cp does not necessarily attack all \users in every round of the protocol.
Instead, it can target one or several specific \users in one or more chosen round(s).
To target a subset of the $\selectusers$ \users at iteration $t$, the \cp can send out different models to \users under attack and other \users~\cite{pasquini2021eluding}:
while the targeted \users receive a model initialized with our \namenoformat, all other \users receive the shared model used to train the ML task.
Attacking only a few \users in a few rounds makes the attack more stealthy and allows the \cp to train a performant shared model based on the gradient updates received in the benign rounds or from non-targeted \users.

After receiving the gradients from \users under attack, the \cp simply projects the appropriate portions these gradients onto the input domain.
In the following section, we will show how this approach can yield perfect extraction of the \users' data points.

\section{Passive Analytical Extraction for FC-NNs}
\label{sec:data_leakage}

Here, we show how the gradients of an FC-NN directly leak the individual training data points they are computed on, even to a passive \attacker who just observes said gradients. 
In \Cref{sub:firstlayer}, we formally show that for a single training data point, \ie a mini-batch size of $B=1$, perfect extraction from the network gradients is possible.
Then, in \Cref{sub:multiple_sample_inversion}, we motivate why it is also possible to perfectly extract a small number of individual data points from gradients, even when working with larger mini-batches of size $B>1$.
However, the success of this passive extraction attack drops as the mini-batch sizes increase. 
This limitation motivates our active adversarial weight initialization attack, which we introduce in \Cref{sec:adv-init-first-fc}.

\subsection{Single-Input Gradients Directly Leak Input}
\label{sub:firstlayer}
It has been shown by Geiping~\etal~\cite{Geiping.2020Inverting} that a single input data point $\mathbf{x}$ can be reconstructed from the gradients of any fully-connected layer which is preceded only by fully-connected layers and contains a bias $\mathbf{b}$.
This holds if the gradient of the loss \wrt the layer's output $\mathbf{y}=\ReLU(W\mathbf{x} +\mathbf{b})=\max(0, W\mathbf{x} +\mathbf{b})$ contains at least one non-zero entry.
For detailed proof of the above see Proposition~D.1 in~\cite{Geiping.2020Inverting}. 
In particular, when considering the first model layer, reconstructing its input data \emph{directly} corresponds to obtaining the original input data point $\mathbf{x}$.
Let $y_i$ denote the output of the $i^{th}$ neuron of the first and fully-connected layer of a model, and let $\mathbf{w}^T_i$ be the corresponding row in the weight matrix and $b_i$ the corresponding component in the bias vector.
Assume $\mathbf{w}^T_i \mathbf{x} + b_i > 0$, and therefore, $\ReLU(\mathbf{w}^T_i \mathbf{x} + b_i) = \mathbf{w}^T_i \mathbf{x} + b_i$.
The reconstruction of the input $\mathbf{x}$ is done by calculating the gradients of the loss \wrt the bias and the weights as follows: 

\begin{equation}
    \frac{\partial \mathcal{L}}{\partial b_i} = \frac{\partial \mathcal{L}}{\partial y_i} \frac{\partial y_i}{\partial b_i} = \frac{\partial \mathcal{L}}{\partial y_i}
\end{equation}
since $\frac{\partial y_i}{\partial b_i} = 1$, where $y_i = \mathbf{w}^T_i \mathbf{x} + b_i$. 

\begin{equation}
\label{eg:scaled_gradients_output}
    \frac{\partial \mathcal{L}}{\partial \mathbf{w}^T_i} = \frac{\partial \mathcal{L}}{\partial y_i} \frac{\partial y_i}{\partial \mathbf{w}^T_i} = \frac{\partial \mathcal{L}}{\partial b_i} \mathbf{x}^T
\end{equation}

Thus, if any $\frac{\partial \mathcal{L}}{\partial b_i} \ne 0$, perfect reconstruction is given by:

\begin{equation}
\label{eq:rescaling}
    \mathbf{x}^T = (\frac{\partial \mathcal{L}}{\partial b_i})^{-1} \frac{\partial \mathcal{L}}{\partial \mathbf{w}^T_i}
\end{equation}

According to~\Cref{eg:scaled_gradients_output}, the gradient of the loss \wrt the weights directly contains a scaled version of the input data. 
The exact scaling factor is $(\frac{\partial \mathcal{L}}{\partial b_i})$, which is the gradient of the loss \wrt the bias. 
This gradient is computed in the regular backward pass together with the gradient of the weights.
Therefore, obtaining the scaling factor by just reading it from the gradients of the bias and
inverting it to $(\frac{\partial \mathcal{L}}{\partial b_i})^{-1}$ comes at zero costs and the factor can be directly applied to rescale the gradient of the weights and obtain the input data point $\mathbf{x}$, see \Cref{eq:rescaling}.
Intuitively, the reason why there is a rescaled version of the input data in the gradients and why this would be beneficial for learning can be motivated by revisiting the simple perceptron algorithm~\cite{gallant1990perceptron}.
When an input is misclassified, the weight update in the perceptron algorithm consists simply in adding this input to the weights, which makes the algorithm learn.

\subsection{Mini-batch Gradients Directly Leak Some Individual Inputs}
\label{sub:multiple_sample_inversion}

It turns out that individual data point leakage is not limited to gradients computed over a mini-batch of size $B=1$: we observe that gradients computed over larger mini-batches also sometimes leak individual training points. 
To forge an intuition for this phenomenon, \Cref{fig:100-gradients-cifar10} in Appendix~\ref{app:additional_results} visualizes the gradients of the first fully-connected layer's weight matrix of the FC-NN described in~\Cref{tab:architectures}. 
We see that we are able to clearly distinguish some of the training data points within the rescaled gradients. This is despite the fact that these gradients were computed over a mini-batch of $B=100$ inputs sampled from the CIFAR10 dataset.

\vspace{0.1cm}
\textbf{Why do some gradients contain individual training data points?}
We denote a training data mini-batch by $X=\{x_1, x_2, \cdots, x_B\} \in \mathbb{R}^{(m \times B)}$ with~$B>1$.
The gradient of this mini-batch $X$ is equal to the average of all gradients computed for each of the data points  $\{x_1, x_2, \cdots, x_B\} $ that make up the mini-batch.
Let  $y_i$ denote again the output of the $i^{th}$ neuron of the fully-connected layer, and let $\mathbf{w_i}$ and $b_i$ be the corresponding row in the weight matrix and the component in the bias vector, respectively.
Then the gradient $\mathbf{G}_{{\mathbf{w}}^T_i}$ and $G_{b_i}$ of $\mathbf{w_i}$ and $b_i$ can be computed as follows: 

\begin{equation}
\label{eq:batch_gradients}
\begin{aligned}
    \mathbf{G}_{{\mathbf{w}}^T_i} &= \frac{1}{B} \sum_{j=1}^B \frac{\partial \mathcal{L}}{\partial y_{(i,j)}} \frac{\partial y_{(i,j)}}{\partial \mathbf{w}^T_i}\\
    G_{b_i} &= \frac{1}{B} \sum_{j=1}^B \frac{\partial \mathcal{L}}{\partial y_{(i,j)}} \frac{\partial y_{(i,j)}}{\partial b_i}
\end{aligned}
\end{equation}
with $y_{(i,j)} = \ReLU(\mathbf{w}_i^T\mathbf{x}_j+b_i)$.
These equations illustrate that the gradient $\mathbf{G}_{{\mathbf{w}^T_i}}$ over the data mini-batch $X$ contains a weighted overlay of all the input data points $\mathbf{x}_j$ from the mini-batch.
The weighting, therein, depends on the contribution of each data point to the model loss $\mathcal{L}$.

We observe that, in some cases, all but one training data point $\mathbf{x^*}$ from the data mini-batch have zero gradients. This is due to the $\max$ operation in $\ReLU(\mathbf{w}_i^T\mathbf{x}+b_i$)$:= \max(\mathbf{w}_i^T\mathbf{x}+b_i,0)$. When $\mathbf{w}_i^T\mathbf{x}+b_i$ is negative, the $\ReLU$ outputs zero, which results in zero gradients for the corresponding data point. 
When the gradients are zero for all data points but for the one data point $\mathbf{x^*}$,
the weight gradient $\mathbf{G}_{{\mathbf{w}^T_i}}$ from \Cref{eq:batch_gradients} becomes $\mathbf{G}_{{\mathbf{w}^T_i}} = \frac{1}{B}\frac{\partial \mathcal{L}}{\partial y_{(i,*)}} \frac{\partial y_{(i,*)}}{\partial \mathbf{w^T_i}} $ with $y_{(i,*)} = \ReLU(\mathbf{w}_i^T\mathbf{x^*}+b_i)$. This reduces the data extraction from the case of $B>1$ to the case of $B=1$, for which we saw in \Cref{sub:firstlayer} that the data point $\mathbf{x}^*$ can be perfectly extracted. In other words, $\mathbf{w}_i^T\mathbf{x}+b_i$ being negative for all data points but one results in accidental leakage of that data point---enabling its exact reconstruction by a passive adversary.

\subsection{Individual Inputs still Leak from Mini-batch Gradients computed in FedAvg}
\label{sub:fedavg_theory}
FedAvg is another popular protocol for FL~\cite{McMahan.2017Communication} where \users do not send their gradients after each local iteration of training. 
Instead, they calculate $T'$ many local epochs over $l$ mini-batches of their data.
After each iteration $t'$ of the total $T'\cdot l$ many local iteration, they update the model according to the respective gradients of the weights and biases, and a learning rate $\eta$ as $\f^{t' + 1}(\cdot) = \f^{t'}(\cdot) - \eta (\frac{\partial \mathcal{L}}{\partial \mathbf{w}^{t'}}, \frac{\partial \mathcal{L}}{\partial b^{t'}})$.
Once the local training is completed, the \users send the updated shared model $\f^{T'}(\cdot)$ to the \cp.
By calculating the difference between the shared model $\f^{0}(\cdot)$ sent to the user and the obtained model, and by re-scaling according to $\eta$, the \cp obtains the value of the \user's local model update as $\sum_{t'=1}^{T'}\frac{\partial \mathcal{L}}{\partial \mathbf{w}^{t'}}, \sum_{t'=1}^{T'} \frac{\partial \mathcal{L}}{\partial b^{t'}}=\frac{\f^{0}(\cdot)-\f^{T'}(\cdot)}{\eta}$.
According to \Cref{eg:scaled_gradients_output}, after every local iteration $t'$ the gradient of the local weights $\frac{\partial \mathcal{L}}{\partial \mathbf{w}^{t'}}= \mathbf{x}\frac{\partial \mathcal{L}}{\partial b^{t'}}$.
Therefore, 
$\sum_{t'=1}^{T'}\mathbf{w}^{t'} = \sum_{t'=1}^{T'} \mathbf{x}b^{t'} = \mathbf{x} \sum_{t'=1}^{T'} b^{t'} $.
Since the server knows $\sum_{t'=1}^{T'} b^{t'}$ from the model update, it can multiply $\left(\sum_{t'=1}^{T'} b^{t'}\right)^{-1} \cdot \sum_{t'=1}^{T'}\mathbf{w}^{t'}=\mathbf{x}$ and extract the user data perfectly.
We experimentally validate this theoretical insight on large mini-batches at the end of \Cref{sub:active_attack_evaluation}.

\section{Active Adversarial Initialization of the First Fully-Connected Layer}
\label{sec:adv-init-first-fc}

\Cref{sec:data_leakage} illustrates under which conditions model gradients leak data points to a passive \attacker capable of observing these gradients. 
In the following, we show how an active \attacker can amplify  previously-accidental leakage during the passive attack by controlling the weights $\mathbf{w}_i$ and biases $b_i$. 
For example, while a passive attacker can extract roughly $20$\% of arbitrary data points from a batch size $B=100$ for 1000 neurons (\ie weight rows in the fully-connected layer) on ImageNet, the active attack can more than double the number of extracted data points to $45$\%. Next, we show how to make such malicious choices to extract a larger number of individual training data points from model gradients.

\subsection{Intuition of our Trap Weights}
\label{sub:adv-initialize-methods}

Without loss of generality, we will suppress the bias term in the following considerations.
The multiplication of a single weight row $\mathbf{w}_i$ corresponding to the $i^{th}$ neuron at the fully-connected layer with some input data point $\mathbf{x}$ can be expressed as a weighted sum of all of the features in $\mathbf{x}$ as follows
\begin{equation}
    \label{eq:weighted-input}
    \mathbf{y}_i = \mathbf{w}_i^T\mathbf{x} = \sum_{j=1}^m w_i^{(j)} x_j\text{.}
\end{equation}
In weight row $\mathbf{w}_i$, let $\negind$
and $\posind$ denote the sets of indices that hold the negative and positive weight components, respectively.
Given $\ReLU$ activation, the $i^{th}$ neuron is only activated on $\mathbf{x}$ if the sum of the features weighted by the negative components is smaller than the sum of the features weighted by the positive components:
\begin{equation}
    \label{eq:weight_rules}
    \sum_{n \in \negind} w_i^{(n)} x_n < \sum_{p \in \posind} w_i^{(p)} x_p\text{.}
\end{equation}
Therefore, $\mathbf{x}$ will yield non-zero gradients at the $i^{th}$ neuron if and only \Cref{eq:weight_rules}, holds for its features. 

When the inequality holds only for a single data point in a mini-batch, this data point can be individually extracted from the gradients, as described in Section~\ref{sub:multiple_sample_inversion}.
The idea behind our \namenoformat is to set the components within each weight row corresponding to the neurons of the first fully-connected layer, such that  \Cref{eq:weight_rules} only holds relatively rarely in inputs, and is therefore likely to only hold for a single data point within a mini-batch.

\subsection{Adversarial Weight Initialization}
\label{sub:adv-initialize-symmetric}

\setlength{\textfloatsep}{0.5cm}

\begin{algorithm2e}[t]
\DontPrintSemicolon
\SetKwComment{Comment}{{\scriptsize$\triangleright$\ }}{}
\caption{Adversarial Initialization of a Weight Row.}
\label{algo:adv_weight_initialization}
        \KwIn{Weight row $\mathbf{w_i}
        $ of length $L$, Gaussian distribution $\mathcal{N}(\mu, \sigma)$ with mean $\mu$ and std $\sigma$, Scaling factor $s<1$, Discrete uniform distribution $\mathcal{U}(\cdot,\cdot)$} 
        \KwOut{Adversarially initialised weight row $\mathbf{w_i}$}
\BlankLine
\begin{minipage}{1.05\hsize}
    \begin{algorithmic}[1]
    \STATE $\negind = \{i | i \sim \mathcal{U}(1, L)\} \text{ where } |\negind| = \frac{1}{2}L$ \Comment*[r]{{\scriptsize Select randomly indices for negative weights}}
    \STATE $ \posind \leftarrow  \{i \notin \negind| i \in [L]\}$  \Comment*[r]{{\scriptsize Select indices for positive weights}}
    \STATE $\mathbf{z}_- \thicksim \mathcal{N}(\mu, \sigma) | \mathbf{z}_- \in \mathbb{R_-}^{\frac{1}{2}L}$ \Comment*[r]{{\scriptsize Negative samples}}
    \STATE $\mathbf{z}_+ = -s \cdot \mathbf{z}_-$ \Comment*[r]{{\scriptsize Positive samples}}
    \STATE $\mathbf{w_i}[\negind] \leftarrow \text{Shuffle}(\mathbf{z}_-)$ \Comment*[r]{{\scriptsize Initialize negative weights}}
    \STATE $\mathbf{w_i}[\posind] \leftarrow \text{Shuffle}(\mathbf{z}_+)$ \Comment*[r]{{\scriptsize Initialize positive weights}}
    
\end{algorithmic}
\end{minipage}
\end{algorithm2e}

Intuitively, our approach adversarially initializes each row of the weight matrix to increase the likelihood that only one data point in a given mini-batch will activate the neuron corresponding to that row. 
To achieve this, we initialize a randomly chosen half of the components of the weight row to negative values, and the other half to the corresponding positive values, by sampling from a Gaussian normal distribution.
The positive components of the weight row are scaled down with a small factor $s<1$ in comparison to the negative components.
This increases the impact of the negative components on the weighted input sum to the corresponding neuron.
This causes most input data points to produce non-positive input to the neuron, such that only a few (in the best case only one) input data point activates the neuron.
See \Cref{algo:adv_weight_initialization} for a formalization of our initialization.

We use the scaling factor $s$ to specify  how much larger the absolute values of the negative weight components should be than the positive values.
This determines how "aggressively" our activation causes weighted inputs to individual neurons to be negative, thereby to be filtered out by the $\ReLU$ function and to have zero gradients for most input data points.
The ideal value of $s$ when it comes to attack effectiveness is dataset-dependent.
The \attacker can to fine-tune $s$ either on a small amount of data from the \users' input domain it holds before sending the \namenoformat to the \users.
Alternatively, they can fine-tune $s$ without any data from the \users' input domain and solely by exploiting the passive data leakage, or using data with the same dimensionality as the \users' data as we show in \Cref{sub:active_attack_evaluation}.

Our adversarial initialization causes the ReLU function for many neurons at the fully-connected layer to activate only for one input data point per mini-batch.
Due to the randomness in the initialization of each weight row corresponding to a neuron, different neurons are likely to be activated by different input data points. 
Thereby, the gradients of different weight rows allow for the extraction of different individual data points.
We demonstrate the success of our \namenoformat for data extraction in \Cref{sub:active_attack_evaluation} by showing that they increase the proportion of neurons that only activate on one random individual data point in a mini-batch by more than factor 10, and thus we are able to extract more than double the number of individual training data points. \Eg our \namenoformat cause $51.4$\% of active neurons out of 1000 to by activated by individual data points from the ImageNet dataset while random model weights with a Gaussian normal initialization with $\sigma=0.5$ only yield $4.4$\%. This allows for an individual extraction of $45.7$\% of the data points in a mini-batch of size $B=100$ for our \namenoformat versus $21.8$\% with random model weights.

\subsection{Trap Weights for Other Architectures}
\label{sub:adv-forwarding-cnns}

To enable perfect extraction, our attack relies on the presence of a ReLU-based fully-connected layer at the beginning of the model architecture.
Since in FL, the \cp is in charge of instantiating the model architecture, this does not represent a practical limitation.

For some application domains, the \cp might want to train ML models beyond pure FC-NNs though, \ie, models where the first layer is not fully-connected. 
In Appendix~\ref{sec:adv_forwarding}, we show how an \attacker can apply malicious manipulations to shared model's weights to extend our attack to CNN-based architectures.
These architectures consist of several convolution layers and some fully-connected layers which the \attacker can leverage for extraction.
The intuition of the attack-extension is to convert the convolution layers to identity functions which transfer the \user's input data to the first fully-connected layer in the model.
The \attacker initializes this layer with our \namenoformat and can then extract \user data.
In Appendix~\ref{sub:cnns_detectability}, we discuss how to also make the manipulations of the convolutional layers most stealthy.

In the following section, we also evaluate extraction for text-data in model architectures that contain an embedding layer before the first fully-connected layer.
Extraction is done from the fully-connected layer whose input consists of the embedding layer's output.
To reconstruct the original text tokens from a sequence of extracted embeddings, the \attacker creates a lookup dictionary, mapping its initialized embeddings back to their corresponding tokens (this is the inverse mapping to the embedding layer).
To avoid vector-comparisons for each lookup, the \attacker uses hash values for vector embeddings as keys.

\section{Experimental Evaluation}
\label{sec:adv-initialize-evaluation}

In this section, we validate that our adversarial weight initialization attack allows a \cp to reconstruct individual training data points from gradients shared by \users. We use three different image datasets, namely MNIST~\cite{LeCun.2010MNIST}, CIFAR10~\cite{krizhevsky2009learning}, and ImageNet~\cite{deng2009imagenet} and the text-based IMDB~\cite{imdb} dataset for sentiment analysis.
Because our approach is applicable to FC-NNs and CNNs, we test it against both of these architectures. We instantiate our attack against an FC-NN for the MNIST dataset, and against a CNN for CIFAR10 and ImageNet. 
For the IMDB dataset, we use a model whose input is 250-token sentences, and consists of an \textit{embedding layer}, which maps each token in a 10,000-word vocabulary to a 250-dimensional floating-point vectors, and inputs these to a fully-connected layer.
The specifics of our model architectures for image and text data are described in~\Cref{tab:architectures} and \Cref{tab:architecture_text} in Appendix~\ref{app:additional_material}, respectively.
We implemented our \namenoformat, and the experiments in TensorFlow~\cite{Abadi.2016TensorFlow} version 2.4.
The code will be open sourced after the peer review process.

\myparagraph{Attack Instantiation.}
Since the \cp has access to the gradients of \textit{all} model layers uploaded by users, it is able to choose which layer to instantiate the attack on.

For the FC-NN, we adversarially initialize the first layer with our \namenoformat  and extract training data points from its gradients. 
In the CNNs, we first initialize the convolutional layers to transmit the input data to the first fully-connected layer of the architecture. Then, we adversarially initialize this layer's weights with our \namenoformat for extraction.

For the text classifier, we initialize the weights of the embedding layer with a random uniform distribution (min=0., max=1.) 
to create the inputs for the fully-connected layer.
We then adversarially initialize this fully-connected-layer's weights with our \namenoformat to perform extraction of the embeddings there.

\subsection{Extraction Success Metrics}
\label{sub:success-metrics}
We introduce three novel metrics to measure the success of individual data point extraction.

\myparagraph{Active Neurons.} 
By measuring the number of \act (A) we can determine for how many neurons their respective weighted inputs are positive.
This is important because data extraction for both overlaying and individual data points is only possible with activated neurons.
If a neuron is not activated by any data point, no information can be transmitted over this neuron and, hence, the gradients will all be zero.

\myparagraph{Extraction-Precision.} 
Our second metric, which we call \precision, captures the percentage of non-zero gradient rows at the given layer's weight matrix from which we can extract any input data point \emph{individually}.
This metric enables us to quantify how well the adversarial weight initialization manages to generate weights that cause activation for exactly one single data point.
\Precision can be calculated as follows:
\begin{equation}
    \label{eq:precision}
    \text{P}=\frac{G_1}{A}\text{,}
\end{equation}
with $A$ denoting the \act, and $G_1$ denoting the number of gradient rows from which we can extract a data point individually and with an $\Ltwo$-distance of zero to any of the input data points.

However, the \precision metric alone would not be expressive enough since a high \precision could be achieved despite the exact same individual training input being reconstructed from all gradient rows.
Therefore, we defined another metric that we call \recall.

\myparagraph{Extraction-Recall.} 
The \recall measures the percentage of input data points that can be perfectly extracted from any gradient row.
We define it by
\begin{equation}
    \label{eq:recall}
    \text{R}=\frac{B_0}{B}\text{,}
\end{equation}
where $B$ is the number of data points in the given mini-batch and $B_0$ is the number of these data points that we can extract with an $\Ltwo$-error of zero from the rescaled gradients.

\myparagraph{Interpretation of Success Metrics.}
Note that our attack seeks to find an adversarial initialization that balances setting enough neurons' outputs to zero (such that a gradient is more likely to isolate individual points from large mini-batches) with, at the same time, having enough neuron outputs' that are non-zero (otherwise, in the limit, no points would be extracted). 
Thus, \act provide additional context for the \precision: with few \act, even a high \precision might not be able to extract many individual training data points, simply because there are very few gradients to perform data extraction from.
However, with many \act, the \recall might become small, due to each neuron being most likely activated by several input data points, preventing individual extraction.

\subsection{Evaluating the Passive Attack}
\label{sub:passive_attack_evaluation}

\begin{table}[]
\scriptsize 
\centering
\begin{tabular}{lcccccc}
\toprule
{} & \multicolumn{2}{l}{\textbf{MNIST}} & \multicolumn{2}{l}{\textbf{CIFAR10}} & \multicolumn{2}{l}{\textbf{ImageNet}}\\
{\textbf{Weights Initializer}}  & P & R & P & R & P & R\\
\midrule
Xavier Normal                         &     .004 &  .037 &     .048 &  .203& .046 &  .213 \\
Xavier Uniform                       &     .005 &  .048 &     .053 &  .229& .040 &  .201 \\
Gaussian  ($\sigma$=0.01)            &     .005 &  .048 &     .051 &  .226&  .041 &  .203 \\
Gaussian  ($\sigma$=0.1)              &     .005 &  .049 &     .053 &  .238& .043 &  .209 \\
Gaussian  ($\sigma$=0.5)              &     .006 &  .050 &     .058 &  .255& .044 &  \textbf{.218} \\
Gaussian  ($\sigma$=1)                &     .006 &  .059 &     .058 &  .256& .045 &  \textbf{.218} \\
Gaussian  ($\sigma$=2)                &     .007 &  \textbf{.061} &     .058 &  \textbf{.259}& .047& .217\\
\bottomrule
\end{tabular}
\caption{\textbf{Extractability with Random Initializations.} Impact of random initialization functions on the \precision (P) and \recall (R) of individual training data points from the model gradients. The displayed numbers refer to a mini-batch of 100 data points and 1000 neurons for extraction in the first model layer (FC-NN architecture from \Cref{tab:architecture_text}). Results are averaged over 10 runs with different random initializations.}
\label{tab:effect_initializer_function}
\end{table}

\begin{table}[]
\centering
\scriptsize
\begin{tabular}{lrrr|rrr}
\toprule
{} & \multicolumn{3}{c}{Passive Attack} & \multicolumn{3}{c}{Our Active Attack} \\
{\textbf{B}} & A & P & R & A & P & R \\
\midrule
20  &                .842 &     .072 &  \textbf{.900} &                .519 &     .610 &  \textbf{1.000}\\
50  &                .885 &     .050 &  \textbf{.552} & .776 &     .376 &  \textbf{.962}\\
100 &               .909 &     .036 &  \textbf{.254} & .910 &     .192 &  \textbf{.654} \\
200 &                .927 &     .030 &  \textbf{.128} &.978 &     .070 &  \textbf{.255}\\
\bottomrule
\end{tabular}
\caption{\textbf{Data Extraction on IMDb Dataset.} The extraction success depends on the size \textbf{B} of the mini-batches for passive attack and active attack with adversarial initialization. The results depict the percentage of \act (A), \precision (P), and \recall (R). All numbers are averaged over 10 runs with different random and adversarial initialization of the model from \Cref{tab:architecture_text}, respectively.}  
\label{tab:text_imdb}
\end{table}

Recall from~\Cref{sec:data_leakage} that extraction of training data from gradients is possible even when model weights are initialized randomly. 
We evaluate this passive attack to obtain a baseline for our adversarial weight initialization strategies.
To evaluate the passive attack, we measure the extraction success of individual training data points from the gradients of randomly initialized models.

\Cref{tab:effect_initializer_function} reports the \precision and \recall of training data point extraction from the gradients of randomly initialized models. These gradients are computed over a mini-batch of 100 data points for 1000 neurons (\ie 1000 weight rows' gradients for extraction) in the first fully-connected layer. We later study the impact of these two parameters on the success of reconstruction attacks. 
Even if this attack is passive, and the \cp has not modified any of the weights adversarially, training data extraction is often successful: for the MNIST dataset, around $6$\% of individual training data points can be directly extracted from the model gradients, whereas for CIFAR10 and ImageNet, roughly $26$\% and $22$\% of the training data points can be perfectly extracted.
The passive attack for extracting embeddings from the IMDB dataset yields roughly $25$\% \recall for 1000 neurons and mini-batches of 100 data points, see~\Cref{tab:text_imdb}.

These results also suggest that setting higher spread, in form of standard deviations to random weight distributions alone can already significantly increase the \recall of individual data points from the model gradients, see~\Cref{tab:effect_initializer_function}.
This is, most likely, due to the larger span within the weight values.

\begin{table}[]
\centering
\scriptsize
\begin{tabular}{lrrrrrrrr}
\toprule
{} & \multicolumn{4}{c}{\textbf{MNIST}} & \multicolumn{4}{c}{\textbf{CIFAR10}} \\
\textbf{Epoch} &  Loss $\mathcal{L}$ & A & P & R &     Loss $\mathcal{L}$ & A & P & R \\
\midrule
0  & .526 &                .998 &     .005 &  .050 &   1.857 &                .907 &     .053 &  .232 \\
5  & .067 &                .997 &     .044 &  .137 &   1.352 &                .900 &     .044 &  .195 \\
10 & .021 &                .997 &     .116 &  .154 &   1.088 &                .913 &     .041 &  .196 \\
15 & .006 &                .997 &     .131 &  .165 &   .768 &                .923 &     .043 &  .206 \\
20 & .002 &                .997 &     .136 &  .167 &   .472 &                .931 &     .050 &  .232 \\
25 & .001 &                .997 &     .140 &  \textbf{.169} &   .282 &               .935 &     .058 &  .241 \\
30 & .001 &                .997 &     .142 &  .168 &   .200 &                .936 &     .062 &  \textbf{.267} \\
\bottomrule
\end{tabular}
\caption{\textbf{Data Extractability from Converging Models.} Results depict the success of passive data extraction based on the training stage of the corresponding models. We show the percentage of \act (A), \precision (P), and \recall (R) for extraction with a mini-batch size of 100 data points from the first layer of the fully-connected network from \Cref{tab:architectures}. All numbers are averaged over 10 runs with different \textit{random} initializations.}
\label{tab:extraction_converging_models}
\end{table}

\begin{figure}[tb]
\centering
\includegraphics[width=0.53\textwidth]{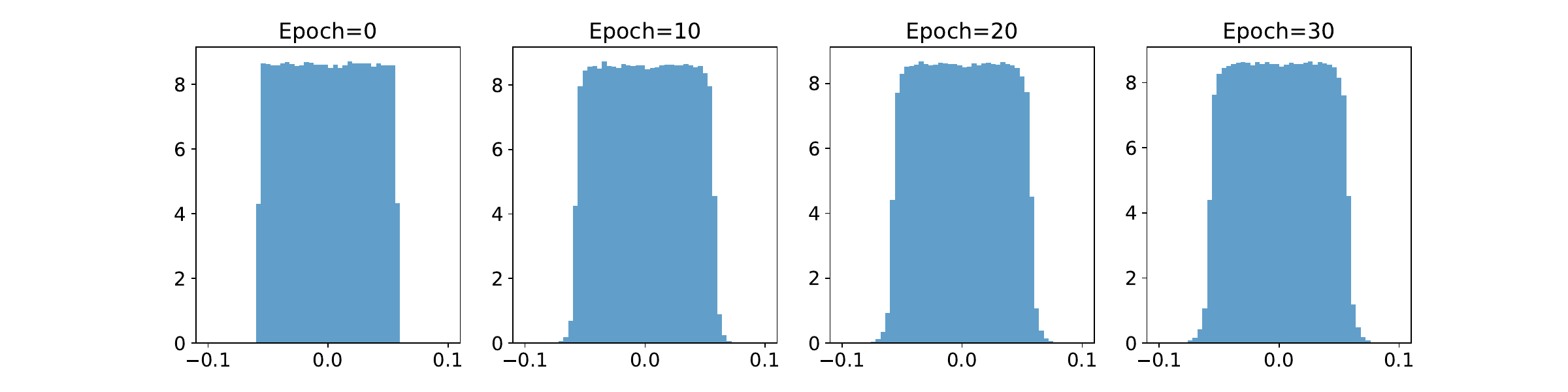}
\caption{\textbf{Evolution of Model Weights over Training.} Distribution of the first layer's weights of the FC-NN from \Cref{tab:architectures} over training on the MNIST dataset. Weights at epoch zero were initialized with a random uniform distribution.}
\label{fig:weights_after_training}
\end{figure}

Additionally, we also set out to investigate how as training progresses, and the model's weights converge, the extraction' success evolves.
We initialized the FC-NN from \Cref{tab:architectures} with a Xavier Uniform distribution and trained the model on MNIST and CIFAR10 for 30 epochs.
\Cref{tab:extraction_converging_models} depicts the results. 
We observe that the \recall increases slightly over the training epochs.
Analyzing the distribution of the model weights in \Cref{fig:weights_after_training} shows that over training, the uniformly initialized weight values resemble more a normal distribution and obtain a wider spread, which might be the reason for the increased extraction success.

\subsection{Evaluating Active Manipulations}
\label{sub:active_attack_evaluation}

\begin{table}[t]
\tiny
\centering
\begin{tabular}{lccccccccc}
\toprule
{} & \multicolumn{3}{c}{MNIST} & \multicolumn{3}{c}{CIFAR10} & \multicolumn{3}{c}{ImageNet} \\
{\textbf{s}} & A & P & R & A & P & R & A & P & R \\
\midrule
$.400$   &                .022 &     .803 &  .114 &                0. &     0. &  0. &                .0. &     0. &  0. \\
 $.500$   &                .149 &     .636 &  .354 &                0. &     0. &  0. &                0. &     0. &  0. \\
$.600$   &                .462 &     .408 &  .526 &                0. &     0. &  0. &                0. &     0. &  0. \\
$.700$   &                .796 &     .203 &  \textbf{.540} &                0. &     0. &  0. &                0. &     0. &  0. \\
$.800$   &                .959 &     .062 &  .334 &                0. &     0. &  0. &                0. &     0. &  0. \\
$.900$   &                .996 &     .010 &  .089 &                .034 &    .946 &  .077 &                0. &     0. &  0. \\
$.950$  &                .999 &     .003 &  .029 &                .729 &     .412 &  \textbf{.540} &                0. &     0. &  0. \\
$.960$  &                .999 &     .003 &  .027 &                .925 &     .175 &  .522 &               0. &     0. &  0. \\
$.970 $ &                1. &     .002 &  .020 &                .993 &     .025 &  .198 &                .002 &     .900 &  .013 \\
$.980$  &                1. &     .002 &  .021 &                1. &     .001 &  .008 &                .043 &     .986 &  .049 \\
$.990$  &                1. &     .002 &  .020 &                1. &     0. &     0. &                .655 &    .514 & \textbf{ .457} \\
$.995$ &                1. &     .002 &  .018 &                1. &     0. &     0. &                .999 &     .007 &  .055 \\
$.999$ &                1. &     .002 &  .017 &                1. &     0. &     0. &                1. &     0. &     0. \\
\bottomrule
\end{tabular}
\caption{\textbf{Impact of Hyperparameter $\mathbf{s}$.} Success of our adversarial weight initialization dependent on the hyperparameter $s$, which downscales the positive weights. The results depict the percentage of \act (A), \precision (P), and \recall (R) with a mini-batch size of 100 data points from the first fully-connected layer of the respective architectures from \Cref{tab:architectures}. All numbers are averaged over 10 runs with different adversarial initializations.}
\label{tab:evaluation_hyperparameters}
\end{table}

\begin{table}[tbh]
\tiny
\begin{tabular}{crrrrrrrrr}
\toprule
          & \multicolumn{3}{l}{\textbf{MNIST}} & \multicolumn{3}{l}{\textbf{CIFAR10}} & \multicolumn{3}{l}{\textbf{ImageNet}} \\
 \textbf{(B, N)}&  A & P & R & A & P & R & A & P & R  \\
\midrule
 (200, 20)  &                .522 &     .436 &  \textbf{.720} &                .454 &     .670 &  \textbf{.695} &                .090 &     .948 &  \textbf{.355} \\
      (200,50)  &                .690 &     .302 &  .428 &                .662 &     .494 &  .452 &                .381 &     .763 &  .304 \\
      (200,100) &                .782 &   .196 &  .218 &                .846 &    .280 &  .269 &                .653 &     .500 &  .240 \\
      (200,200) &                .859 &     .121 &  .086 &                .954 &     .124 &  .096 &                .886 &     .233 &  .113 \\
 (500,20)  &                .535 &     .451 &  \textbf{.915} &                .452 &     .689 &  \textbf{.870} &                .096 &     .939 &  \textbf{.490} \\
      (500,50)  &                .697 &     .301 &  .624 &                .653 &     .505 &  .614 &                .387 &     .767 &  .426 \\
      (500,100) &                .792 &     .205 &  .397 &                .845 &     .290 &  .422 &                .646 &     .508 &  .358 \\
      (500,200) &                .871 &     .129 &  .185 &                .950 &     .119 &  .177 &                .892 &     .240 &  .199 \\
 (1000,20) &                .539 &     .444 &  \textbf{.950} &                .441 &     .703 &  \textbf{.915} &                .102 &     .942 &  \textbf{.595} \\
      (1000,50)  &                .705 &     .300 &  .760 &                .648 &     .504 &  .724 &                .388 &     .770 &  .516 \\
      (1000,100) &                .796 &     .203 &  .540 &                .844 &     .297 &  .556 &                .655 &     .514 &  .457 \\
      (1000,200) &                .871 &     .124 &  .293 &                .951 &     .120 &  .256 &                .892 &     .238 &  .288 \\
 (3000,20)  &                .541 &     .442 &  \textbf{1.} &                .441 &     .696 &  \textbf{.945} &                .101 &     .934 &  \textbf{.640} \\
      (3000,50)  &                .704 &     .299 &  .888 &                .646 &     .503 &  .812 &                .386 &     .764 &  .586 \\
      (3000,100) &                .797 &     .203 &  .746 &                .840 &     .286 &  .711 &                .649 &     .518 &  .579 \\
      (3000,200) &                .873 &     .129 &  .504 &                .951 &     .122 &  .414 &                .889 &     .243 &  .404 \\
\bottomrule
\end{tabular}
\caption{\textbf{Effect of Mini-Batch Size and Number of Neurons on Data Extraction.} Success of our adversarial weight initialization is dependent on the mini-batch size \textbf{B} and the number of neurons \textbf{N} that corresponds to the number of weights rows. The results depict the percentage of \act (A), \precision (P), and \recall (R). All numbers are averaged over 10 runs with different adversarial initializations.}
\label{tab:evaluation_batch_size_neurons}
\end{table}

We now turn to our active attack, which implements our trap weights to amplify the vulnerability exploited by passive attacks. 
This amplification is controlled by the scaling factor $s$ in the \namenoformat. We first  evaluate the impact of this scaling factor on the reconstruction quality of individual training data points over a mini-batch of 100 data points and 1000 neurons.

\begin{figure}[t]
\centering

\centering
\begin{subfigure}[b]{0.22\textwidth}
\centering
\includegraphics[width=\linewidth]{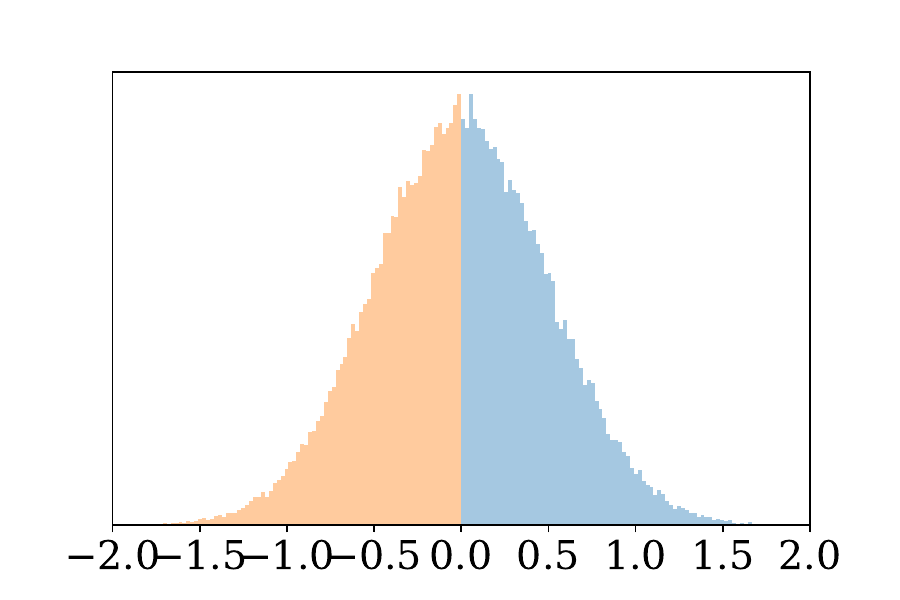}
\caption{$s=1$, Baseline}
\label{fig:CDP}
\end{subfigure}
\begin{subfigure}[b]{0.22\textwidth}
\centering
\includegraphics[width=\linewidth]{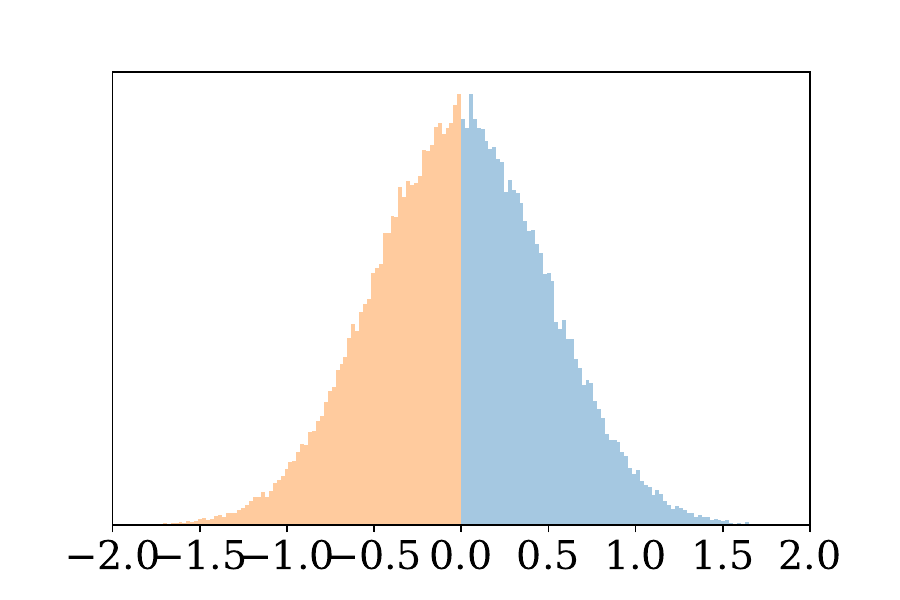}
\caption{$s=0.99$ (ImageNet)}
\label{fig:ldp}
\end{subfigure} 
\begin{subfigure}[b]{0.22\textwidth}
\centering
\includegraphics[width=\linewidth]{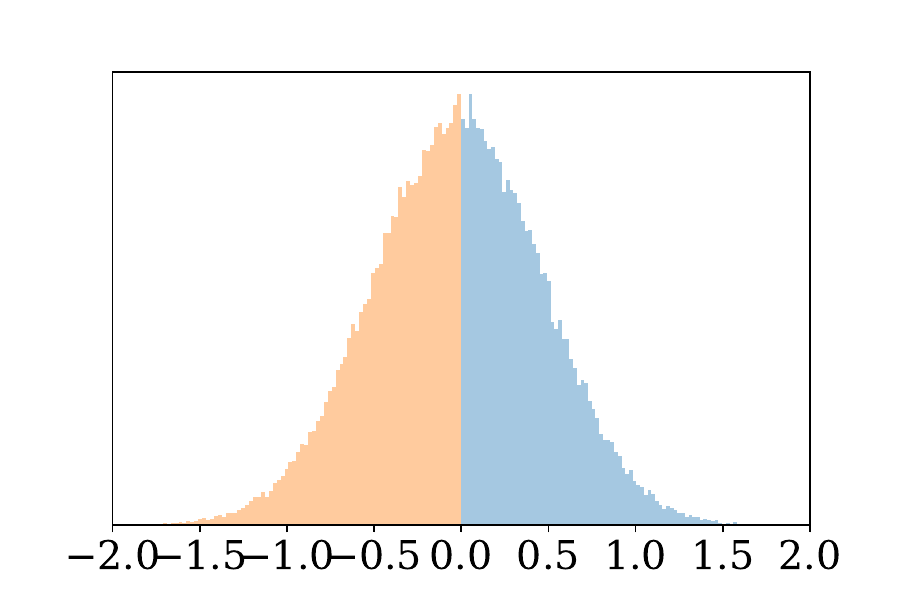}
\caption{$s=0.95$ (CIFAR10)}
\label{fig:ddp10}
\end{subfigure} 
\begin{subfigure}[b]{0.22\textwidth}
\centering
\includegraphics[width=\linewidth]{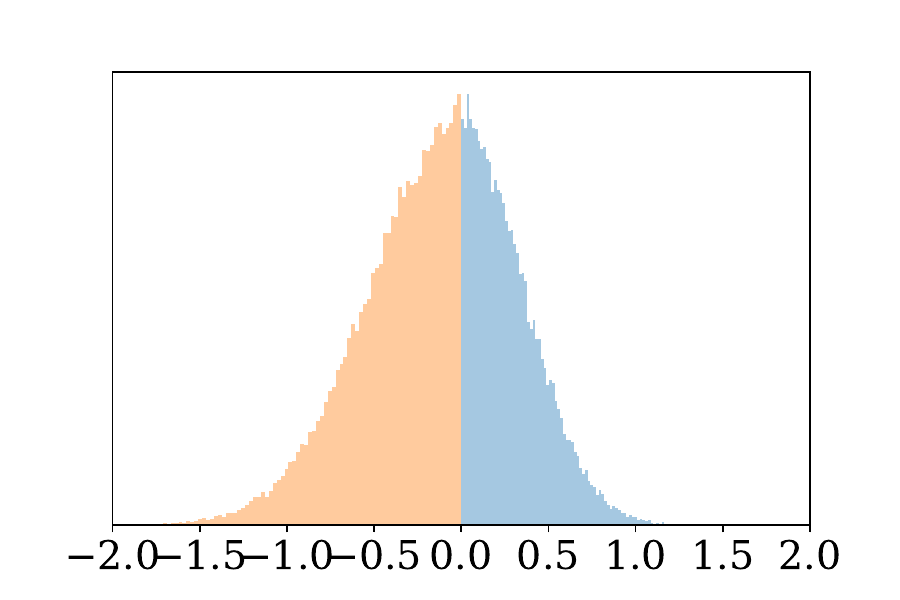}
\caption{$s=0.7$ (MNIST)}
\label{fig:ddp1000}
\end{subfigure}  

\caption{\textbf{Influence of s on Trap Weight Distribution.} When $s=1$, the distribution of weights follows the standard Gaussian normal distribution (here $\sigma=0.5$). 
This corresponds to the baseline of random initialization. 
For ImageNet and IMDB (b) and CIFAR10 (c), the difference in distribution to the random Baseline (a) is negligible.
} 
\label{fig:values_s}
\end{figure}

\Cref{tab:evaluation_hyperparameters} depicts the results, averaged over ten different random adversarial initializations. 
We can see that the best scaling factor for MNIST, when it comes to the \recall, is $s=0.7$.
With this scaling factor, we are able to extract on average $54.0$\% of the individual training data points which were involved in the \users' gradient computations. 
This is an improvement by around factor nine to the passive attack.
For CIFAR10 and Imagenet, the best scaling factors concerning \recall are $s=0.95$, and $s=0.99$, which allow for a perfect reconstruction of $54.0$\%, and $45.7$\% of the individual training data points, respectively, for 1000 neurons and a mini-batch size of $100$ data points.
Thereby, the active attack is more than twice as successful as the passive attack for extracting individual training data points in these datasets.
\Cref{fig:mnist_first_layer}, \Cref{fig:cifar_all}, and \Cref{fig:imagenet_first_layer} in the Appendix~\ref{app:additional_results} show the visual reconstruction results of the best run for the MNIST, CIFAR10, and ImageNet dataset, respectively.
For CIFAR10, we additionally present extraction success when all local data stems from the same single class.

Similar improvements of performance could be achieved for the IMDB dataset.
The best extraction was achieved also with $s=0.99$, for which, with 1000 neurons and mini-batches of 100 data points, we obtained an \recall of $65.4$\%, which is around 2.5 time as high as the passive attack, see \Cref{tab:text_imdb}.

In \Cref{fig:values_s}, we show the influence of the scaling factor $s$, our method's hyperparameter, on the distribution of our \namenoformat.
The case $s=1$ corresponds to the baseline where positive components in the \namenoformat are not scaled down.
The figure shows that the more $s$ deviates from $1$, the larger the difference between a random distribution and our \namenoformat. 
For CIFAR10 and ImageNet ($s=0.95$, and $s=0.99$), our \namenoformat's distribution is very close to the the random distribution, making our \namenoformat more stealthy.
The best scaling factor for MNIST, $s=0.7$, is significant smaller than for ImageNet and CIFAR10 due to the sparsity in the data (the background in MNIST images consists of zero pixels).
Our experiments indicate that with decreasing sparsity and increasing data dimensionality, $s$ approaches 1. 
Especially the last observation makes sense since scaling more positive components with a factor closer to 1 is in effect of the weighted sum equivalent to scaling fewer positive components with a factor much smaller than 1. 
Thereby, our \namenoformat increase in stealthiness with increasing complexity of the data to be extracted.

As hypothesized above, from \Cref{tab:evaluation_hyperparameters}, we furthermore confirm that the \recall of our attack is related to the percentage of \act: 
When very few neurons are activated, it is not possible to extract large numbers of individual data points due to the lack of gradients to extract them from.
However, when the percentage of \act is high, the \recall also becomes very small, which is due to the fact that each neuron gets activated by several input data points, and thereby, individual extraction is impossible. 

\myparagraph{Attacker without Auxiliary Data.}
We experiment with an attacker who does not have access to a small mini-batch of data from the \users' distribution to tune the scaling factor $s$ of our \namenoformat.
In this setup, the only knowledge an attacker holds is about the dimensionality of the \users' data which it needs to instantiate an adequate model architecture.
We evaluate three attacks in this setup.
\textit{1) Exploiting passive data leakage and composing a tuning dataset:} the attacker randomly initializes the model in a first round of the protocol.
Our results in \Cref{tab:effect_initializer_function} show that also randomly initialized models' gradients leak significant fractions of the \users' data (MNIST 6.1\%, CIFAR10 25.9\%, and ImageNet 21.7\%).
By plotting the \user's gradients and eyeballing which data points resemble natural images, the attacker can build a tuning set for $s$. 
Since we only require a maximum of 100 data points to find the optimal values for $s$ per dataset in \Cref{tab:evaluation_hyperparameters}, the attacker only has to inspect the gradients of 17, 4, and 5 \users for MNIST, CIFAR10, and ImageNet, respectively in the first round of the protocol.
On the selected data, they can tune $s$ and use it in every subsequent iteration.
We performed tuning on 100 data points obtained through passive extraction and obtained the same $s$ as through tuning on a random mini-batch of data (0.7, 0.95, and 0.99 for MNIST, CIFAR10, and ImageNet, respectively).
\textit{2) Exploiting raw passive data leakage:} Since manually, selecting suitable data points is time-consuming, we propose an alternative approach where the attacker uses all extracted data points with are in a valid range for input pixels ([0,1]) from the passive extraction on non-adversarially initialized model weights in the first round of the protocol.
These data points are not necessarily individually extracted \user data points as we show in \Cref{fig:extracted_first_passive} in Appendix~\ref{app:additional_insights_trap_weights}.
But the attacker can still consider them as a tuning dataset for $s$ and evaluate the \recall on this dataset when initializing the shared model with different \namenoformat to tune $s$.
Our results in \Cref{tab:extraction_from_random_gradients} show that for CIFAR10 and ImageNet, the best $s$ found on these passively reconstructed data points are equal to the best $s$ obtained directly by tuning on one mini-batch of the original data.
For MNIST, the $s$ on the extracted gradients differs slightly from the original best $s$ (0.75 vs. 0.7).
We suspect these changes to result from MNIST data being much sparser (many more zero features) than the extracted gradients in \Cref{subfig:MNIST_first}.
\textit{3) Using a surrogate dataset of same data dimensions:} Lastly, the attacker can tune $s$ on a surrogate dataset of the same dimension (but potentially different distribution) than the \users' data.
We compare \recall of an adversarial weight initialization with $s$ found on a surrogate dataset and the optimal $s^*$ found on the actual dataset for Fashion MNIST, SVHN, CIFAR100, and Open Images
~\cite{OpenImages} in \Cref{tab:surrogate_data}.
Our results highlight that extraction with the surrogate $s$ obtained through tuning on MNIST, CIFAR10, and ImageNet, already yields a significantly higher success than passive extraction on non-manipulated weights.
Furthermore, the closer the surrogate dataset's distribution is to the users' dataset, the closer $s$ and $s^*$.
Especially for CIFAR10 and CIFAR100, and ImageNet and Open Images, we find that $s=s^*$ which leads to highest extraction success.

\begin{table}[tb]
\centering
\scriptsize
\begin{tabular}{cccccc}
\toprule
  \textbf{Dataset}      & Passive  R & $s$ &  R with $s$  & $s^*$ & R with $s^*$ \\
\midrule
 Fashion MNIST      &   0.09    &   0.7     & 0.22  & 0.77  &   \textbf{0.31}\\
 SVHN               &   0.22    &    0.95   & 0.26  & 0.97  &   \textbf{0.40}\\
 CIFAR100           &   0.25    &    0.95   &\textbf{0.42}   & 0.95  &   \textbf{0.42}\\
 Open Images        &   0.21    &     0.99  &\textbf{0.44}  & 0.99  &   \textbf{0.44}\\
\bottomrule
\end{tabular}
\caption{\textbf{Surrogate Data for Tuning $s$.} We report \recall for passive extraction and extraction under adversarial weight initializations.
For the latter, we compare the extraction under an $s$ found on a surrogate dataset, and the optimal $s^*$ found through tuning on 100 data points from the given datasets.
As surrogate datasets, we use MNIST for Fashion MNIST, CIFAR10 for SVHN and CIFAR100, and ImageNet for Open Images.
Results are averaged over 5 runs.}
\label{tab:surrogate_data}
\end{table}

\myparagraph{Impact of Data Labels.} Additionally, we investigated whether this high reconstruction success could also be achieved =in a non-IID setting when \users hold local mini-batches of data that belongs to one single class, different from the other \users. 
This is a particularly challenging setting for prior work on optimization-based attacks that end up reconstructing average points rather than individual points exactly. 
Instead, \Cref{fig:cifar_first_layer_dog} and \Cref{tab:effect_nonIID} in Appendix~\ref{app:additional_insights_trap_weights} show on CIFAR10, how our method remains able to perfectly extract individual data points from the gradients even when all points stem from the same class.

\begin{figure}[t]
\centering

\begin{subfigure}[b]{0.35\textwidth}
\centering
\includegraphics[width=\linewidth]{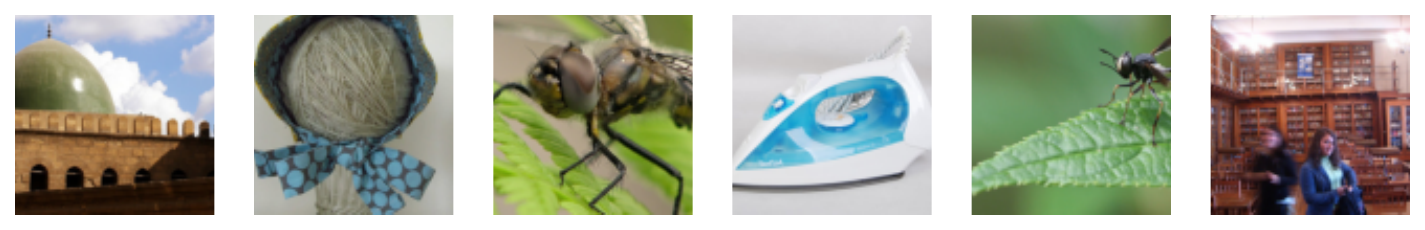}
\vspace{-0.6cm}
\caption{Original Data. Dimension=(224,224,3).}
\label{subfig:original}
\end{subfigure} 
\vfill
\begin{subfigure}[b]{0.35\textwidth}
\centering
\includegraphics[width=\linewidth]{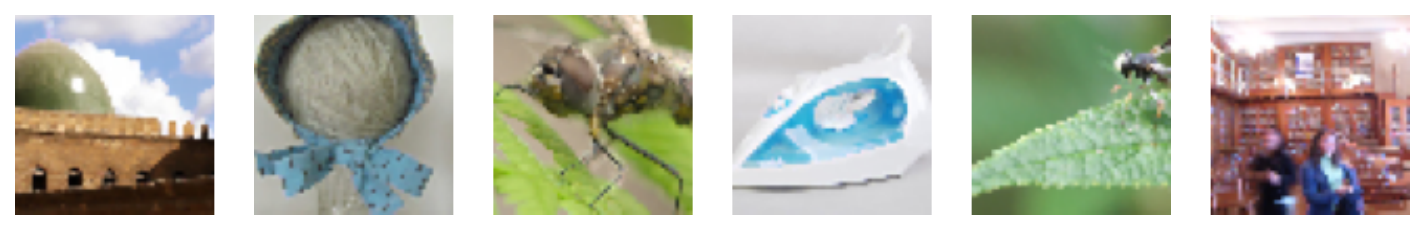}
\vspace{-0.6cm}
\caption{VGG7 Extracted. Dimension=(56,56,3).}
\label{subfig:vgg7}
\end{subfigure}  
\vfill
\begin{subfigure}[b]{0.35\textwidth}
\centering
\includegraphics[width=\linewidth]{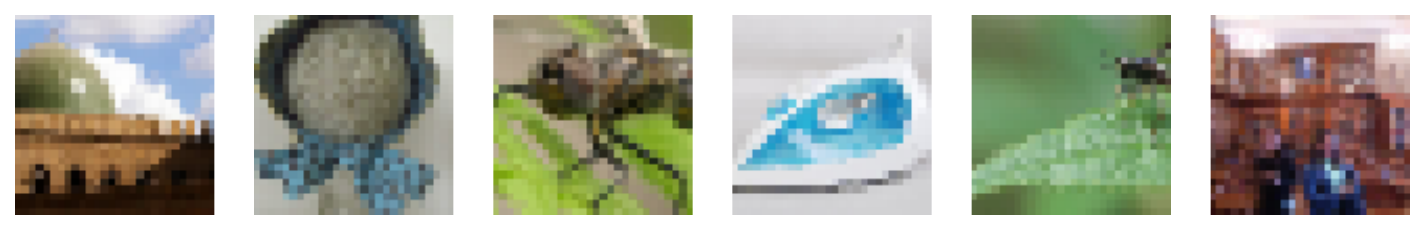}
\vspace{-0.6cm}
\caption{ResNet Extracted. Dimension=(28,28,3).}
\label{subfig:resnet20}
\end{subfigure}  
\caption{\textbf{Extraction from Standard Architectures.}
We extract individual user data points from the first fully-connected layer after the convolutional layers.
The compression of extracted data in comparison to the original data results from the pooling layers in the architectures.}
\label{fig:vgg_resnet20}
\end{figure}

\myparagraph{Impact of Mini-Batch Sizes.} We also set out to investigate the impact of the mini-batch size $B$ and the number of weight rows that we can use for extraction.
\Cref{tab:evaluation_batch_size_neurons} depicts the resulting metrics.
The metrics show that the smaller the mini-batch sizes are, and the more weight rows there are for extraction, the more individual training data points can be individually reconstructed.
For $3000$ weight rows, even up to $50$\% of the individual training data points for mini-batch sizes as large as 200 in the MNIST dataset can be perfectly extracted.
Small mini-batches of $20$ training data points are entirely extractable without any loss in this setting.
Also for the IMDB dataset, smaller batch-sizes for the same number of neurons yield much higher \recall, and embeddings of data from small mini-batches of $20$ training data points are perfectly extractable, see \Cref{tab:text_imdb}.
This suggests that in practice, the success of the extraction attack can be significantly increased by the \cp demanding smaller mini-batch sizes from the \users or initializing larger models. 

\myparagraph{Impact of Lossy Layers.}
For perfect extraction of data points in CNN architectures, our attack requires the input to the first fully-connected layer to have at least as many parameters as the original input data point. 
CNN architectures can contain pooling layers to reduce input size.
In Appendix~\ref{sub:dropout_maxpool}, we evaluate the impact of pooling on the fidelity of the extracted data.
Our evaluation shows that pooling results in some form of compression of the \user's input data, see for example \Cref{fig:pooling_alone}.
In Appendix~\ref{sub:cnns_reduceSize}, we show how the \cp can implement an alternative to pooling for size-reduction in CNNs based on convolutional layers which still allows for prefect extracability, as long as there are enough model parameters.
We also evaluate the effect of dropout on the fidelity of the extracted data.
\Cref{fig:dropouts} and \ref{fig:dropouts_20} visualize the effect of pure dropout, while \Cref{fig:droppools} and \ref{fig:droppools_20} visualize the joint effect of dropout and pooling.
To increase fidelity of extraction under lossy layers, an attacker can apply post-processing, such as de-compression.

\myparagraph{VGG and ResNet.}
In addition to our custom FC-NN and CNN architecture from \Cref{tab:architectures}, we experimented with a VGG7 and a ResNet20~\cite{resnet20} architecture.
For VGG7, we initialize all convolutional layers as illustrated in \Cref{fig:conv} in Appendix~\ref{app:size_preserving_filters}, and the fully-connected layer directly after the convolutional layers with our \namenoformat.
For the ResNet20, we only initialize the first convolutional layer according to \Cref{fig:conv}. 
Thanks to the skip connections, the remaining convolutional layers can remain unchanged, apart from the convolutional filters whose output is added to the output of the skip connections.
These need to be set to zero, such that the input data can be propagated unaltered over the skip-connections to the fully-connected layer that we initialize with our \namenoformat.
We set the last pooling layer before this fully-connected layer in ResNet20 to implement average pooling. 
Our extraction results for ImageNet are depicted in \Cref{fig:vgg_resnet20}.
The compression of extracted data in comparison to the original data results from the pooling layers in both architectures that reduce input dimensions.

\begin{table}[t]
\tiny
\centering
\begin{tabular}{ccrrrrrr}
\toprule
 local & local  & \multicolumn{3}{c}{MNIST} & \multicolumn{3}{c}{CIFAR10} \\
  epochs&  B & \textuparrow$\Delta_\text{acc}$ & P & R & \textuparrow$\Delta_\text{acc}$ & P & R \\
\midrule
1 & 10 &     0.296 &     0.238 &  0.704 &  0.264 &     0.279 &  0.584 \\
  & 20 &     0.220 &     0.175 &  0.496 &  0.238 &     0.277 &  0.466 \\
  & 40 &     0.289 &     0.109 &  0.280 & 0.188 &     0.127 &  0.213 \\
2 & 10 &     0.384 &     0.252 &  0.704 & 0.316 &     0.311 &  0.560 \\
  & 20 &     0.772 &     0.184 &  0.478 & 0.296 &     0.295 &  0.476 \\
  & 40 &     0.671 &     0.111 &  0.264  &0.282 &     0.126 &  0.247 \\
3 & 10 &     0.604 &     0.241 &  0.712 & 0.372 &     0.329 &  0.596 \\
  & 20 &     0.790 &     0.189 &  0.544  &0.420 &     0.277 &  0.496 \\
  & 40 &     0.823 &     0.111 &  0.283  &0.324 &     0.138 &  0.257 \\
4 & 10 &     0.644 &     0.251 &  0.692 & 0.396 &     0.332 &  0.632 \\
  & 20 &     0.848 &     0.178 &  0.494&  0.440 &     0.288 &  0.478 \\
  & 40 &      0.825 &     0.113 &  0.273& 0.415 &     0.138 &  0.256 \\
5 & 10 &     0.604 &     0.270 &  0.732 & 0.412 &     0.354 &  0.620 \\
  & 20 &     0.870 &     0.178 &  0.494 & 0.492 &     0.289 &  0.502 \\
  & 40 &     0.873 &     0.119 &  0.283 & 0.461 &     0.139 &  0.277 \\
\bottomrule
\end{tabular}
\caption{\textbf{Local Accuracy Improvement and Extraction Success with FedAvg.} 
We present results for FedAvg where each user holds five mini-batches of size B and computes 1,2,3,4, or 5 local epochs of training with the FC-NN from \Cref{tab:architectures}.
The $\Delta_\text{acc}$ indicates the accuracy improvement on the user's local data w.r.t. the received shared model (initially around 10\% accuracy).
The \precision (P) and \recall(R) for every B stay at the same high level even after multiple local epochs of training.}
\label{tab:fedavg}
\end{table}

\myparagraph{Impact of Local Mini-Batch-Averaging.} 
Additionally, we looked into the effect of averaging over the gradients of multiple mini-batches, \eg the average of gradients received from multiple \users.
The results in \Cref{tab:effect_averaging} in Appendix~\ref{app:additional_insights_trap_weights} show that through averaging, the attack success is significantly reduced.
Already when averaging over $20$ mini-batches of size $B=100$ in the MNIST dataset, the average \recall drops from $54.0$\% to $2$\% because multiple data points overlay in the gradients.
This highlights that the \cp needs to perform the extraction before the averaging operation.
The following section shows that this simple change to the protocol is easily implemented by an actively dishonest \cp, even for standard FL libraries.

\myparagraph{FedAvg.}
To validate our theoretical insights from \Cref{sub:fedavg_theory} which highlights that even under FedAvg, perfect extraction of individual data points is possible, we run FedAvg experiments in which \users hold five mini-batches of \{10, 20, or 40\} different data points (yielding a total of 50, 100, and 200 local data points per user), and perform \{1, 2, 3, 4, or 5\} local training epochs.
In \Cref{tab:fedavg}, we depict results from the FC-NN architecture from \Cref{tab:architectures} on MNIST and CIFAR10.
Our results highlight that the while accuracy on the \users' data significantly increases through the local training, the extraction success stays constant over multiple local epochs.
We even observe a slight increase in \recall. 
For example, we can extract $58.4$\% of data points from users who hold five mini-batches with ten data points each after one epoch, while this number increases to $62$\% after five local epochs of training.
This finding is congruent with our finding in \Cref{tab:extraction_converging_models}, where we show that extraction success increases with convergence.
For CNNs, the extraction success degrades over multiple local epochs of training. 
This is due to the convolutional filters that, after a local update, do not have the zero-elements anymore which prevent features in the forward-pass from overlapping.
For our CNN architecture from \Cref{tab:architectures}, we report $\Ltwo$-distances between original and extracted data of $[3.81\mathrm{e}{-5}, 5.06\mathrm{e}{-5}, 0.07,  0.27, 0.92]$ and $[1.4\mathrm{e}{-3},138.94,199.69,264.45,269.32]$ for CIFAR10 and ImageNet after 1,2,3,4, and 5 epochs, respectively.

\myparagraph{TensorFlow Federated.}
\label{sub:tf-federated}
We experimented with TensorFlow Federated~\cite{tffed}---a standard open source library for FL deployments.
In Appendix~\ref{app:tf-federated}, we show that a dishonest \cp only requires minimal code changes to implement our \namenoformat.

\subsection{Comparison To Previous Work}
\label{sub:comparison_previous_work}

To compare the success of our attack, we compare to the three approaches conceptually closest to ours.
These are~\cite{Geiping.2020Inverting} which is the first to describe individual extractability of single-data point gradients,~\cite{fowl2021robbing} which relies on direct extraction from a fully-connected layer in the model architecture, and~\cite{pasquini2021eluding} which exploits manipulations model parameters to extract data from \user-gradients.

\myparagraph{Comparison to~\cite{Geiping.2020Inverting}.}
For the sake of correctness, we build on their code base and adopt it to also run with neural architectures we used in our other experiments. We use the parameters that~\cite{Geiping.2020Inverting} found to perform best. Here, we are mainly interested in the quality of the other attack's reconstruction in comparison to our method, and in the number of passes over the model, \ie the computing time required to obtain the reconstructions.
We perform evaluation on the MNIST and CIFAR10 datasets.

\Cref{fig:mnist_recon_results} in Appendix~\ref{app:additional_results} shows an example of the gradient-inversion fidelity obtained with~\cite{Geiping.2020Inverting} for MNIST with $B=1$.
Within $10^4$ iterations, the pixel-wise $\Ltwo$ errors observed go as low as $10^{-4}$ for FC-NNs with architecture as depicted in \Cref{tab:architectures} in the Appendix, and $10^{-3}$ for LeNet-5. Similar results for CIFAR10 are available in Appendix~\ref{fig:cifar10_recon_results} in~\Cref{app:additional_results}. In contrast to these results, our attack allows us to extract data points perfectly, \ie with $\Ltwo$ error of zero, and without \emph{any} back-propagation iterations. 
These results make clear that gradient inversion in practice suffers from local minima and requires a very large number of iterations to converge to a comparable reconstruction to our method. On a practical note, reconstruction of a single CIFAR10 image with 32 restarts from a seven-layer FC-NN takes on average 1 hour and 3 minutes on a high-end GPU, in comparison to milliseconds needed for extraction with the help of our \namenoformat. Most importantly, it is clear that reconstruction is not a lossless process and full data recovery is almost never possible, even in the simple cases where gradients of only a single data point are considered. To better understand limitations of prior literature we refer the reader to~\cite{wainakh2021federated}.

\myparagraph{Comparison to~\cite{fowl2021robbing}.}
The success of~\cite{fowl2021robbing}' data extraction depends on the size of their imprinting module. 
Using their code-base and extending it with our success metrics, we evaluated what size of imprinting module they require to obtain the same extraction recall as we do, \ie, to extract the same number of data points from the model gradients perfectly.
We compared our methods for all three vision dataset, using a batch-size of $B=100$.
For ImageNet and CIFAR10, following their baseline, we instantiated their model with a ResNet18, for MNIST, we used LeNet5.
We always inserted their imprinting module before the first layer to allow for perfect extractability with their method.
Our results show that to obtain the same extraction recall as we do (46\%, 54\%, and 54\% for ImageNet, CIFAR10, and MNIST, respectively), their imprinting module needs to be of size roughly 150, 200, and 400, respectively.
The fact that they require the largest imprint module for MNIST is due to the similarity in the data points (sparsity in the background with all zero pixels) which makes their binning less effective.

\myparagraph{Comparison to~\cite{pasquini2021eluding}.}
Note that~\cite{pasquini2021eluding}'s main goal is not to extract large amounts individual \user \textit{data points} but \user \textit{updates}, by circumventing the secure aggregation~\cite{bonawitz2017practical} used to protect the FL protocol. %
The updates (gradients) that~\cite{pasquini2021eluding} recover do usually not correspond to full and perfectly individual data points. 
Instead, for FC-NNs, their extracted gradients will resemble our passive extraction results from \Cref{fig:100-gradients-cifar10} where most of the gradients are a blurry overlay of all underlying data points.
For CNNs, their results will not be able to extract any individual data point since non-maliciously initialized convolution filters overlay input features.
Thereby, their attack mainly violates confidentiality of the users' model updates in a setup where users believe to obtain protection though an aggregate with other users.
Still, the resulting gradients can then be used as a departure point for additional privacy-attacks, such as reconstruction.
In contrast, our work directly violates the \users' privacy by manipulating the shared model weights to make individual data points directly extractable from the model updates sent from \users to the \cp.
To assess individual extractability in their setup, we use their gradient suppression and model inconsistency attack to make all but one \user in a round of the FL protocol return zero gradients.
The one target-user receives a randomly initialized FC-NN (Gaussian with $\sigma=0.5$) with architecture from~\Cref{tab:architectures}.
Note that the data extraction from gradients in this setup corresponds to our passive extraction.
For MNIST, with $B=100$, extraction in their setup yields $~5$\% of perfectly extractable data points, while our \namenoformat yield $54$\%.
In the same setup for CIFAR10, their method yields $26$\%, in contrast to our \namenoformat which yield again $54$\%.

\section{Defending Privacy in Federated Learning}
\label{sec:defenses}

This section discusses potential mitigations against our \namenoformat attack. %
We start with explaining DP which provides formal privacy guarantees and then move on to defenses specifically tailored to our \namenoformat attack.

\subsection{Formal Guarantees: Differential Privacy}
To bound the leakage of private information from model gradients, a gold standard for reasoning about privacy guarantees is the framework of differential privacy (DP)~\cite{Dwork.2006Differential}.

There exist three main ways of integrating DP in the FL protocol, namely Centralized Differential Privacy (CDP), Local Differential Privacy (LDP), and Distributed Differential Privacy (DDP).

In CDP, \users clip their gradients locally according to a clip norm $c$ and the \cp performs the addition of noise with a scale dependent on the noise multiplier $\sigma$~\cite{ramaswamy2020training}.  
CDP cannot provide DP guarantees with a malicious \cp because this \cp can simply extract \user data before adding noise or not add noise at all.\footnote{
For a private aggregation of sensitive statistics (instead of high-dimensional ML model gradients), there exist solutions of CDP without a trusted aggregator~\cite{roth2020orchard, roth2019honeycrisp}.}
    
LDP reduces the trust required in the \cp since every \user locally adds noise to their gradients according to their privacy requirements~\cite{truex2020ldp}.
Independent of other \users, the noise is drawn from $\mathcal{N}(0, \sigma^2c^2)$. 
However, previous work has shown that this setup leads to poor privacy-utility trade-offs, such that LDP is not popular in practical applications~\cite{wei2020federated}.

DDP is supposed to combine the advantages of CDP and LDP. 
In DDP, before aggregation, each \user locally adds some (small) amount of noise to their gradients~\cite{truex2019hybrid}.
The noise distribution depends on the number $\selectusers$ of other selected \users.
It is specified by $\mathcal{N}\left(0,  \frac{\sigma^2}{\selectusers-1}c^2\right)$~\cite{truex2019hybrid}.
While the individual noise levels do not offer sufficient protection, the aggregates provide rigorous privacy guarantees.
There exist different forms of performing the aggregation.
One popular approach %
is to use secure aggregation (SA)~\cite{bonawitz2017practical}, which adds significant computational overhead and requires tailored DP mechanisms that operate on integer values, \eg~\cite{kairouz2021distributed, agarwal2021skellam}.
Finally, prior work has shown that in FL, SA can be eluded~\cite{pasquini2021eluding}. 
This motivates defenses dedicated to protecting specifically against our \namenoformat attack.

\subsection{Specific Defenses against Trap Weights}
The following defenses can be applied to mitigate the success of our \namenoformat attack.
Since these defenses do not provide rigorous theoretical privacy guarantees but rather empirical protection, we recommend combining them with DP.

\myparagraph{Hardware-Based Protection.}
Using protocols that rely on Trusted Execution Environments (TEE), \eg~\cite{mo2021ppfl} prevents the \cp from performing active manipulations.
However, TEEs are prone to side-channel attacks~\cite{dessouky2020hybcache,jauernig2020trusted}.
Hence, there is a remaining risk for the privacy of \users' data.

\myparagraph{Local Averaging and Large Mini-Batches.} 
Our results in \Cref{tab:evaluation_batch_size_neurons} and \Cref{tab:effect_averaging} highlight that calculating gradients over large mini-batches and local averaging reduces the fraction of data points that can be perfectly reconstructed.
We, therefore, argue that \users should perform gradient calculation on large mini-batches of local data points and average gradients over multiple mini-batches before sending them to the \cp.
This is, however, only possible if \users have actual control on the local execution of the FL protocol and the execution is not inaccessibly encapsulated inside an application.

\myparagraph{Choice of the Activation Function.}
Our \namenoformat are designed to exploit properties of the ReLU activation function, namely the fact that it yields zero-gradients for inputs at some neurons.
Yielding zero-gradients is not unique to the ReLU activation. 
Other popular activation functions such as sigmoid and tanh have flat areas that also yield zero-gradients.
Hence, by adapting the initialization of our \namenoformat to these functions' properties, we could also achieve perfect extractability of individual data points with these functions.
However, using activation functions, such as leaky ReLU, which propagate information on every input through each neuron, can prevent individual extractability of training data points.

\myparagraph{Lossy Layers.}
Our evaluation in \Cref{sec:adv-initialize-evaluation} highlights that the application of layers that compress the input data or cause information loss, such as pooling or dropout reduce fidelity of the extracted data.
Therefore, relying on architectures that have aggressive compression and/or dropout reduces the leakage of individual \user data to the \cp.

Given that in FL, the \cp is in charge of instantiating the shared model (with its hyperparameters, such as the activation function), \users can \textit{only} rely on additional protection through the model itself \textit{if this \cp is trusted}.
An untrusted \cp, in contrast, has incentives to choose model architectures and hyperparameters that facilitate data extraction.

\section{Discussion and Future Directions}
\label{sec:discussion}

In this section, we first discuss the detectability of our adversarial initialization and integration of our attack in the training process of the shared model.
We then analyze the potential and capabilities of adversarial weight initialization for future privacy attacks.
Finally, we argue that dedicated privacy-protection should be implemented as a default option into FL protocols to prevent accidental or malicious privacy leakage.

\subsection{Detectability of Trap Weights}
\label{sub:detectability}

To detect the presence of our \namenoformat, the \users can apply one of the following two strategies:
(1) analyzing the weights of the shared model in \textit{one} or \textit{multiple} iterations over the FL protocol, or (2) analyzing the behavior of the shared model on their data.

\myparagraph{Analyzing Model Weights.}
Assuming the \user has access to the model only in \textbf{one iteration} of the protocol\footnote{Given that in practical deployments of FL, $\totalusers >> \selectusers$, an individual \user will be sampled for participation very rarely.}, they can run a detection method that aims at deliberately looking for characteristic elements of our \namenoformat, such as a normal distribution with high standard deviation, or the presence of higher absolute values for negative components than positive components in the first fully-connected layer's weight matrix.
\Cref{fig:weights_after_training} shows that even when initialized with a uniform distribution and relatively low deviation, model weights after several epochs of training resemble more a normal distribution and exhibit a larger standard deviation, making the former characteristic of our \namenoformat an unreliable attack detector.
When it comes to the magnitude of positive and negative components, \Cref{fig:values_s} shows that for $s$ close to one, the distribution of the model weights still resembles a standard normal.
However, the more $s$ deviates from one, the more the distribution of weights deviates from a standard normal distribution.
Yet, without knowledge of the prior training procedure and the other \users' data, we argue that a target \user can still not determine with certainty whether the received model weights are the result of the prior training or of a manipulation~\cite{Shumailov.2021Manipulating}.

Having access to the shared model over \textbf{multiple iterations} of the FL protocol \textit{additionally} enables \users to compare the received shared model's parameters to the parameters from previous FL iterations.
Therefore, the success of detection boils down to the following question: \textit{Can a local \user tell that a given set of model parameters came from legitimate updates of other local \users?} 
We argue that this is not possible, even for non-FL setups where the entire training procedure is transparent. 
The stochastic nature of training algorithms, combined with the non-determinism of modern hardware, makes it difficult to reproduce training runs~\cite{jia2021proofoflearning}. 
Because of this reproducibility error, an \attacker can assemble a mini-batch of natural data points that produce \textit{any} desired gradient update~\cite{Shumailov.2021Manipulating}. 
In other words, given two different sets of model weights, the \user cannot tell if the gradient descent step between these weights was a result of a legitimate optimization step. 
This is exacerbated in FL because the data of any given \user is invisible to other \users, further complicating the verification of gradient descent integrity.

\myparagraph{Analyzing Model Performance.}
In addition to analyzing the received model weights for detection of our attack, a \user can also evaluate the \emph{functionality} of the shared model.
We observe that, for the vast majority of classification tasks, the model's loss across training data points significantly reduces after just a few iterations. 
Thus, after the initial training iterations, FL \users would expect to encounter low loss values for their own examples. 
However, research has shown that, in particular for \users whose data stems from the tails of the data distribution, FL does not necessarily lead to an improvement of model accuracy on their data \cite{yu2021salvaging}.
Therefore, detection mechanisms that rely on analyzing the convergence of the model's accuracy over multiple FL iterations are also no reliable detectors of our manipulation.%

\subsection{Training Success and Model Performance}
\label{sub:training_success}
Increasing the utility of the shared model over the course of training in the FL protocol is important because the \cp is expected to provide a well-performing model after several training iterations.
Therefore, the \cp in our attack leverages two main points over the course of the protocol.
(1) Instead of aiming at reconstructing the \user data over all communications, it sends the adversarially (re-)initialized model out only at a few communication rounds.
In all other rounds, it sends out the actual shared model for training without adversarially re-initializing it.
(2) Instead of targeting all \users, the \cp only targets a subset of \users, and send out an adversarially initialized model to them, and the continuously trained shared model to all other \users.
The \cp can even combine both strategies by sending out an adversarially initialized model only to a subset of users in a few iterations.

\subsection{The Power of Weight Initialization}
In general, even outside of the FL context, our attack shows that controlling and manipulating the weights of neural networks opens a new attack surface against ML.
We argue that weight manipulation could be used to design further privacy attacks outside of the FL context.
Our adversarial initialization of convolutional and fully-connected model layers is able to transmit input data points to any subsequent layer in the model, practically modulating data perfectly over them.
Additionally by setting our \namenoformat, we can increase the leakage of individual training data points from model gradients.
Therefore, the \namenoformat basically create a simple if-else logic based on $>$ and $<$ relations between weighted inputs to model neurons.
Future work could investigate whether the weights could also be set in order to implement more complex logical structures and if-else cases depending on the input.
Based on these, it might be possible to craft hybrid attacks that first initialize the model weights and then use that to later extract information, for example, on membership of individual data points, or these data points' sensitive attributes.
Note that adversarially setting weights also does not need to be limited to initializing the model weights.
Instead, given an already initialized (and trained model), it might be possible to craft additional training data that leads to the weights taking the adversarial values that an \attacker wants.

\subsection{Using Dedicated Privacy Protection in FL}
\label{sub:privacy_confidentiality}
FL was originally designed as an alternative to centralized ML in which no large datasets would have to be moved from \users to a \cp in order to train an ML model on the joint data.
The approach does not only reduce communication costs but also spares the \cp from having to build up the infrastructure by outsourcing training and data storage costs to the \users.
Indeed, FL is more communication cost effective since the data itself is not shared directly. 

Attacks like our \namenoformat highlight, however, that the protocol does not guarantee protection for the individual \users' private training data.
This is not surprising since nothing in the design the FL protocol protects against leakage of private information.
Without dedicated privacy-protection, the \cp even has an upper hand over how much data a local model will leak, as demonstrated in our work.
However, FL is still often marketed as a data-minimizing technology.
Our work highlights that such marketing is misleading since in order to deploy FL as a privacy-technology, it is necessary to implement dedicated additional protection methods, such as the ones discussed in~\Cref{sec:defenses}.%
We argue that, to prevent malicious or accidental leakage in FL, these protection methods should be implemented in FL as a default when deploying the protocol to actual \users.
That is, vanilla federated learning does not provide privacy advantages for \users---unless it is combined with additional defense methods, such as DP learning.

\section{Conclusion}
In this work, we presented a new privacy attack against FL that is based on an active attacker who holds the ability to maliciously manipulate the shared model and its weights.
Our attack allows for perfect reconstruction of a significant portion of the \users' private training data.
Even for very high-dimensional complex datasets, such as ImageNet, we are able to perfectly extract roughly 50\% of the individual data points from mini-batches of sizes as large as 100.
The extraction is computationally highly efficient and even allows to perfectly extract individual training data points from data mini-batches containing all data points from one single class.

Our attack underscores the deficiency of the ``data never leaves the device'' approach to preserving privacy. For FL to have a chance of truly preserving privacy, it must incorporate appropriate mitigations against our attack. Those either have expensive overheads or are tailored for this specific attack (see Section~\ref{sec:defenses}).

\section*{Acknowledgments}
We would like to acknowledge our sponsors, who support our research with financial and in-kind contributions: Amazon, Apple, CIFAR
through the Canada CIFAR AI Chair, DARPA through the GARD project, Intel, Meta, NFRF through an Exploration
grant, NSERC through the COHESA Strategic Alliance, the Ontario Early Researcher Award, and the Sloan Foundation. Resources used in preparing this research were provided,
in part, by the Province of Ontario, the Government of Canada through CIFAR, and companies sponsoring the Vector
Institute.

\newpage
\bibliographystyle{plain}
\bibliography{library}

\appendices

\label{app:1}

\section{Extended Related Work on Passive Data Reconstruction Attacks}
\label{app:passive_reconstruction}

Table~\ref{tab:ComparisionAttacks} summarizes these data reconstruction attacks (described below) and compares them with our attack. 

\begin{table}[h!]
 \setlength{\tabcolsep}{2pt}
 \scriptsize
    \centering
    \begin{tabular}{l|cc|ccc|c|cc|c}
    \toprule
        \multirow{2}{*}{\textbf{Method}} & \multicolumn{2}{c|}{\textbf{\Attacker}} & \multicolumn{3}{c|}{\textbf{Rep.}} & \textbf{Label-} & \multicolumn{2}{c|}{\textbf{\textbf{B}}} & \textbf{Opt.-/}    \\
       & \textbf{U} & \textbf{S} & \textbf{C} & \textbf{U} & \textbf{ID} & \textbf{Free} & 1 & $\geq 1$ & \textbf{Train.-Free} \\
      &&&\multicolumn{3}{c|}{\tiny$\xrightarrow{stronger}$} &&\multicolumn{2}{c|}{\tiny$\xrightarrow{stronger}$} & \\
    \toprule
          DMU-GAN~\cite{Hitaj.2017Deep} &  \checkmark & &\checkmark & & &  & & \checkmark&\\
           mGAN-AI~\cite{Wang.2019Beyond} & & \checkmark & &\checkmark & & & & \checkmark &    \\
           DLG~\cite{Zhu.2020Deep} & & \checkmark & & & \checkmark & \checkmark &  \checkmark & &    \\
          iDLG~\cite{Zhao.2020iDLG} & & \checkmark& & & \checkmark & \checkmark & \checkmark &  &    \\
          GradInv~\cite{Yin.2021See} & & \checkmark & & & \checkmark & \checkmark &  & \checkmark &      \\
        \midrule
         \name~[Ours]& & \checkmark &   &  & \checkmark & \checkmark & & \checkmark & \checkmark\\
         \bottomrule
    \end{tabular}
    \caption{Comparison of data reconstruction attacks. KEY-- U: User, S: Server, C: Class, ID: Individual Data Points, B: Mini-batch size, Opt.: Optimization, Train.: Training, Rep.: Representative.}
    \label{tab:ComparisionAttacks}
\end{table}

\myparagraph{Class-wise Representation Reconstruction Attacks}.
Hitaj~\etal~\cite{Hitaj.2017Deep} were the first to propose a GAN-based data reconstruction attack, called DMU-GAN.
The attacker must know the dataset's classes, and the reconstructed data points are generic representations of class-wise properties rather than individual \user data points or classes. 
Wang~\etal~\cite{Wang.2019Beyond} suggested mGAN-AI, which extends DMU-GAN's reconstruction attack to per-\user class-wise representations, but still does not extract individual data points. 
Additionally, both methods require access to data from the same distribution as the \users' data. \cite{sun2020provable} observes that class-wise representations are embedded in model updates even without the need to reconstruct them using a GAN, and suggest defenses.

\begin{algorithm2e}[ht]
\DontPrintSemicolon
\SetKwComment{Comment}{{\scriptsize$\triangleright$\ }}{}
\caption{Optimization-Based Data Rec.~\cite{Zhu.2020Deep}. }\label{algo:opt_based_reconstruction}
        \KwIn{Gradients, $\Grad_i\att{t}$, received from victim \user $u_i$ at iteration $t$, Shared model $\f\att{t}(\cdot)$ at iteration $t$.} 

        \KwOut{Reconstructed training data, ($\mathbf{x}_i^*$, $y_i^*$)}
\BlankLine
\begin{minipage}{1.05\hsize}
    \begin{algorithmic}[1]
    \STATE $(\hat{\mathbf{x}}^{\att{1}},\hat{y}^{\att{1}}) \leftarrow (\mathcal{N}(0,1),\mathcal{N}(0,1))$ \Comment*[r]{{\scriptsize Initialize}} 

    \FOR{$\hat{t} \in [1,\hat{T}]$}
        \STATE $\hat{\Grad}\att{\hat{t}} = \nabla_{\WAll}\Loss(\f\att{t}(\hat{\mathbf{x}}^{\hat{t}}),\hat{y}^{\hat{t}})$ \Comment*[r]{{\scriptsize Dummy gradients}}
        \STATE $D^{\hat{\att{t}}}=\| \Grad_i\att{t} - \hat{\Grad}\att{\hat{t}} \|^2$ \Comment*[r]{{\scriptsize Dummy vs \user}} 
        \STATE $\hat{\mathbf{x}}^{\att{{\hat{t}+1}}} \leftarrow \hat{\mathbf{x}}^{\att{{\hat{t}}}}-\alpha \nabla_{\hat{\mathbf{x}}^{\att{{\hat{t}}}}}D^{\hat{\att{t}}},$ 
        \STATE $\hat{y}^{\att{{\hat{t}+1}}} \leftarrow \hat{y}^{\att{{\hat{t}}}}-\alpha \nabla_{\hat{y}^{\att{{\hat{t}}}}}D^{\hat{\att{t}}}$

        \ENDFOR    
    \STATE $(\mathbf{x}^*_i$, $y^*_i) \leftarrow (\hat{\mathbf{x}}^{\att{\hat{T}+1}}$, $\hat{y}^{\att{\hat{T}+1}})$ 
\end{algorithmic}
\label{algo:SOA_opt_based_data_reconstruction}
\end{minipage}
\end{algorithm2e}

\begin{figure*}[t]
	\centering
	\begin{subfigure}[b]{0.3\textwidth}
		\centering
		\includegraphics[width=\linewidth]{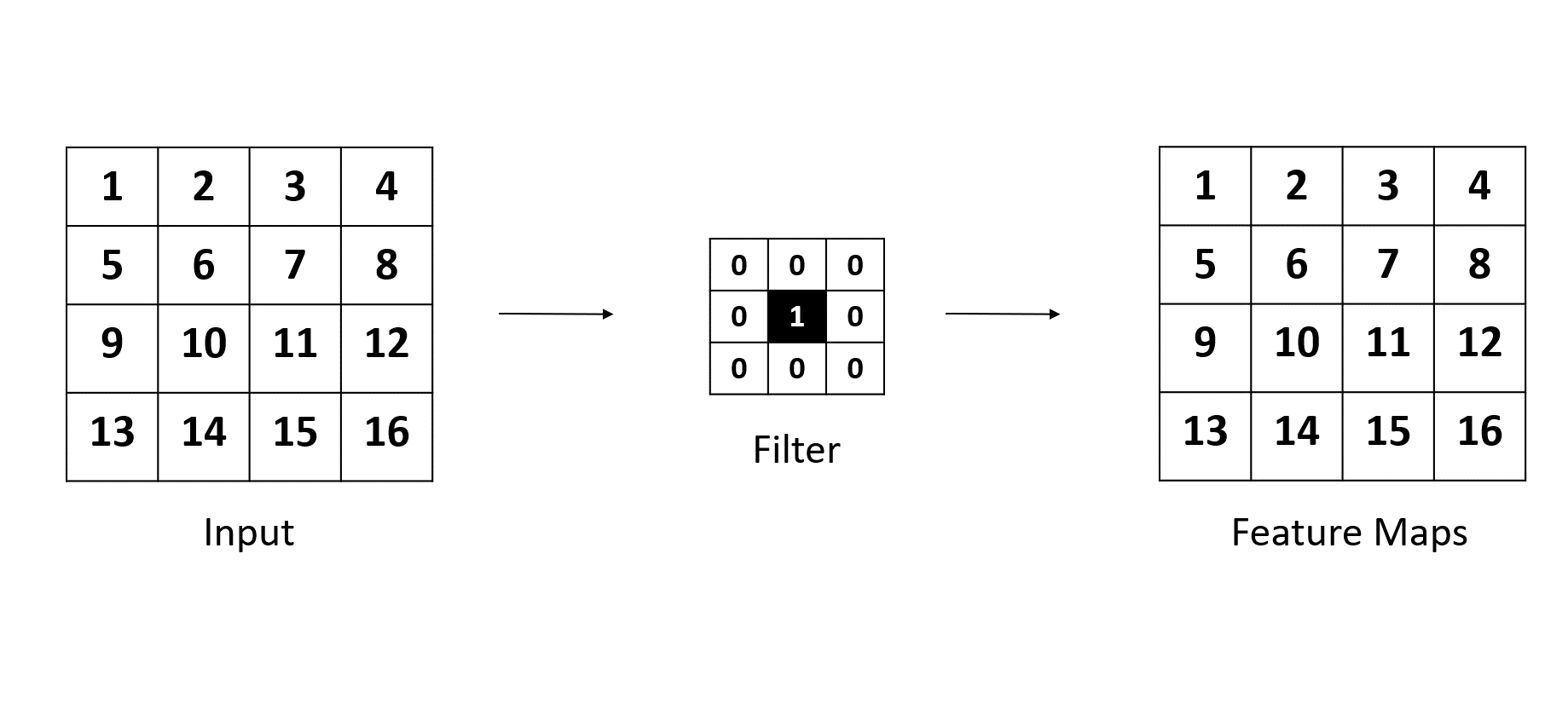}
		\caption{2D Example.}
		\label{subfig:2D-conv-preserve}
	\end{subfigure}  
	\hspace{3cm}
	\begin{subfigure}[b]{0.3\textwidth}
		\centering
		\includegraphics[width=\linewidth]{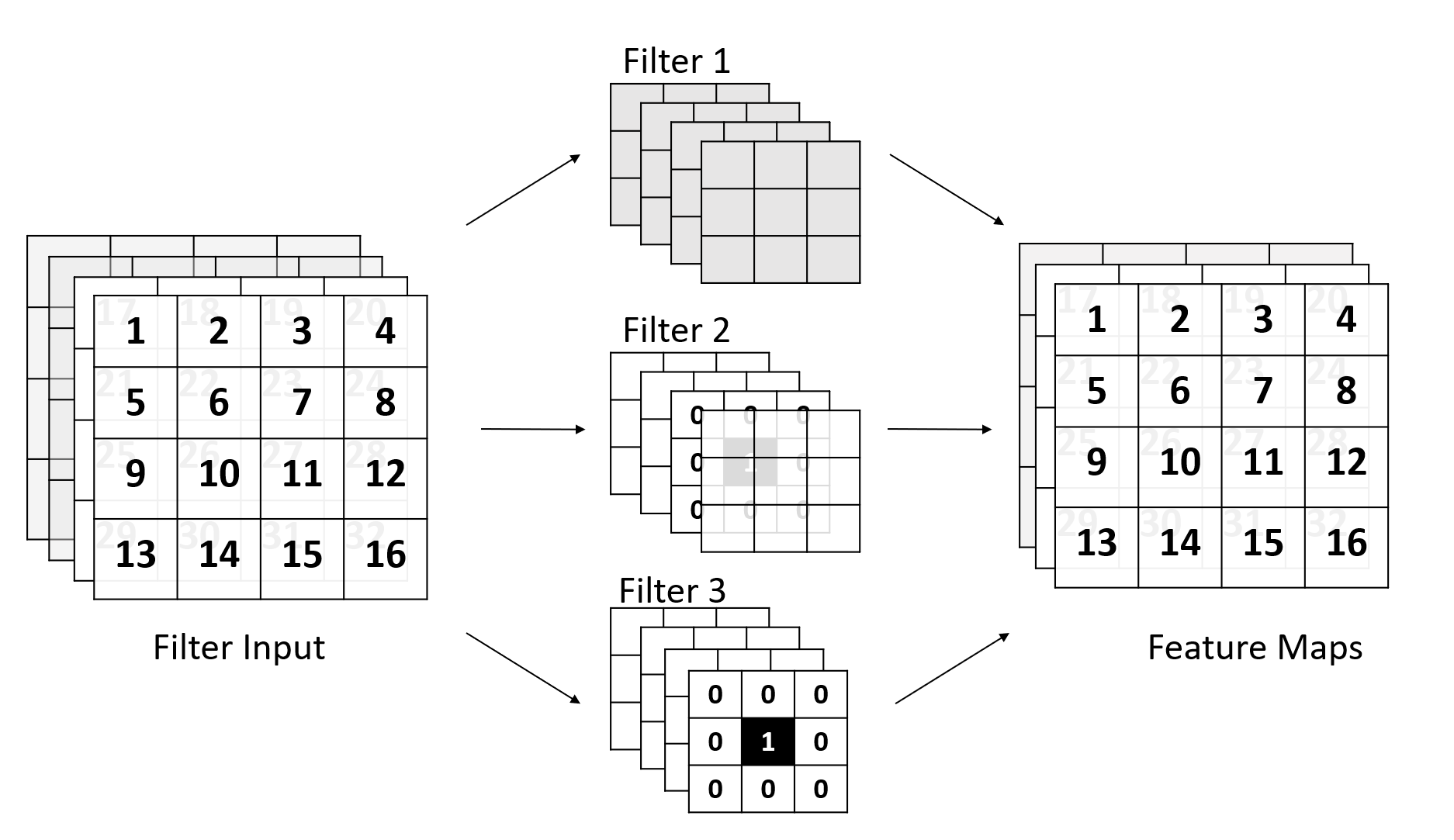}
		\caption{3D Example.}
		\label{subfig:3D-conv-preserve}
	\end{subfigure}   
	\caption{\textbf{Size-Preserving Adversarially Initialized Convolutional Filters.} Adversarially initialized convolutional filters that transmit their input to the next layer. The numbers indicated in the input and feature map represent the features, the numbers in the filter represent the weight initialization. Grey layers indicate random weight values while white layers indicate zero weights. }
	\label{fig:conv}
\end{figure*}
\setlength{\textfloatsep}{0.1cm}

\section{Generalization of Data Extraction Attack to Convolutional Neural Networks}
\label{sec:adv_forwarding}

So far, both the passive and active attacks we described are tailored to extracting data from the gradients computed to update a fully-connected layer. However, modern neural network architectures often rely on convolutional layers to model image and text data alike.
It is difficult to directly apply our attack strategy to these convolutional layers  because they rely on the weight sharing principle: to decrease the effective number of parameters that need to be trained, the same weight values are applied to multiple locations of the image to extract patterns regardless of their location in the image. 
In this section, we thus propose a second instantiation of our adversarial weight initialization  strategy that  generalizes extraction attacks to convolutional neural networks (CNNs).

Our solution reduces networks with convolutional layers to the setting we previously considered with fully-connected neural networks. To do so, we observe that a CNN typically composes a few convolutional layers with fully-connected layers. We thus initialize the weights of the convolutional layers such that they transmit the model input \textit{unaltered} up to the fully-connected layers of the model architecture.%

There are two important requirements for our approach to transmitting, or forwarding, model inputs through convolutional layers.
The first  is to make sure that no feature of the input data is lost. This requires having at least as many parameters at every convolutional layer as the number of input features.
The second  is to make sure that different features do not get overlaid. We explain how to ensure this next. 

\subsection{Preserving Input-Size}
\label{app:size_preserving_filters}
\myparagraph{Two Dimensional Input.}
In general, preserving input size over a convolutional layer can be achieved through an adequate combination of padding, stride, and filter sizes.
Specifically, we use stride \emph{one} and an adequate zero-padding to preserve the size of the layer input. 
In order to transmit the input features, we create a filter with uneven dimensions $(w,h)$, where $w=h$, and we initialize it with \zero~everywhere apart from the element in the middle which we set to \one.
For a two dimensional input (\eg a grey-scale image), the described filter perfectly transmits the information to the next layer and creates a feature map that exactly replicates the input. 
See \Cref{subfig:2D-conv-preserve} for this adversarially initialized filter.

\myparagraph{Three Dimensional Input.}
Some input data to CNNs is distributed over several input channels, such as color images, that consist of three channels.
At every layer, we, therefore need three adversarially initialized convolutional filters to "transmit a copy" of the input channels.
A standard architecture can have many more filters per layer, which can, in the case of our attack be randomly initialized since they will be ignored by the \attacker.
Assume now that the original input features at the current layer $\li$ are distributed over $a_{\li}$ of the total $b_{l_{i-1}}$many feature maps.
For example, in the first model layer, $a_{\li}= b_{l_{i-1}}$ corresponds to the number of color channels required to encode the image.
In subsequent layers, the remaining $b_{l_{i-1}}-a_{\li}$ many feature maps contain random noise, introduced by random filters that do not transmit the input features (\eg Filter 1 in \Cref{subfig:3D-conv-preserve}).
We denote the indices of the feature maps where the input features are located by $\vec{\alpha_{\li}}$.
We then need $a_{\li}$ many filters, initialized as described above to transmit the information to the next layer.
The filters differ from each other only by the placement of the matrix that contains the \one~element.
This placement must correspond to different indices in $\vec{\alpha_{\li}}$.
See \Cref{subfig:3D-conv-preserve} for a visualization of this setting.
Note that the placement of the feature-transmitting filters at layer $\li$ will determine the indices $\vec{\alpha_{l_{i+1}}}$ of the feature maps that are input to the next layer.

In the last convolutional layer before the fully-connected layer that we want to transmit the input to, the filters containing noise should be initialized such that they yield negative input to the ReLU function. 
Thereby, the output of the last convolutional layer becomes zero everywhere apart from the feature maps produced by the filters transmitting input data features.
The flattened output then serves as input to the fully-connected layer, and reconstruction can be conducted as described in \Cref{sec:adv-init-first-fc}.

\subsection{Reducing Input-Size}
\label{sub:cnns_reduceSize}

\begin{figure*}[t]
	\centering
	\begin{subfigure}[b]{0.45\textwidth}
		\centering
        \includegraphics[width=0.45\linewidth, angle=270]{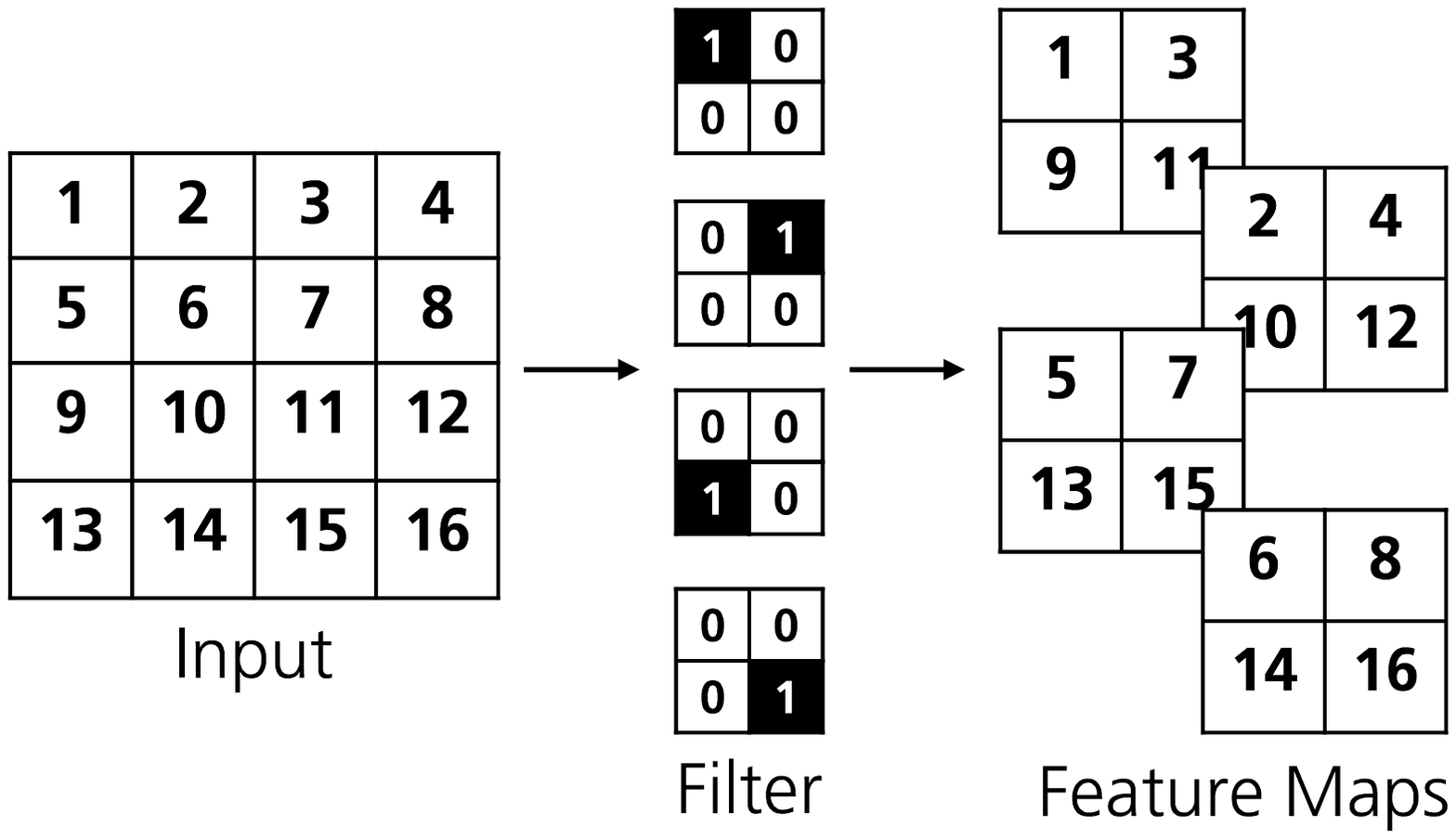}
		\caption{2D Example.}
		\label{subfig:2D-conv-reduce}
	\end{subfigure}  
	\hspace{1cm}
	\begin{subfigure}[b]{0.45\textwidth}
		\centering
		\includegraphics[width=0.45\linewidth, angle=270]{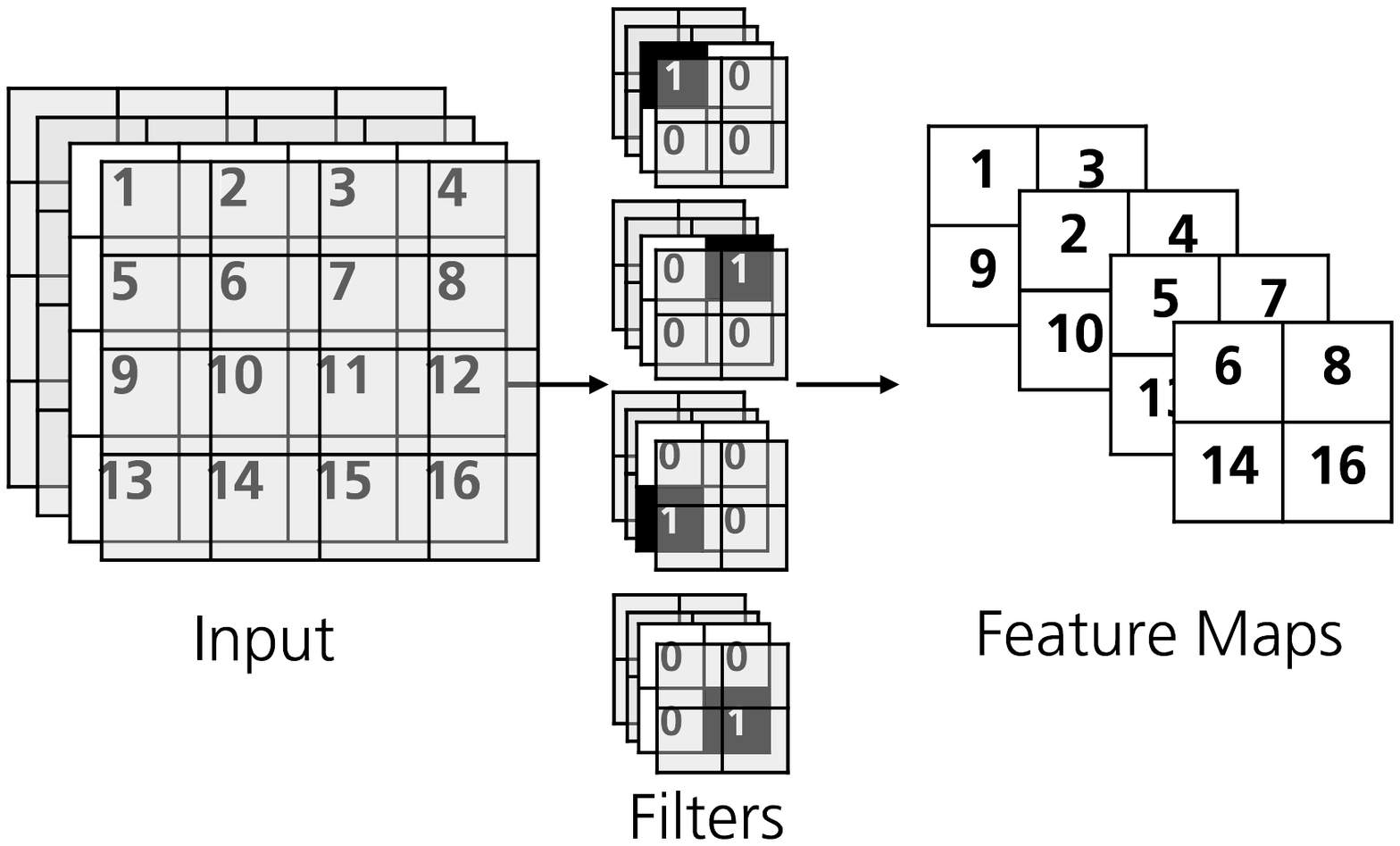}
		\caption{3D Example.}
		\label{subfig:3D-conv-reduce}
	\end{subfigure}
	\caption[Size-Reducing Adversarially Initialized Convolutional Filters]{\textbf{Size-Reducing Adversarially Initialized Convolutional Filters.} Adversarially initialized convolutional filters that transmit their input to the next layer. The numbers indicated in the input and feature map represent the features, the numbers in the filter represent the weight initialization. Grey layers indicate random weight values while white fields indicate zero weights. Feature maps in the 3D example that only contain noise are suppressed for improved visualization.}
	\label{fig:conv-reduce}
\end{figure*}

\myparagraph{Two Dimensional Input.}
The reduction of size in convolutional layers can be achieved by increasing the stride.
However, thereby, the number of features in the next feature map is reduced such that this feature map cannot accommodate all features from the previous layer.
To overcome this, we propose distributing the features of one input feature map over several feature maps in the following layer.
\Cref{subfig:2D-conv-reduce} depicts this approach for two-dimensional inputs.
Note that the stride is set to the dimensions of the convolutional filters $(w,h)$ to prevent features from overlapping in the following layer.
Additionally, to transmit all the features, the dimensions of the filters $w$ and $h$ $(w=h)$ need to be integer dividers of the previous feature map's dimensions.
Finally, in total, for each layer that reduces the size of the input by a factor $\frac{1}{w}$, we require $w^2$ many filters to transmit every feature from one input feature map.
Hence, assuming that at layer $l_i$ the original features are distributed over $a_{l_i}$ many input feature maps, we require $a_{l_i}\cdot w^2$ many filters to transmit all original input features.

\myparagraph{Three Dimensional Input.}
The same approach as for the two dimensional input can be extended to the case with three input dimensions.
The approach is visualized in \Cref{subfig:3D-conv-reduce}.
For improved visualization, we do not present the feature maps in the layer's output which contain only noise.
Again, in both the two and three dimensional case, in the last convolutional layer before flattening, the noise filters should produce negative input to the ReLU function.
This enables only extracting the original input features and no noise from the following fully-connected layer.

\subsection{Reducing Detectability}
\label{sub:cnns_detectability}
In principle, our adversarial weight initialization for CNNs only requires the number of filters per layer that actually transmit the features.
However, using only a small number of filters, \eg \one~as in the case of the size-preserving adversarial convolutional filters, leads to models architectures that deviate strongly from standard architectures.
Therefore, we propose using a standard number of convolutional filters in every layer and initializing the filters that are not used to transmit features at random.
Additionally, to prevent the simple detection strategy which relies on probing after every convolutional layer whether its input is equal to its output, one can replace the ones in the adversarially initialized convolutional filters by other positive constants.
Data extraction at the fully-connected layer then yields data points where features of the original input data are scaled by (multiple different) factors. 
By applying the inverse of the factors encoded in the model weights this scaling can then be reverted.
As a consequence, the rescaled extracted data points still perfectly correspond to the input data.

\section{Additional Material}
\label{app:additional_material}

The following table describes the model architectures both for the FC-NNs and CNNs use throughout the paper. Note that the our method could also be applied to much larger CNNs with more layers: in fact, as long as each layer contains as many parameters as the data holds input features, our approach is applicable.

\begin{table}[!h]
    \centering
    \scriptsize
    \begin{tabular}{cc}
    \toprule
    FC-NN Architecture & VGG-inspired CNN Architecture \\
    \midrule 
        Dense(n=1000, act=relu)   &  Conv(f=128, k=(3,3), s=1, p=same, act=relu)\\
        Dense(n=3000, act=relu) &  Conv(f=256, k=(3,3), s=1, p=same, act=relu)\\
        Dense(n=3000, act=relu) &   Conv(f=512, k=(3,3), s=1, p=same, act=relu)\\
        Dense(n=2000, act=relu) &  Flatten\\
        Dense(n=1000, act=relu) &  Dense(n=1000, act=relu)\\
        Dense(n=\#classes, act=None)  &   Dense(n=\#classes, act=None)\\
    \bottomrule
    \end{tabular}
    \caption{Architectures of models used in the experiments on image data. f: number of filters, k: kernel size, s: stride, p: padding act: activation function, n: number of neurons.}
    \label{tab:architectures}
\end{table}

\begin{table}[!h]
    \centering
    \begin{tabular}{c}
    \toprule
    IMDB-Model Architecture                  \\
    \midrule 
        Embedding(feat=10000, dim=250)    \\
        Dense(n=1000, act=relu) )       \\
        Dense(n=1, act=None)    \\
    \bottomrule
    \end{tabular}
    \caption{Architecture of models used in the experiments on the IMDB dataset. feat: vocabulary size, dim: embedding size, act: activation function, n: number of neurons.}
    \label{tab:architecture_text}
\end{table}

\section{Additional Experimental Results}
\label{app:additional_results}
This section presents additional experimental results.

\subsection{Passive Extraction}

\begin{figure}[h]
\centering
\includegraphics[width=0.5\textwidth, trim={0cm 16.8cm 0cm 2cm},clip]{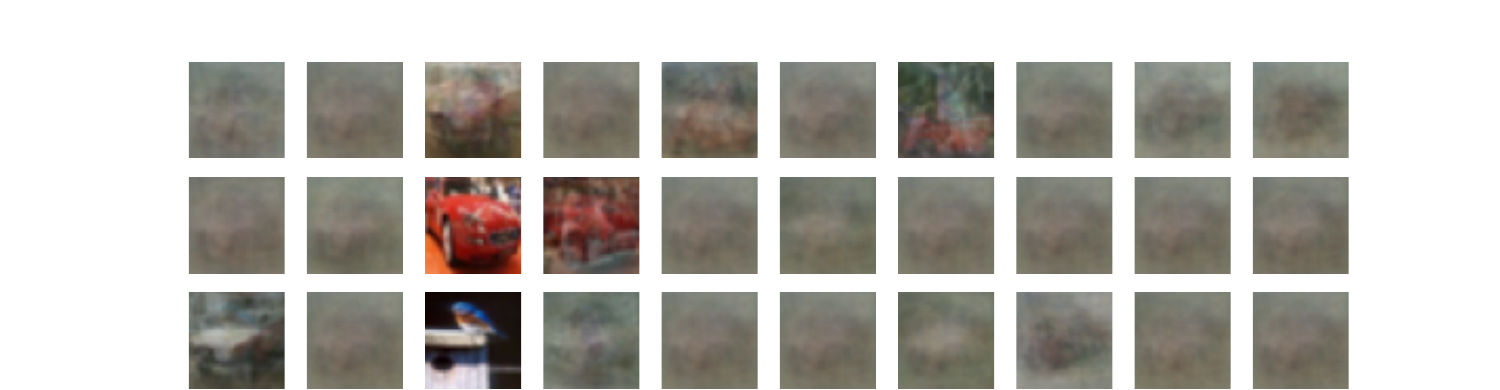}
\caption{\textbf{Baseline: Passive Attack.} Data from the CIFAR10 dataset, extracted from the gradients of the first $30$ weight rows at the first fully-connected layer of a randomly initialized FC-NN with architecture from \Cref{tab:architectures}.}
\label{fig:100-gradients-cifar10}
\end{figure}

\Cref{fig:100-gradients-cifar10} shows extraction from a randomly initialized FC-NN with architecture presented in \Cref{tab:architectures}.

\subsection{Trap Weights and Active Extraction}
\label{app:additional_insights_trap_weights}
\begin{figure}[t]
\centering
\begin{subfigure}[b]{0.43\textwidth}
\centering
\includegraphics[width=\linewidth, trim={0cm 17.7cm 0cm 0cm},clip]{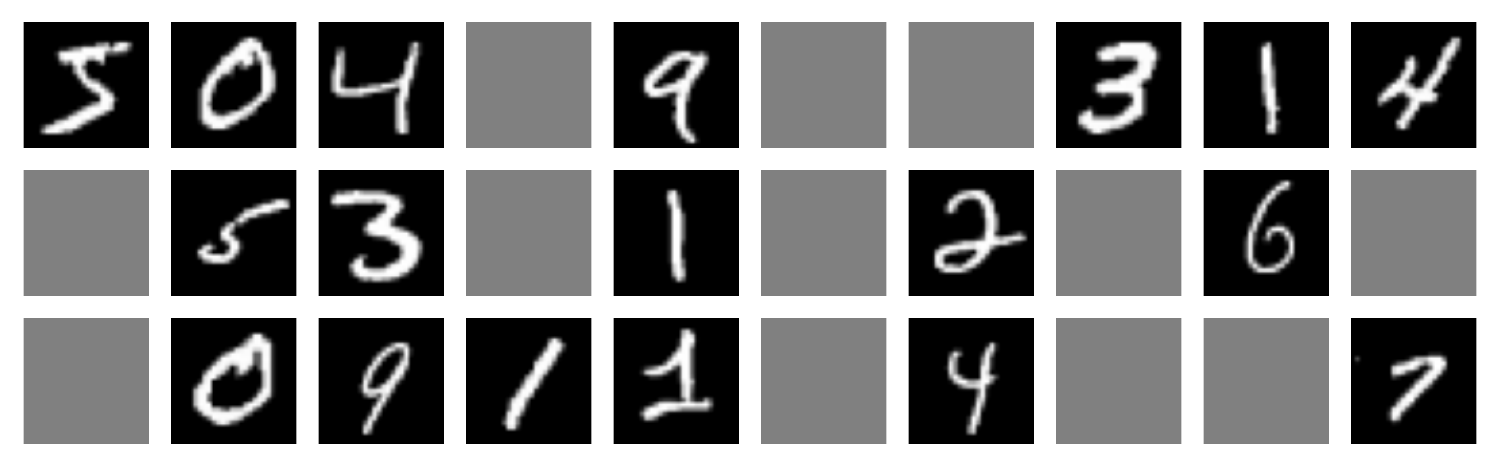}
\caption{Reconstructed data points.}
\label{subfig:}
\end{subfigure}     
\begin{subfigure}[b]{0.43\textwidth}
\centering
\includegraphics[width=\linewidth, trim={0cm 17.7cm 0cm 0cm},clip]{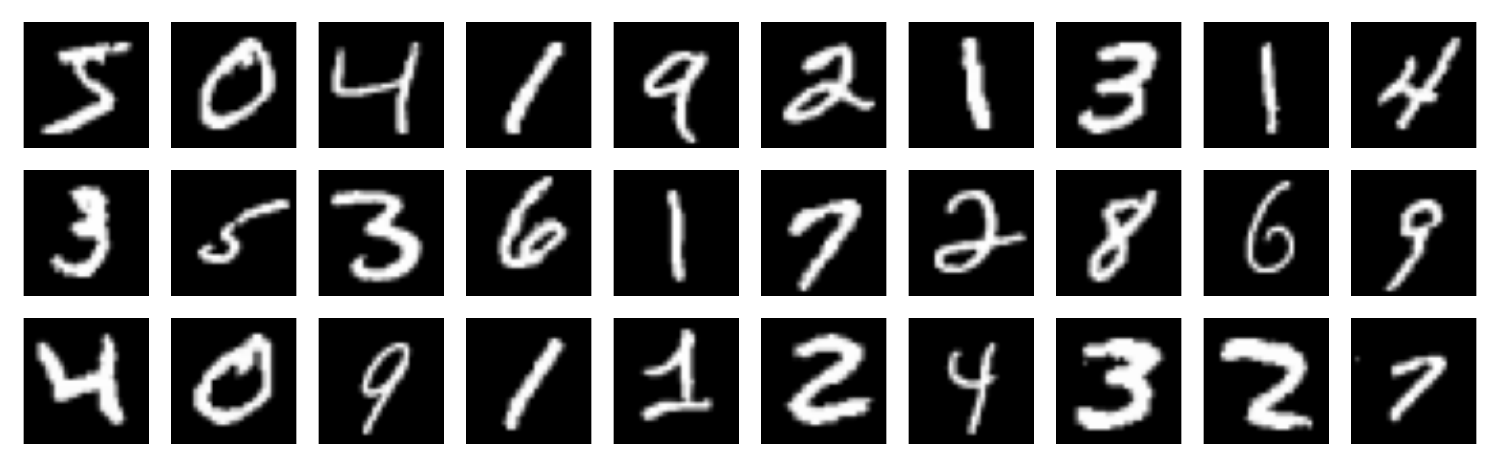}
\caption{Original data points.}
\label{subfig:}
\end{subfigure}   
\caption{MNIST. Reconstruction success of our adversarial initialization: first 30 images from a mini-batch of 100 data points, extracted at the first fully-connected layer of the FC-NN from \Cref{tab:architectures}. Gray images indicate that the corresponding original data point could not be extracted individually from the model gradients.}
\label{fig:mnist_first_layer}
\end{figure}

\begin{figure*}[b]
\centering
\begin{subfigure}[b]{0.43\textwidth}
\centering
\includegraphics[width=\linewidth]{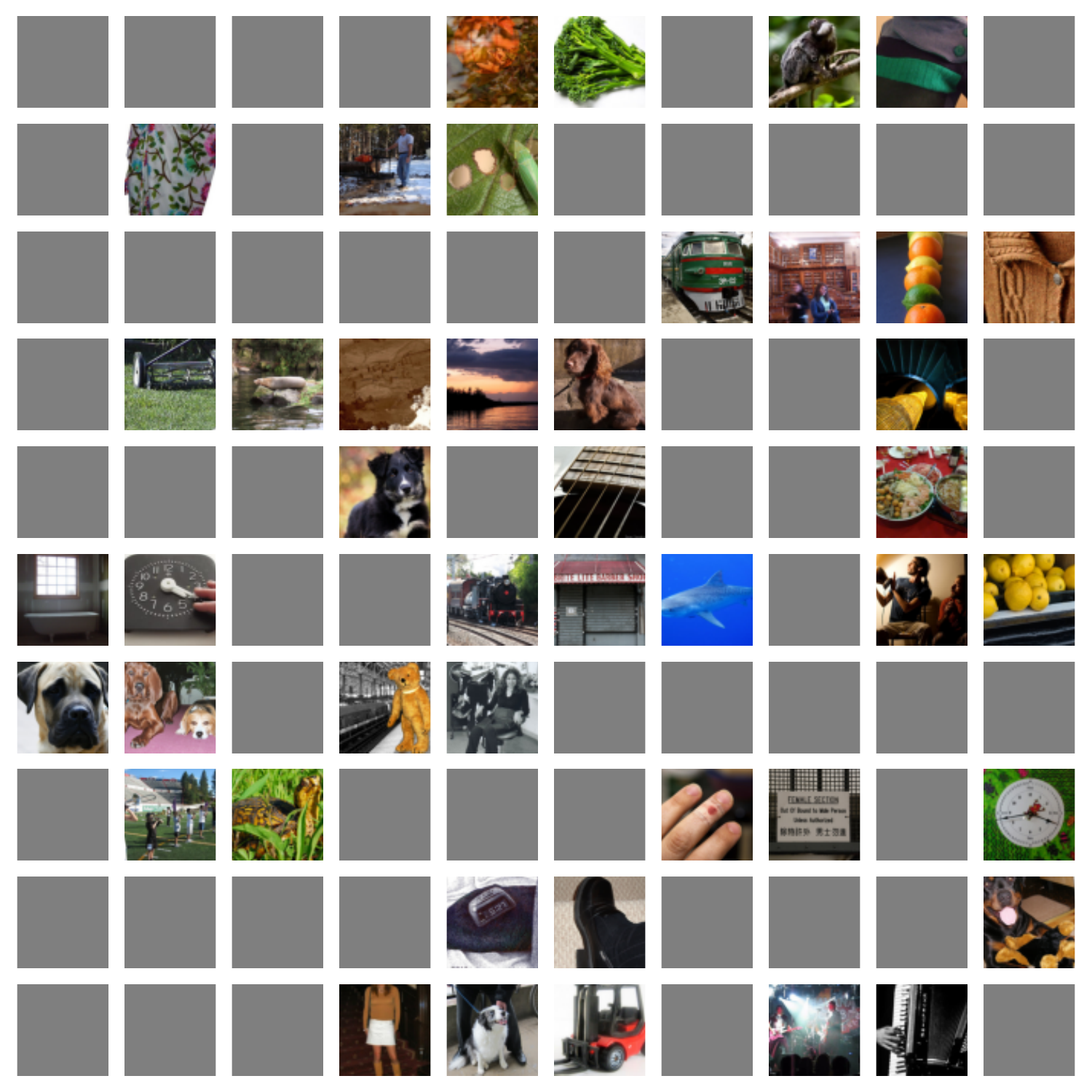}
\caption{Reconstructed data points.}
\end{subfigure}     
\hfill
\begin{subfigure}[b]{0.43\textwidth}
\centering
\includegraphics[width=\linewidth]{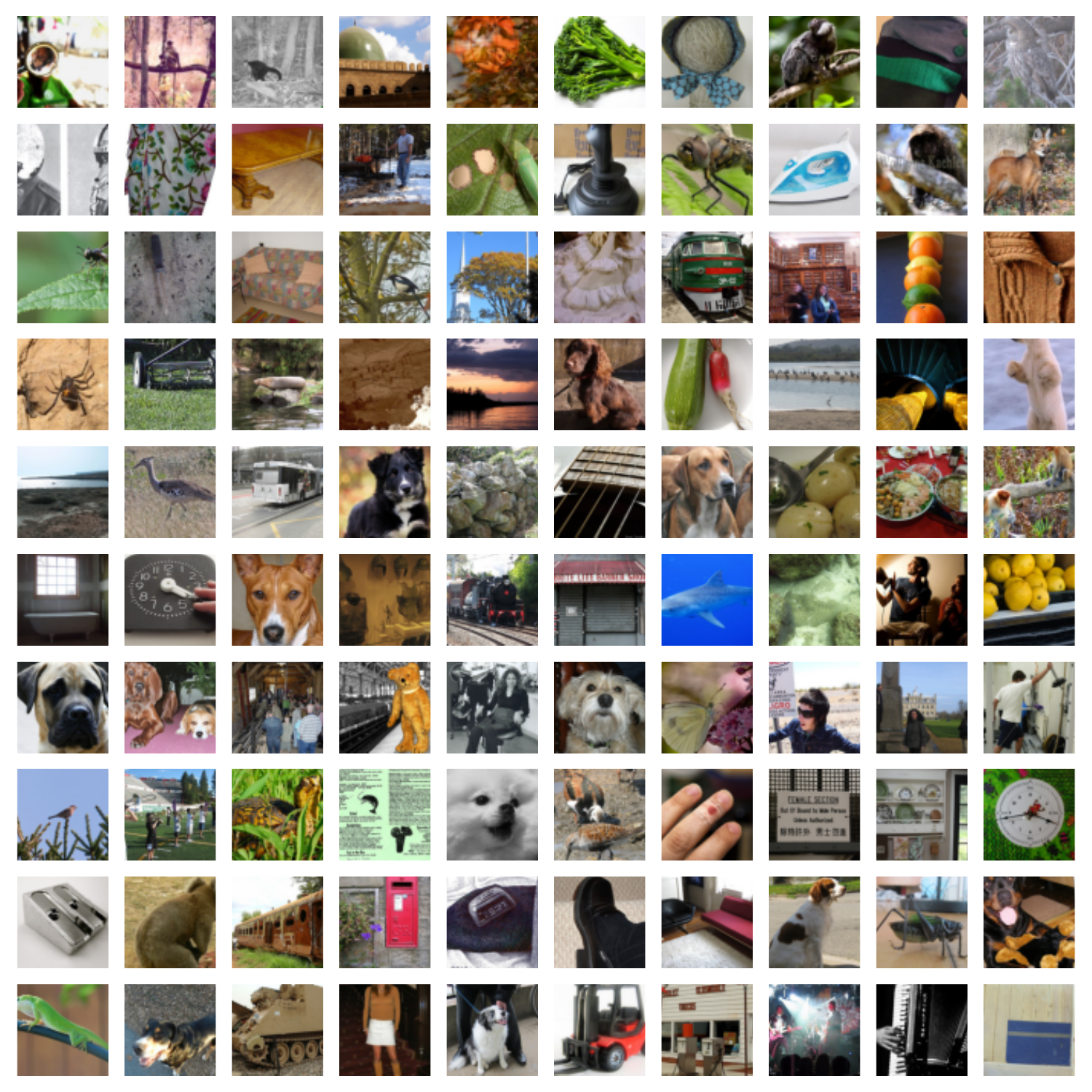}
\caption{Original data points.}
\end{subfigure}   
\caption{ImageNet. Reconstruction success of our adversarial initialization: all reconstructed data points from a mini-batch of 100 data points, extracted at the first fully-connected layer of the CNN from \Cref{tab:architectures}. Gray images indicate that the corresponding original data point could not be extracted individually from the model gradients..}
\label{fig:imagenet_first_layer}
\end{figure*}

Figures \ref{fig:mnist_first_layer} and \ref{fig:imagenet_first_layer} depict the extracted data points for MNIST and ImageNet, respectively.

\begin{figure}[tbh]
\centering
\includegraphics[width=0.8\linewidth]{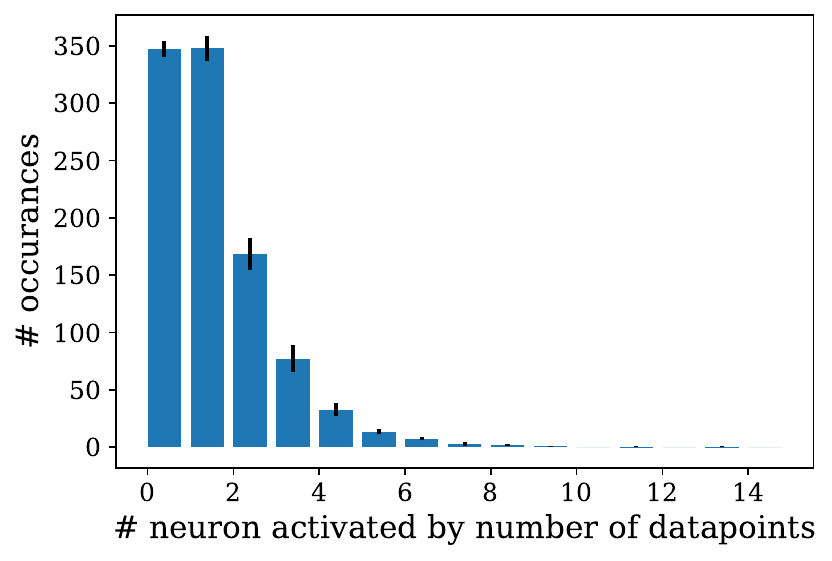}
\caption{ImageNet. To study partial extractability, we analyze by how many data points each neuron gets activated. Results are averaged over five different random and trap weight model initializations.}
\label{fig:extractability_q3}
\end{figure}

We, furthermore, study partial extractablity, \ie, the case when a data point is not individually extractable, but still leaks meaningful private information about a training data point.
Partial leakage occurs when an extracted gradient represents the overlay of only a few data points.
In this case, the individual signal of each data point is still distinguishable, see for example the first data point in the third row of \Cref{fig:100-gradients-cifar10}.
We plot in \Cref{fig:extractability_q3} by how many data point each of the neurons with our trap weight initialization gets activated.
This corresponds to the number of data points that will be present in the overlay of the respective gradients.
We can see that nearly as many neurons get activated by two data points as by one data point (\ie perfect extractability).
In general, with our trap weights, neurons get activated by small numbers of data points.
This indicates that the \cp can still extract meaningful partial information on many data points, also if these are not perfectly extractable.
Results an average over five runs with different trap weight initializations for a mini-batch of 100 data points from the ImageNet dataset.

\begin{figure*}[thb]
\centering
\begin{subfigure}[b]{0.46\textwidth}
\centering
\includegraphics[width=\linewidth]{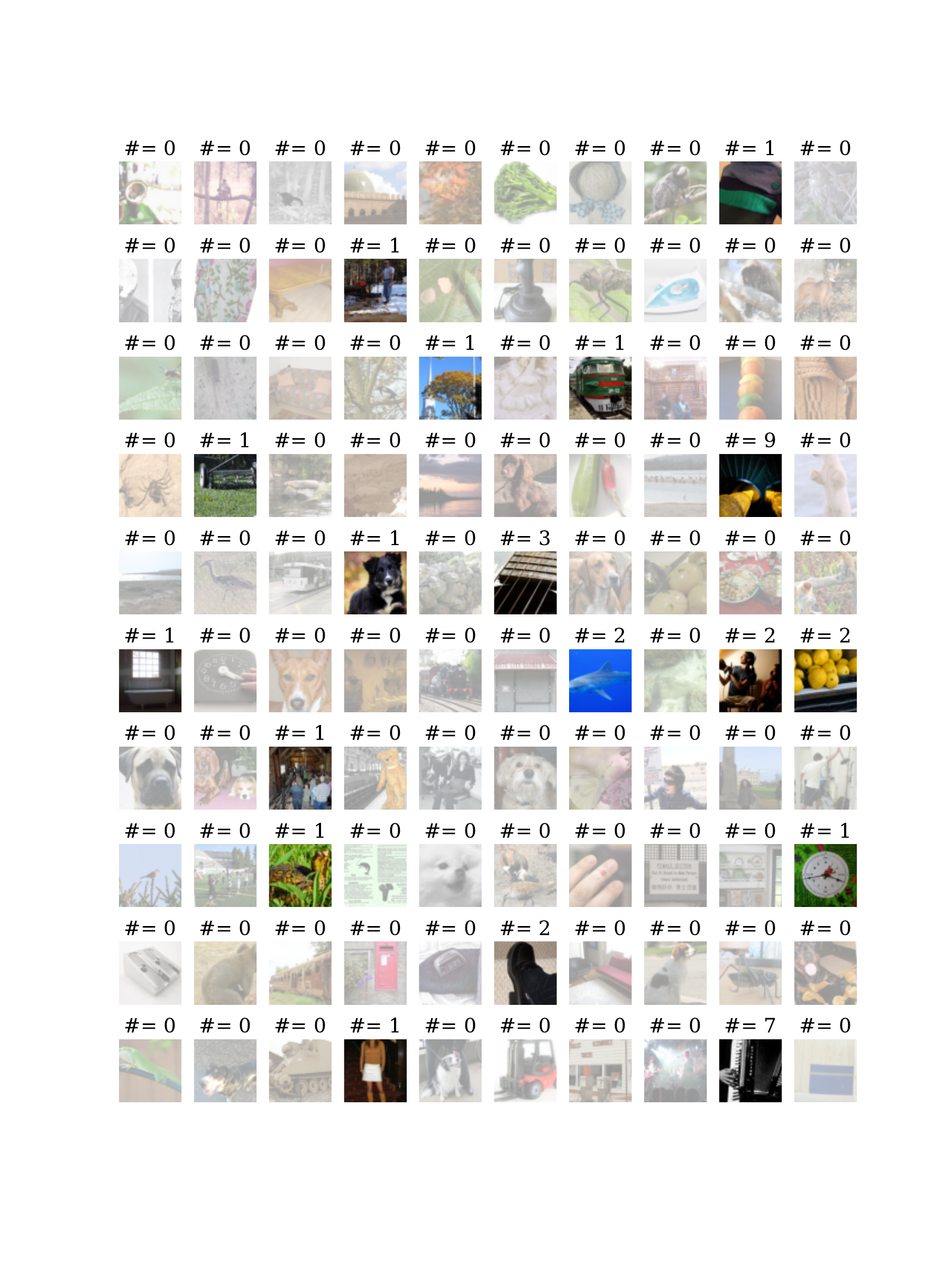}
\caption{Randomly initialized model weights.}
\end{subfigure}     
\hfill
\begin{subfigure}[b]{0.46\textwidth}
\centering
\includegraphics[width=\linewidth]{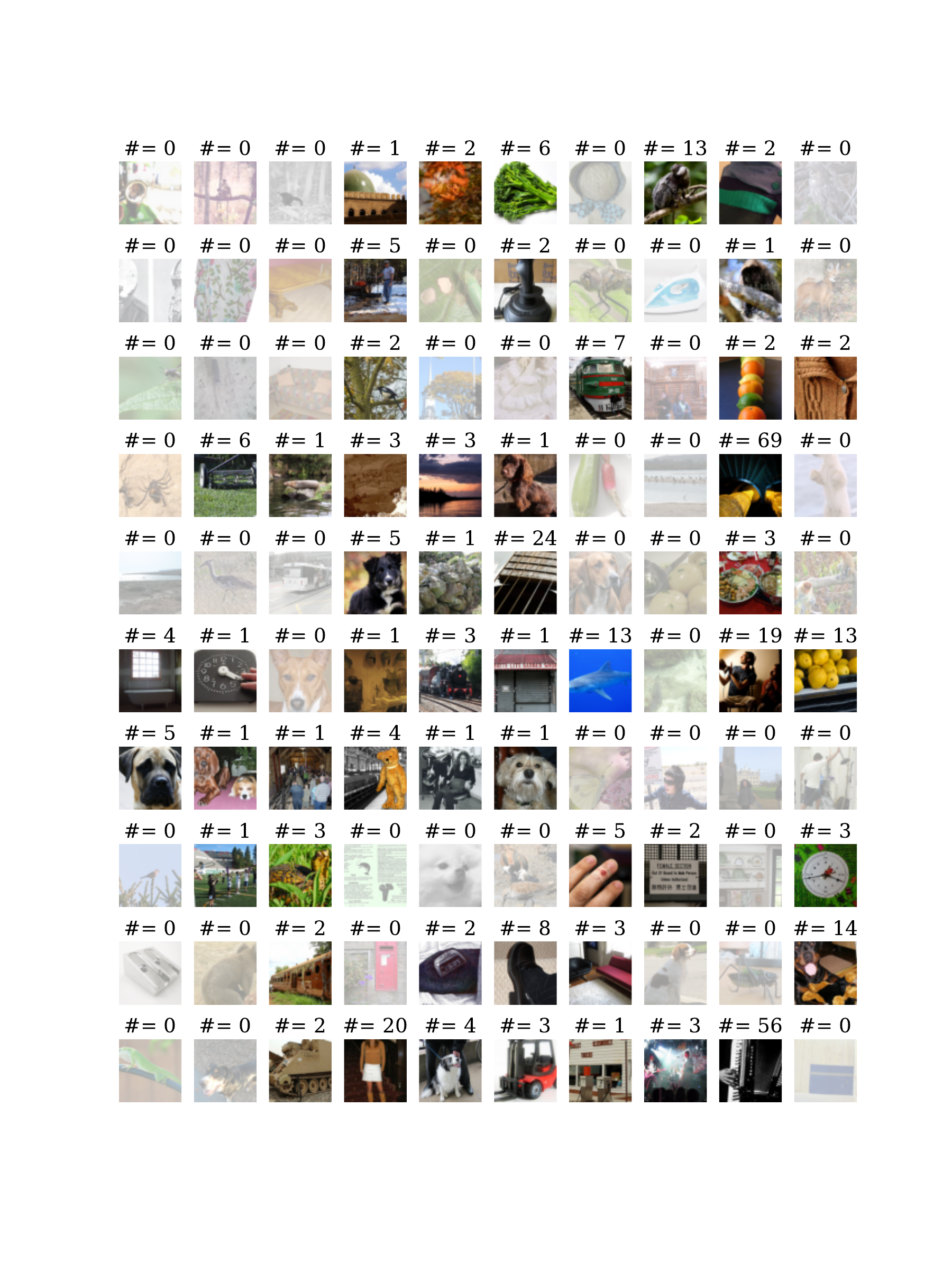}
\caption{Trap weights.}
\end{subfigure}   
\caption{ImageNet. Number of individual occurrences in the rescaled gradients over a mini-batch of 100 data points, extracted at the first fully-connected layer of the CNN from \Cref{tab:architectures}. To provide insights into what data points could not be individually extracted, we plot data points with zero occurrences with low saturation.}
\label{fig:extractability_q2}
\end{figure*}

To provide additional insights on data points that can and cannot be individually extracted, in \Cref{fig:extractability_q2} which data points can and cannot be extracted.
For the individually extractable data points, we, furthermore, depict how often each of them is individually extractable, \ie for how many neurons this data point is the only one activating it. 
We see that our trap weights first amplify natural leakage \ie, data points that are extractable from random weights are usually also extractable with our trap weight and our trap weights make other data points extractable.
Second, our trap weights yield redundancy, \ie data points are extractable multiple times from different weight rows' gradients.

We depict extraction success for local averaging over multiple mini-batches in \Cref{tab:effect_averaging}.

\begin{table}[tb]
\centering
\scriptsize
\begin{tabular}{crrr}
\toprule
  \textbf{B, Num}      & A & P & R \\
\midrule
 (20,1)  &                .496 &     .486 &  .950 \\
 (20,5)  &                .787 &     .213 &  .572 \\
 (20,10) &                .851 &     .157 &  .412 \\
 (20,20) &                .898 &     .116 &  .251\\
 (50,1)  &                .687 &     .307 &  .790 \\
 (50,5)  &                .901 &     .107 &  .230 \\
 (50,10) &                .928 &     .080 &  .138 \\
 (50,20) &                .953 &     .053 &  .067 \\
 (100,1)  &                .800 &     .200 &  .562 \\
 (100,5)  &                .936 &     .066 &  .116 \\
 (100,10) &                 .966 &     .046 &  .054\\
 (100,20) &                .982 &     .028 &  .020 \\
\bottomrule
\end{tabular}
\caption{\textbf{Effect of Mini-Batch Averaging.} Success of our adversarial weight initialization on MNIST under averaging over multiple mini-batches on the same model parameters. The number of mini-batches is denoted by \textbf{Num} and their respective size by \textbf{B}. The results depict the percentage of \act (A), \precision (P), and \recall (R) for extracting from 1000 neurons at the first layer of the FC-NN depicted in~\Cref{tab:architectures}. All numbers are averaged over 10 runs with different adversarial initializations.}
\label{tab:effect_averaging}
\end{table}

We also study the non-IID setup where \users hold data from a single class, different from other \users in the protocol.
We present the \recall and \precision per class on the CIFAR10 dataset in \Cref{tab:effect_nonIID}.

\begin{table}[h]
\centering
\scriptsize
\begin{tabular}{ccccc}
\toprule
  \textbf{Class}      & P (Passive) & P (Active) & R (Passive) & R (Active)\\
\midrule
 0  &  .064  &\textbf{.570} & .185 & \textbf{.352} \\
 1  &    .041 & \textbf{.276} & .208 &\textbf{.560} \\
 2 &    .056  &\textbf{.480} & .195 &\textbf{.384} \\
 3 &     .044  &\textbf{.318} & .208 &\textbf{.489}\\
 4  &    .056  & \textbf{.516} & .225 &\textbf{.426} \\
 5  &     .045  &\textbf{.356} & .238 &\textbf{.534} \\
 6 &      .049  & \textbf{.358} & .209 &\textbf{.442} \\
 7 &     .051  &\textbf{.367} & .205 &\textbf{.515} \\
 8  &      .055  & \textbf{.536} & .209 &\textbf{.386} \\
 9  &      .043  & \textbf{.395} &  .240 &\textbf{.559} \\
 
\bottomrule
\end{tabular}
\caption{\textbf{Non-IID Extraction on CIFAR10.} Success of our adversarial weight initialization (active) versus non-manipulated  model weights (passive) on CIFAR10 in a non-IID setup where each user only holds data from a single class, different from all other users. The results depict the \precision (P) and \recall (R) for extracting from 1000 neurons at the first layer of the FC-NN depicted in~\Cref{tab:architectures}. All numbers are averaged over 10 runs with different (adversarial) initializations.
While in both passive and active extraction, extraction success between the classes differs, our adversarial weight initialization significantly increases leakage over all classes.
Results are averaged over 5 runs.
}
\label{tab:effect_nonIID}
\end{table}

\begin{figure}[t]
\centering

\begin{subfigure}[b]{0.43\textwidth}
\centering
\includegraphics[width=\linewidth]{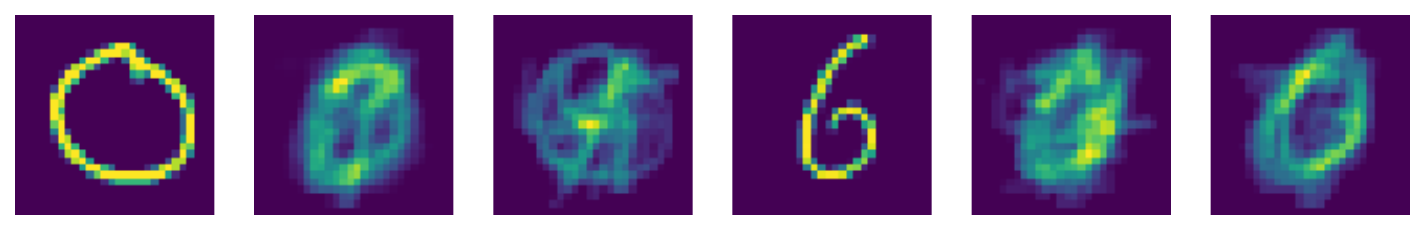}
\vspace{-0.6cm}
\caption{MNIST.}
\label{subfig:MNIST_first}
\end{subfigure} 
\vfill
\begin{subfigure}[b]{0.43\textwidth}
\centering
\includegraphics[width=\linewidth]{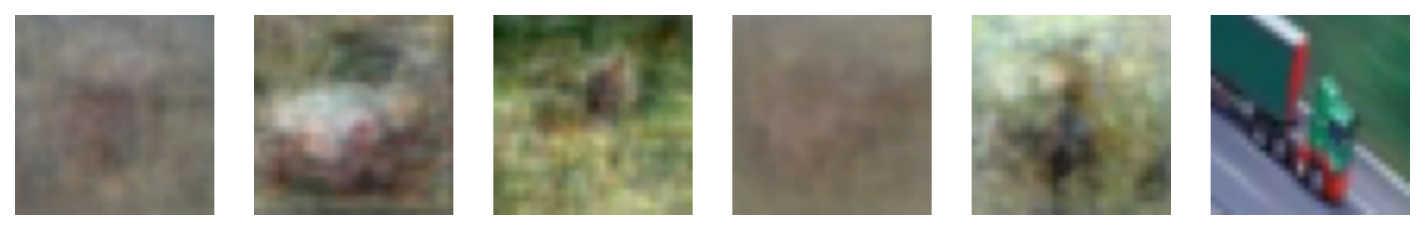}
\vspace{-0.6cm}
\caption{CIFAR10.}
\label{subfig:}
\end{subfigure}  
\vfill
\begin{subfigure}[b]{0.43\textwidth}
\centering
\includegraphics[width=\linewidth]{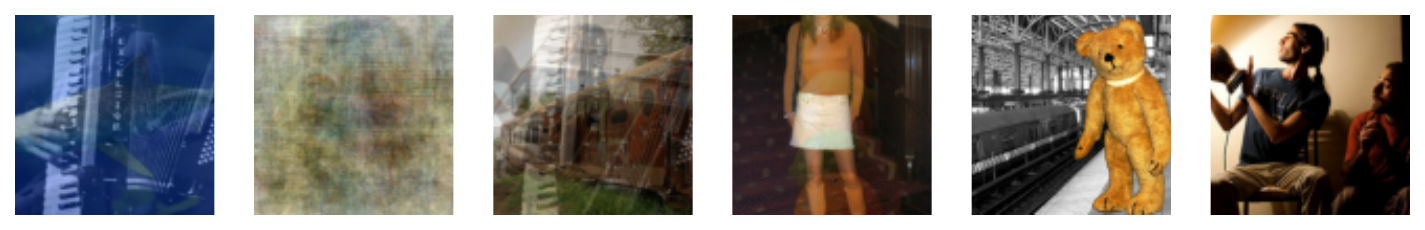}
\vspace{-0.6cm}
\caption{ImageNet.}
\label{subfig:}
\end{subfigure}  
\caption{\textbf{Passive Extraction from Non-Manipulated Model Weights.}
We extract from the gradients of the weight rows at the first fully-connected layer of a randomly initialized FC-NN with architecture from \Cref{tab:architectures} and depict data points whose features are in range [0,1]. 
These passively extracted data points can be used by an attacker to tune the hyperparameter $s$ for active extraction.}
\label{fig:extracted_first_passive}
\end{figure}

Finally, we study how an attacker without any prior knowledge can tune $s$.
One way to proceed is that the attacker does not adversarially initializes the model in the first FL iteration.
It then extracts data points from the gradients, which are not necessarily individually extracted data points.
From these data points, the attacker keeps the one in a valid image input range with features in range [0, 1], and uses these data points for fine-tuning $s$.
We depict the resulting data points for MNIST, CIFAR10, and ImageNet in \Cref{fig:extracted_first_passive} and show extraction success for different $s$ on 100 such data point in \Cref{tab:extraction_from_random_gradients}.

\begin{table}[t]
\scriptsize
\centering
\begin{tabular}{lccc}
\toprule
{\textbf{s}} & MNIST R  & CIFAR10 R  & ImageNet R \\
\midrule
$.650$   &0.468&0.0& 0.0\\
$.700$   &0.477&0.0& 0.0\\
$.750$   &\textbf{0.603}&0.0& 0.0\\
$.800$  &\textbf{0.603}&0.085&0.0\\
$.900$  &0.531&0.121&  0.0\\
$.910$ &0.504& 0.147&0.0\\
$.920$ &0.513 &0.178& 0.0\\
$.930$ &0.459 &0.210& 0.0\\
$.940$ &0.450 &0.222& 0.0\\
$.950$ & 0.513&\textbf{0.238}& 0.0 \\
$.960$ & 0.378&0.229& 0.0\\
$.970 $ & 0.450&0.191& 0.073\\
$.980$   &0.315 &0.012& 0.232\\
$.990$   & 0.360&0.0& \textbf{0.422}\\
$.995$  & 0.342&0.0& 0.330\\
$.999$ &0.270 &0.0& 0.0\\
\bottomrule
\end{tabular}
\caption{\textbf{Tuning Factor $s$ on Data Points from Passive Extraction.} We model an attacker who does not hold auxiliary data to tune the scaling factor $s$. 
Such an attacker can, during the firs round, of the protocol extract the data points from the non-manipulated model weights' gradients.
The points (we select those with features in range [0,1], see \Cref{fig:extracted_first_passive}), can be used to tune $s$.
The identified optimal $s$ (bold) w.r.t. the \recall are close (0.75 or 0.80, MNIST) or identical (0.95, CIFAR10 and 0.99, ImageNet) to the original datasets' optimal $s$, 0.7, 0.95, and 0.99 for MNIST, CIFAR10, ImageNet.}
\label{tab:extraction_from_random_gradients}
\end{table}

\begin{figure*}[t]
\centering
\begin{subfigure}[b]{0.44\textwidth}
\centering
\includegraphics[width=\linewidth, trim={0cm 17.7cm 0cm 0cm},clip]{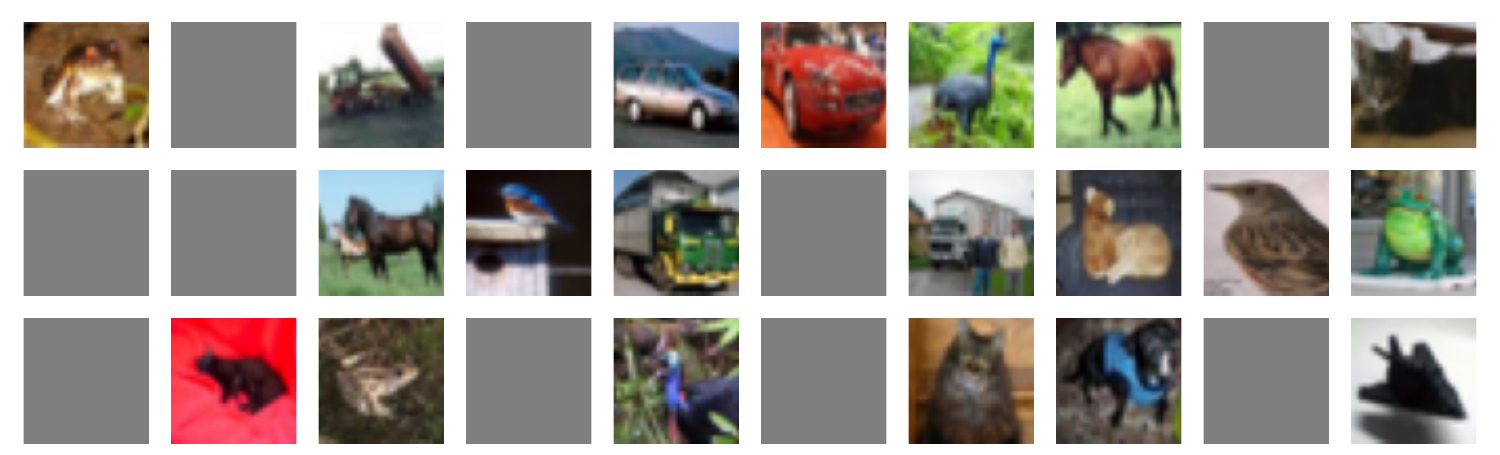}
\caption{Reconstructed data points from different classes.}
\label{fig:cifar_first_layer}
\end{subfigure}
\hfill
\begin{subfigure}[b]{0.44\textwidth}
\centering
\includegraphics[width=\linewidth, trim={0cm 17.7cm 0cm 0cm},clip]{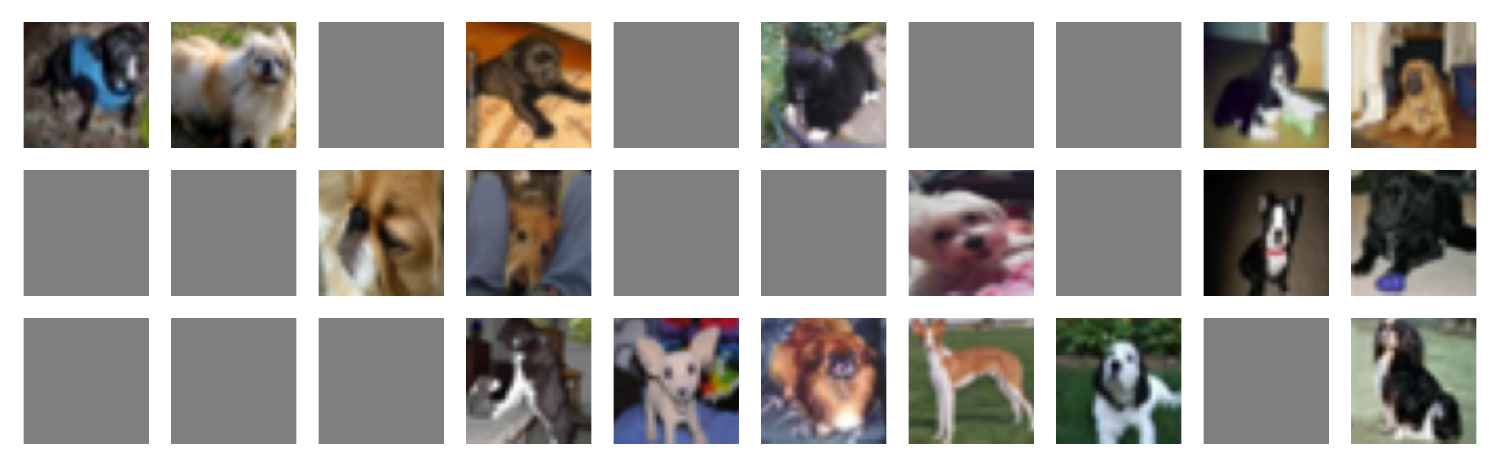}
\caption{Reconstructed data points from class "dog".}
\label{fig:cifar_first_layer_dog}
\end{subfigure} 
\begin{subfigure}[b]{0.44\textwidth}
\centering
\includegraphics[width=\linewidth, trim={0cm 17.7cm 0cm 0cm},clip]{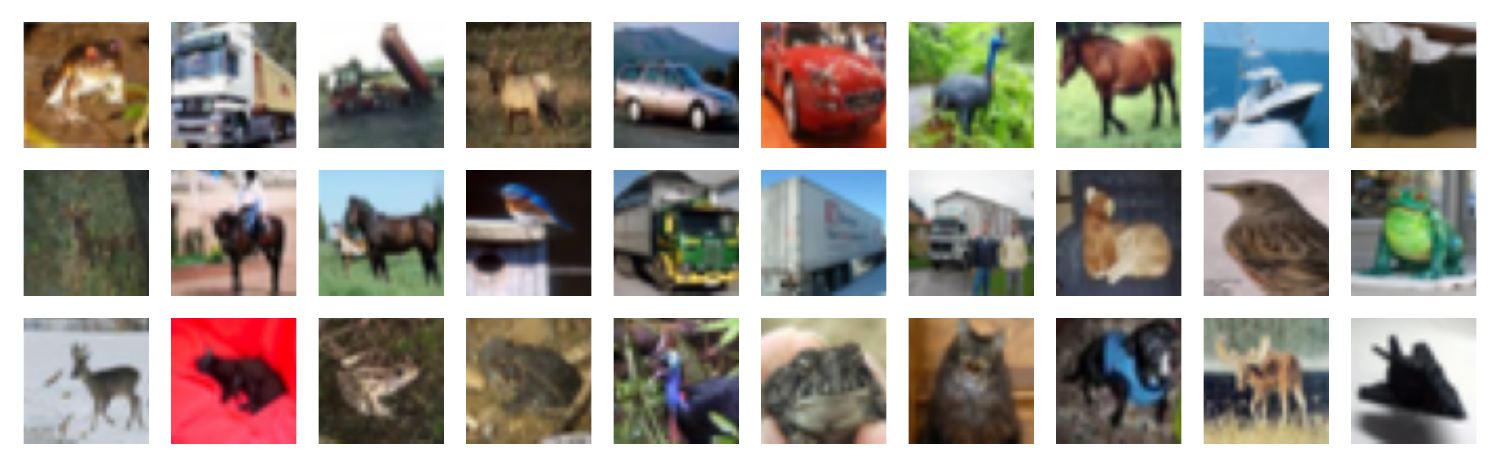}
\caption{Original data points from different classes.}
\label{fig:cifar_orig_all}
\end{subfigure} 
\hfill
\begin{subfigure}[b]{0.44\textwidth}
\centering
\includegraphics[width=\linewidth, trim={0cm 17.7cm 0cm 0cm},clip]{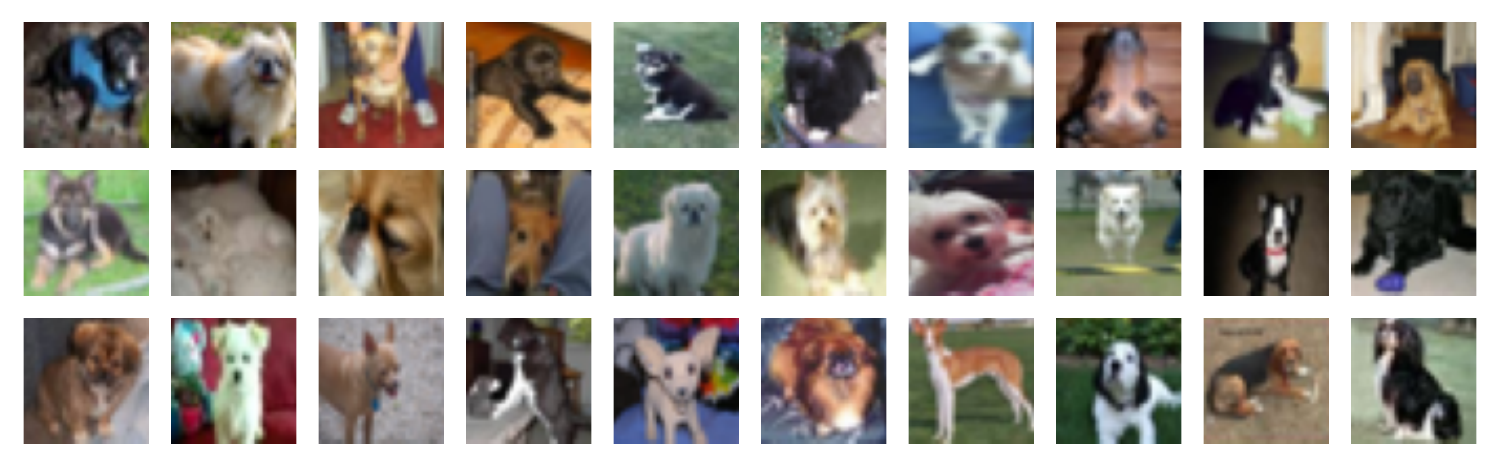}
\caption{Original data points from class "dog".}
\label{fig:cifar_orig_dog}
\end{subfigure}   
\caption{\textbf{CIFAR10 Data Extracted from Mini-Batches with Data Points from Different and the Same Class.} Reconstruction success of our adversarial initialization: first 30 images from a mini-batch of 100 data points, extracted at the first fully-connected layer of the CNN from \Cref{tab:architectures}. Gray images indicate that the corresponding original data point could not be extracted individually from the model gradients. Our extraction success for data from the same class is as high as for data from different classes.}
\label{fig:cifar_all}
\end{figure*}

\subsection{Extraction under Lossy Layers}
\label{sub:dropout_maxpool}

We also study the effect of "lossy" layers, such as dropout and pooling on our data extraction success.
Therefore, we rely on the following architecture proposed by~\cite{agarwal2021skellam} for FL, see \Cref{tab:architectures_fl+dp+ssa}.
\begin{table}[tbh]
    \centering
    \scriptsize
    \begin{tabular}{c}
    \toprule
    CNN Architecture by~\cite{agarwal2021skellam} \\
    \midrule 
        Conv(f=32, k=(3,3), s=1, p=same, act=relu)\\
        MaxPool() \\
        Conv(f=64, k=(3,3), s=1, p=same, act=relu)\\
        Dropout()\\
        Flatten\\
        Dense(n=1000, act=relu)\\
        Dropout()\\
        Dense(n=\#classes, act=None)\\
    \bottomrule
    \end{tabular}
    \caption{CNN Architecture by~\cite{agarwal2021skellam} used to evaluate the data extraction attack under the impact of Dropout and Pooling. f: number of filters, k: kernel size, s: stride, p: padding act: activation function, n: number of neurons.}
    \label{tab:architectures_fl+dp+ssa}
\end{table}

\begin{figure}[tbh]
\centering
\begin{subfigure}[b]{0.43\textwidth}
\centering
\includegraphics[width=\linewidth]{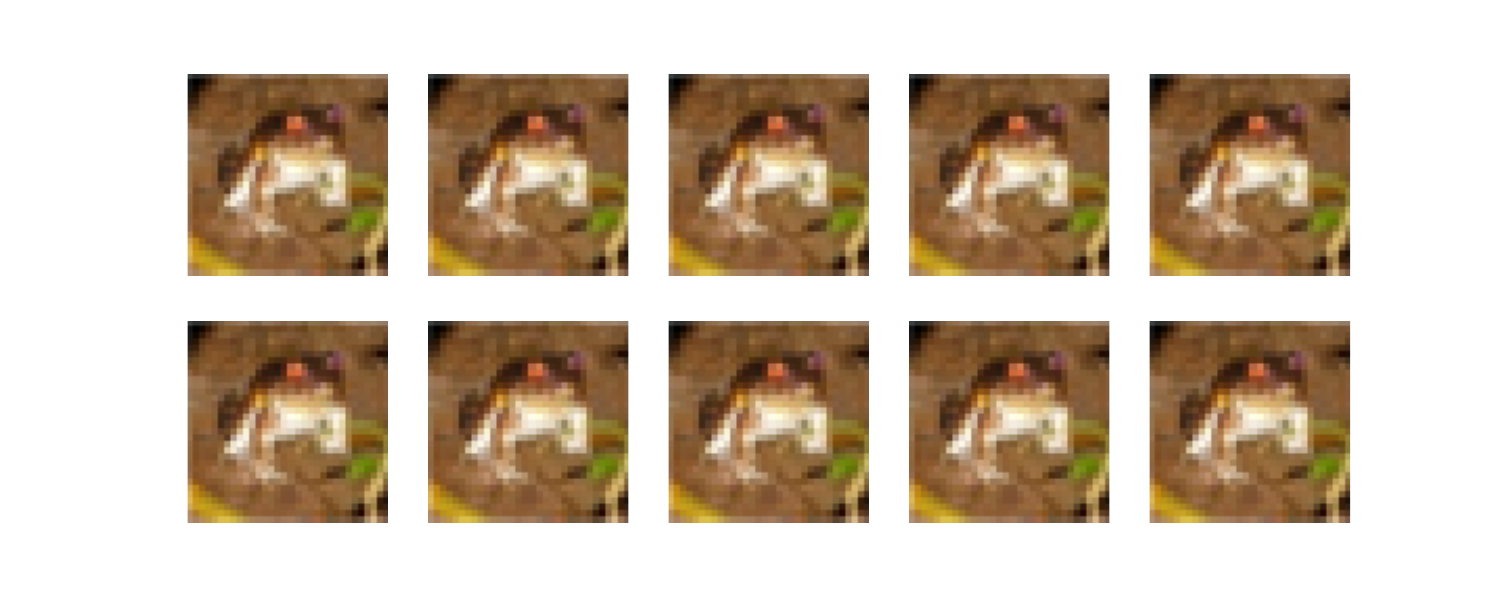}
\caption{Dropout with $p=0.0$.}
\end{subfigure}     
\begin{subfigure}[b]{0.43\textwidth}
\centering
\includegraphics[width=\linewidth]{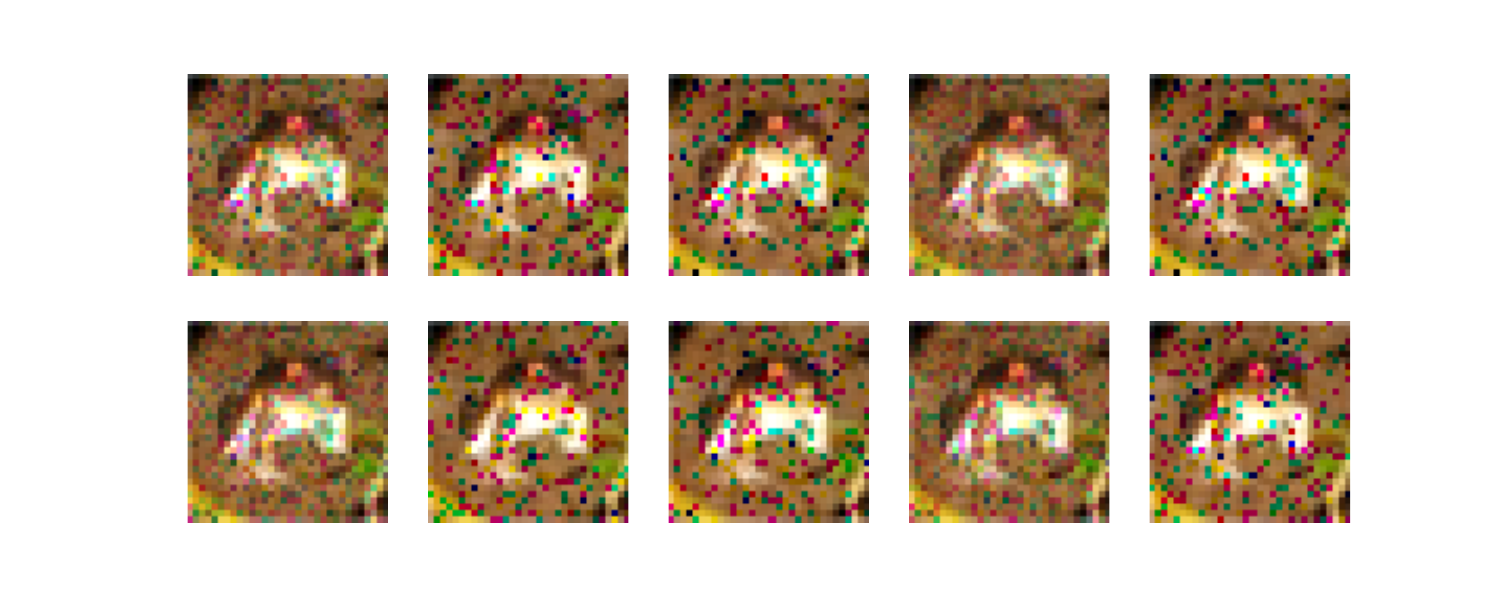}
\caption{Dropout with $p=0.1$.}
\end{subfigure}   
\begin{subfigure}[b]{0.43\textwidth}
\centering
\includegraphics[width=\linewidth]{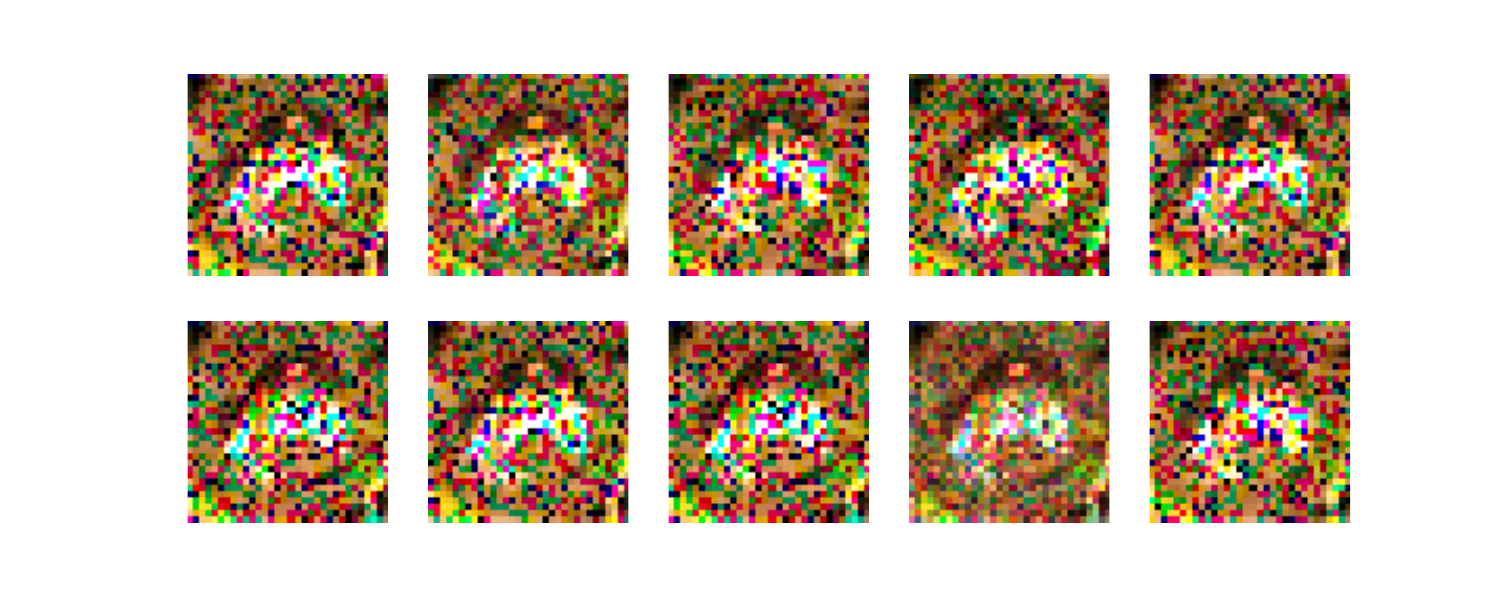}
\caption{Dropout with $p=0.3$.}
\end{subfigure}   
\begin{subfigure}[b]{0.43\textwidth}
\centering
\includegraphics[width=\linewidth]{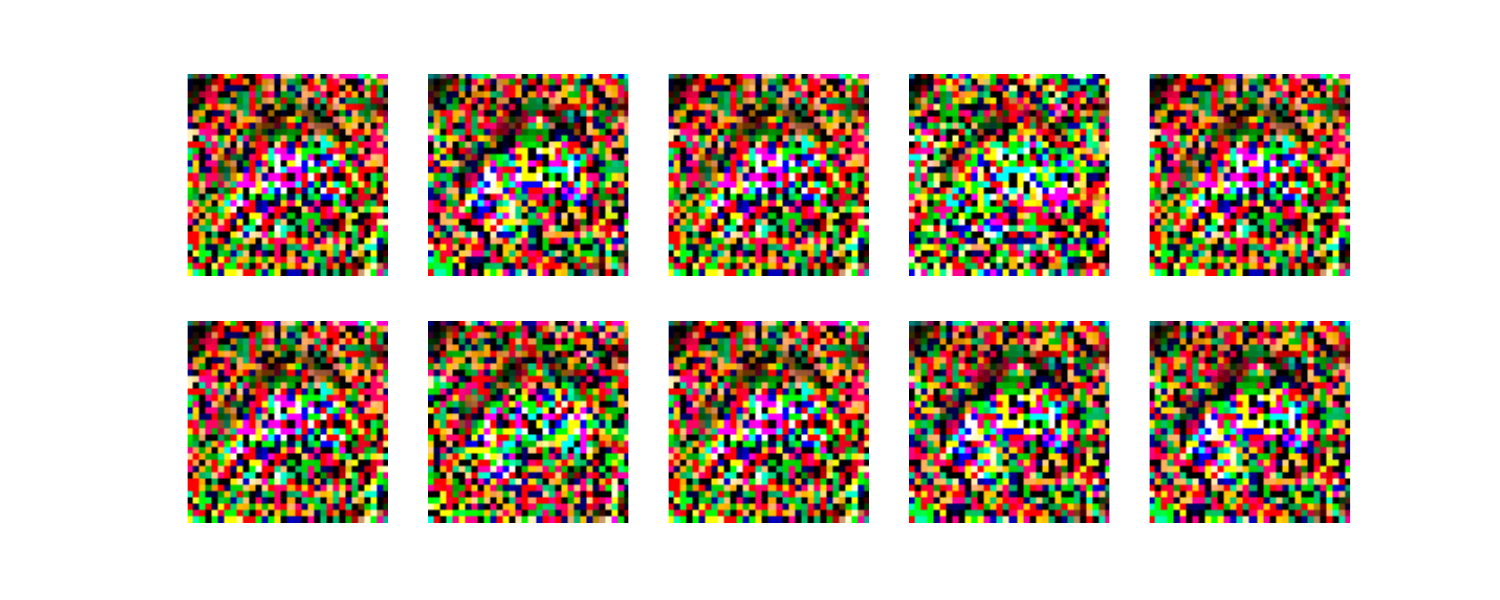}
\caption{Dropout with $p=0.5$.}
\end{subfigure}   
\begin{subfigure}[b]{0.43\textwidth}
\centering
\includegraphics[width=\linewidth]{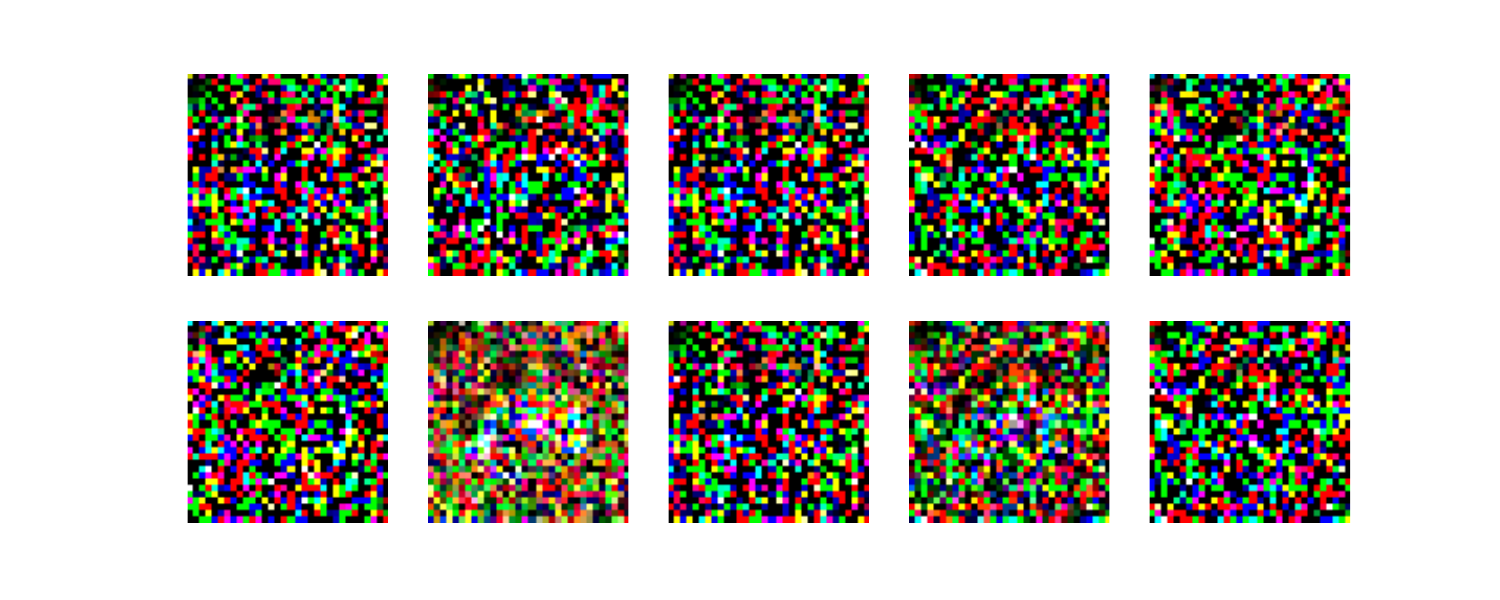}
\caption{Dropout with $p=0.7$.}
\end{subfigure}   
\begin{subfigure}[b]{0.43\textwidth}
\centering
\includegraphics[width=\linewidth]{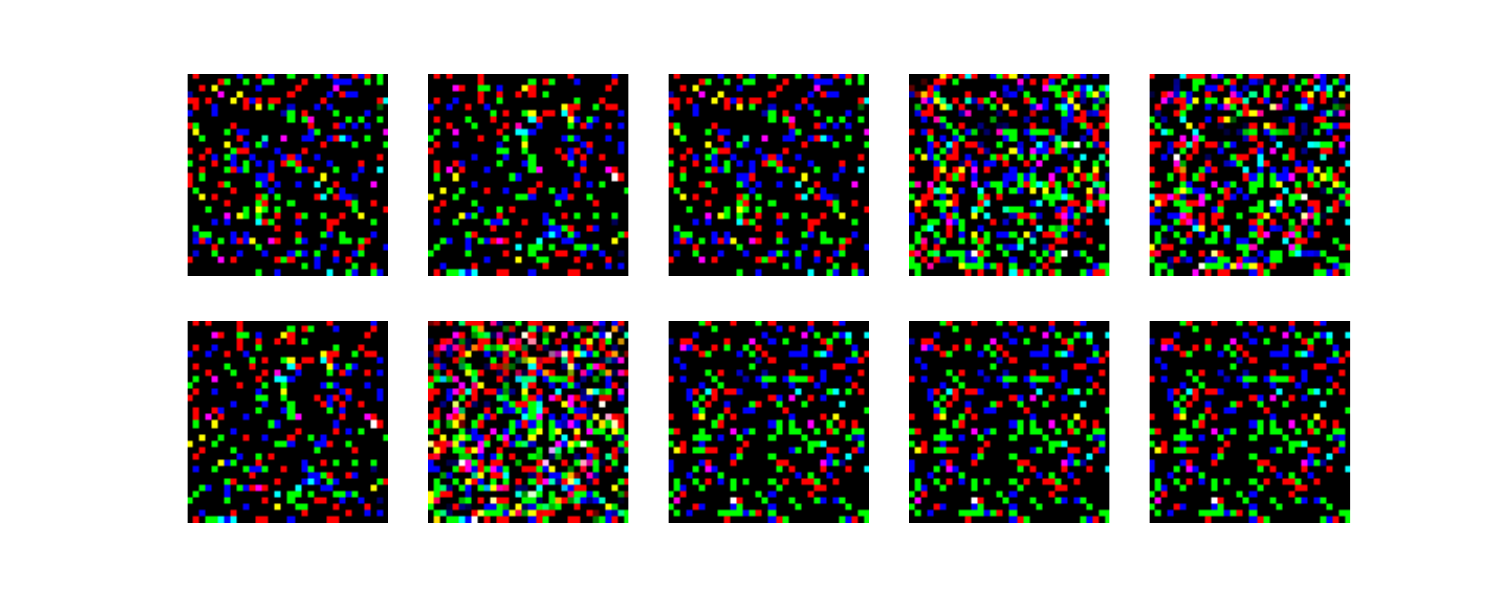}
\caption{Dropout with $p=0.9$.}
\end{subfigure}   
\caption{Batch size 1.}
\label{fig:dropouts}
\end{figure}

\begin{figure}[htb]
\centering
\begin{subfigure}[b]{0.43\textwidth}
\centering
\includegraphics[width=\linewidth]{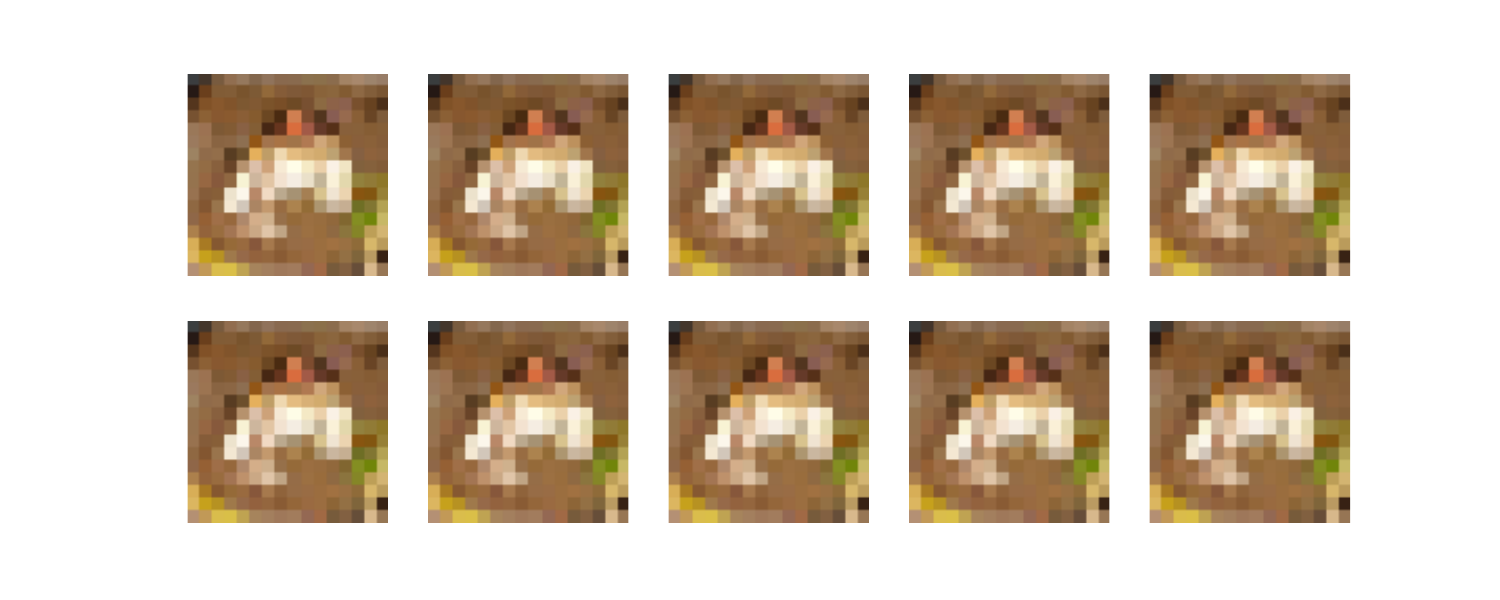}
\caption{Dropout with $p=0.0$ and pooling.}
\label{fig:pooling_alone}
\end{subfigure}     
\begin{subfigure}[b]{0.43\textwidth}
\centering
\includegraphics[width=\linewidth]{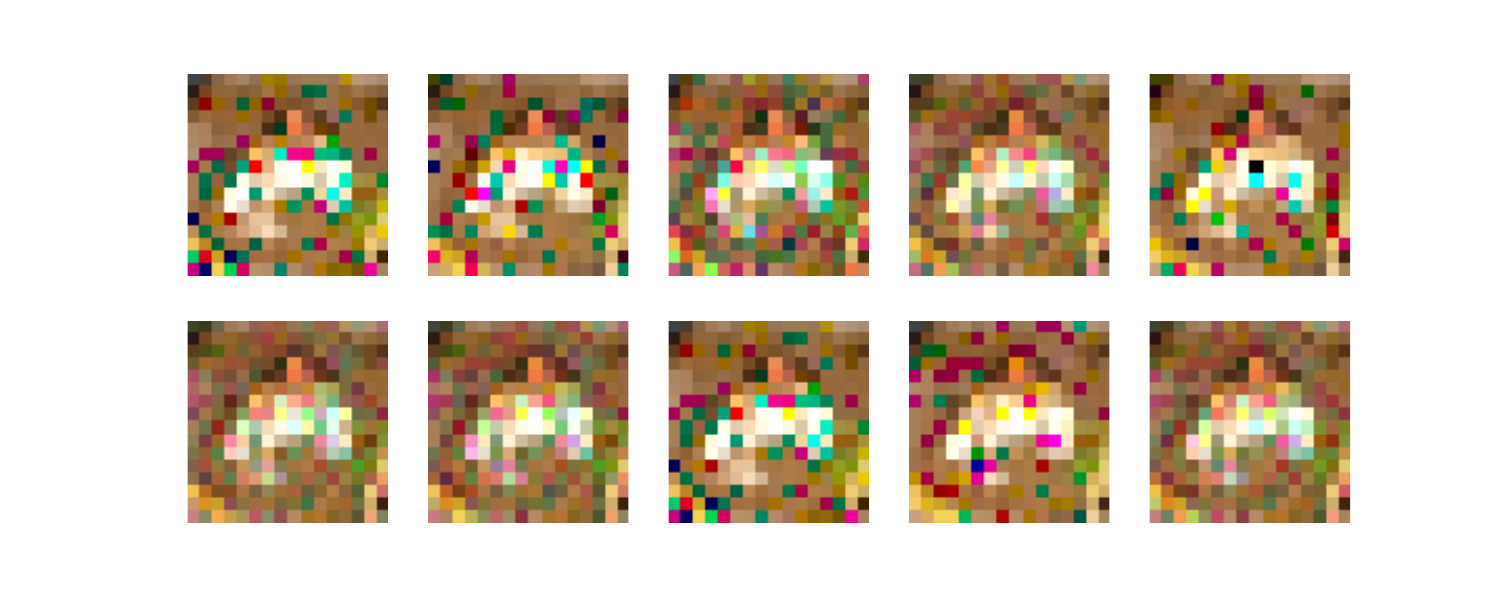}
\caption{Dropout with $p=0.1$ and pooling.}
\end{subfigure}   
\begin{subfigure}[b]{0.43\textwidth}
\centering
\includegraphics[width=\linewidth]{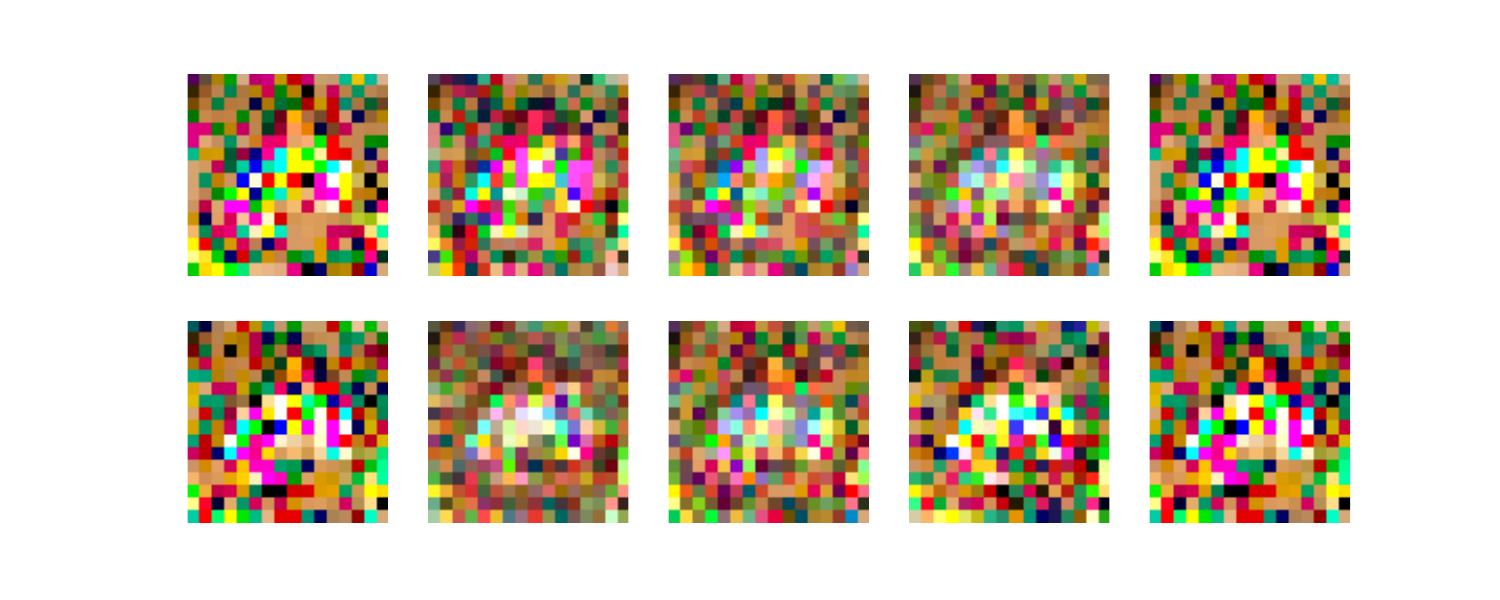}
\caption{Dropout with $p=0.3$ and pooling.}
\end{subfigure}   
\begin{subfigure}[b]{0.43\textwidth}
\centering
\includegraphics[width=\linewidth]{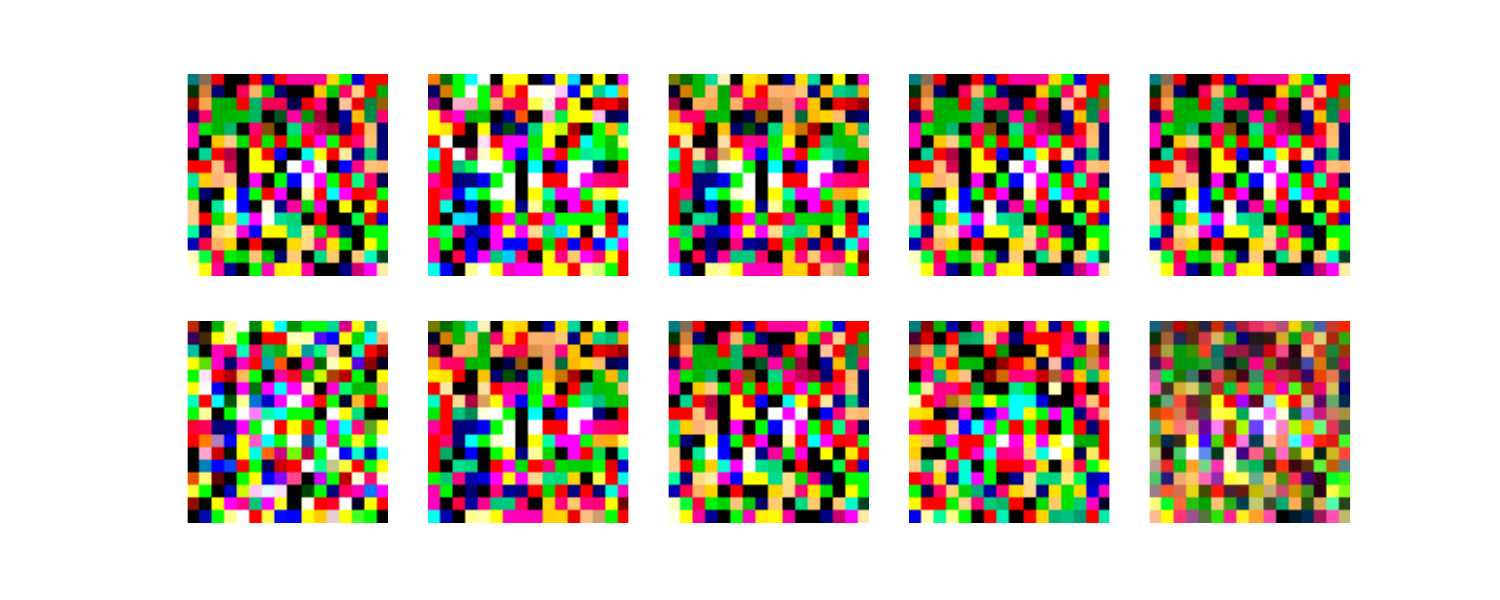}
\caption{Dropout with $p=0.5$ and pooling.}
\end{subfigure}   
\begin{subfigure}[b]{0.43\textwidth}
\centering
\includegraphics[width=\linewidth]{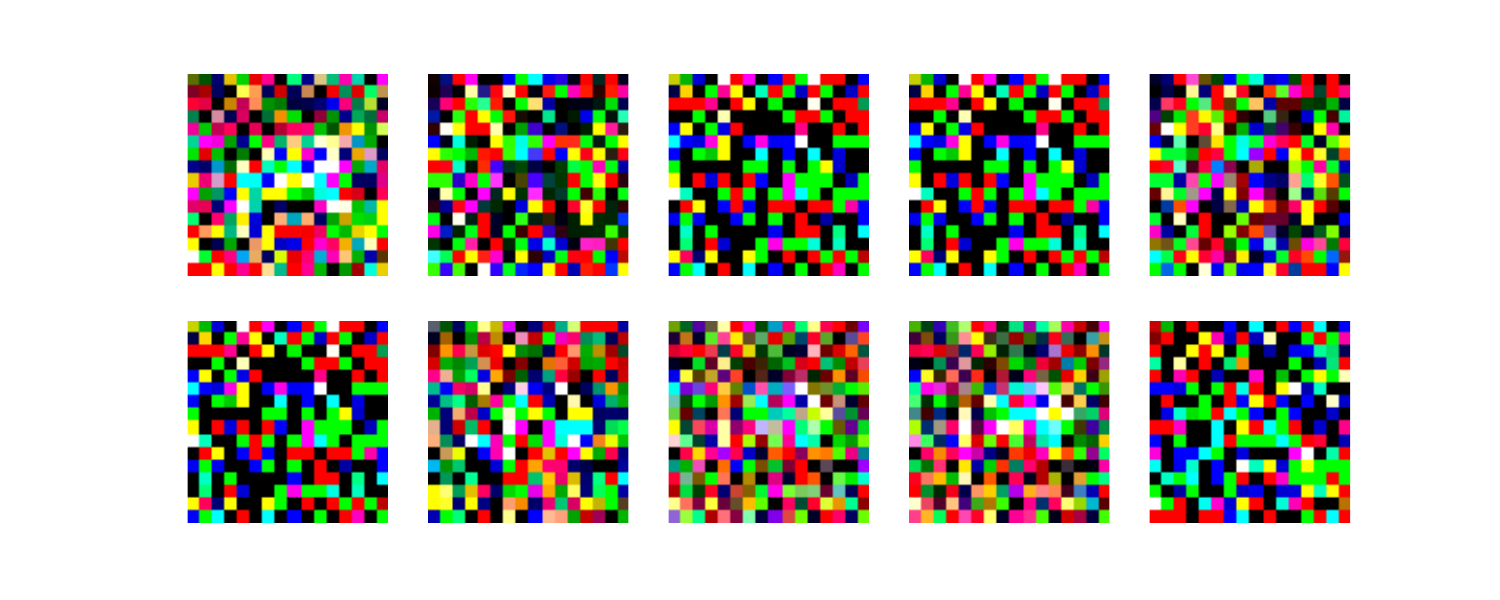}
\caption{Dropout with $p=0.7$ and pooling.}
\end{subfigure}   
\begin{subfigure}[b]{0.43\textwidth}
\centering
\includegraphics[width=\linewidth]{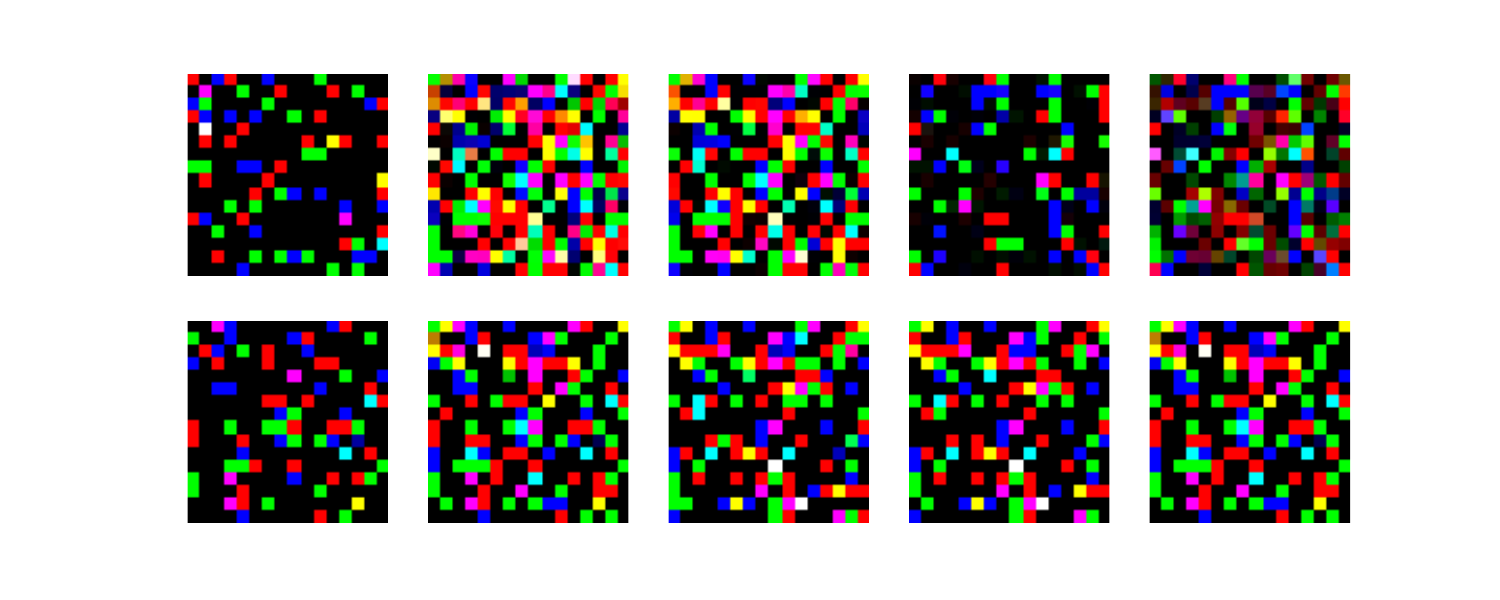}
\caption{Dropout with $p=0.9$ and pooling.}
\end{subfigure}   
\caption{Batch size 1.}
\label{fig:droppools}
\end{figure}

\begin{figure}[htb]
\centering
\begin{subfigure}[b]{0.43\textwidth}
\centering
\includegraphics[width=\linewidth]{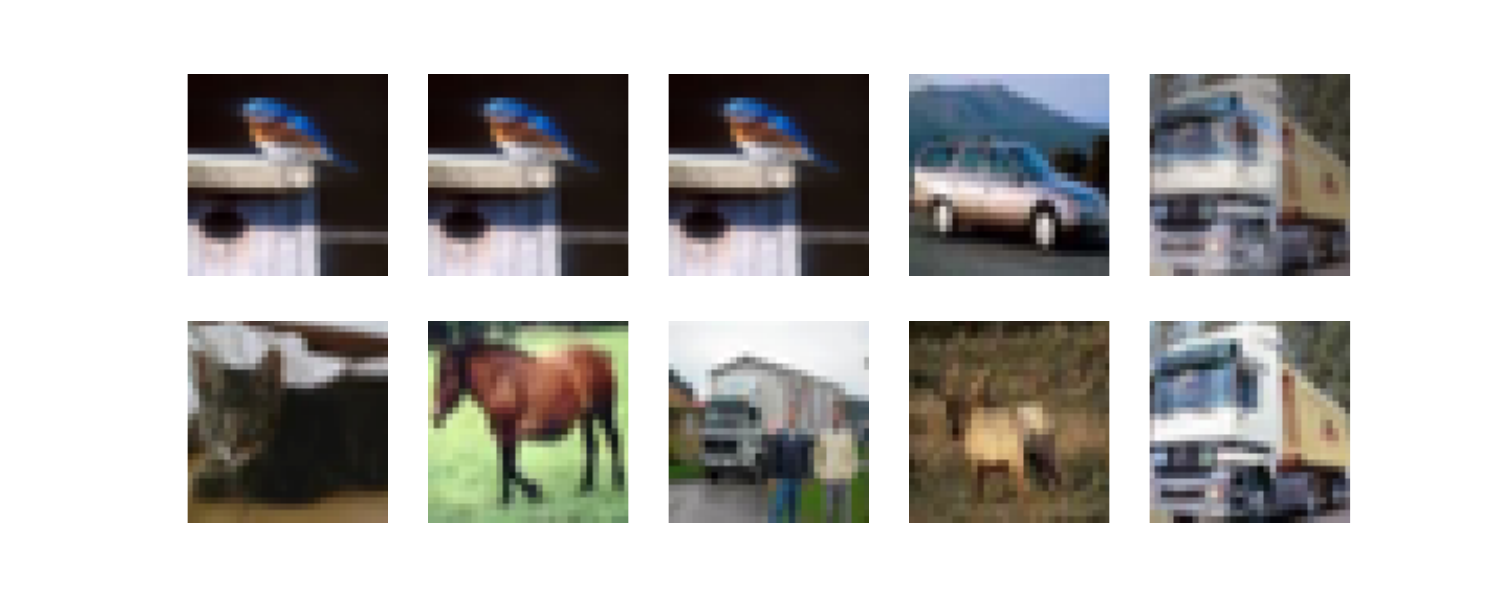}
\caption{Dropout with $p=0.0$.}
\end{subfigure}     
\begin{subfigure}[b]{0.43\textwidth}
\centering
\includegraphics[width=\linewidth]{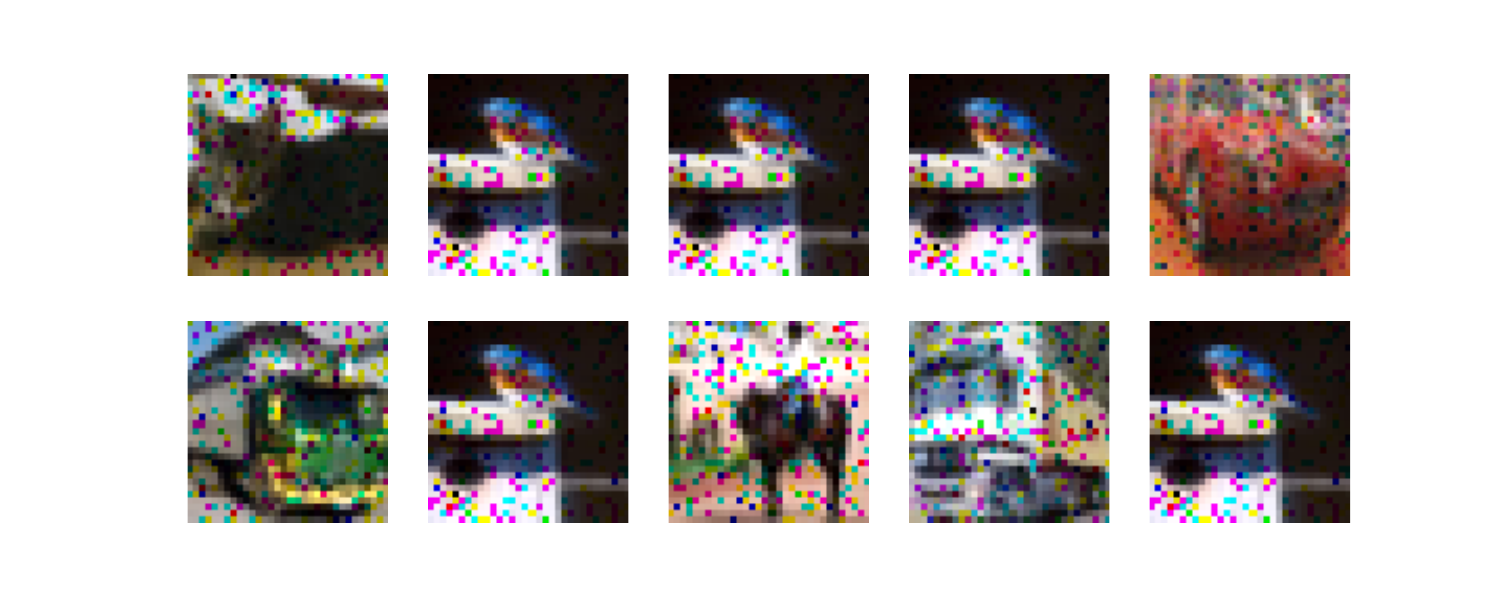}
\caption{Dropout with $p=0.1$.}
\end{subfigure}   
\begin{subfigure}[b]{0.43\textwidth}
\centering
\includegraphics[width=\linewidth]{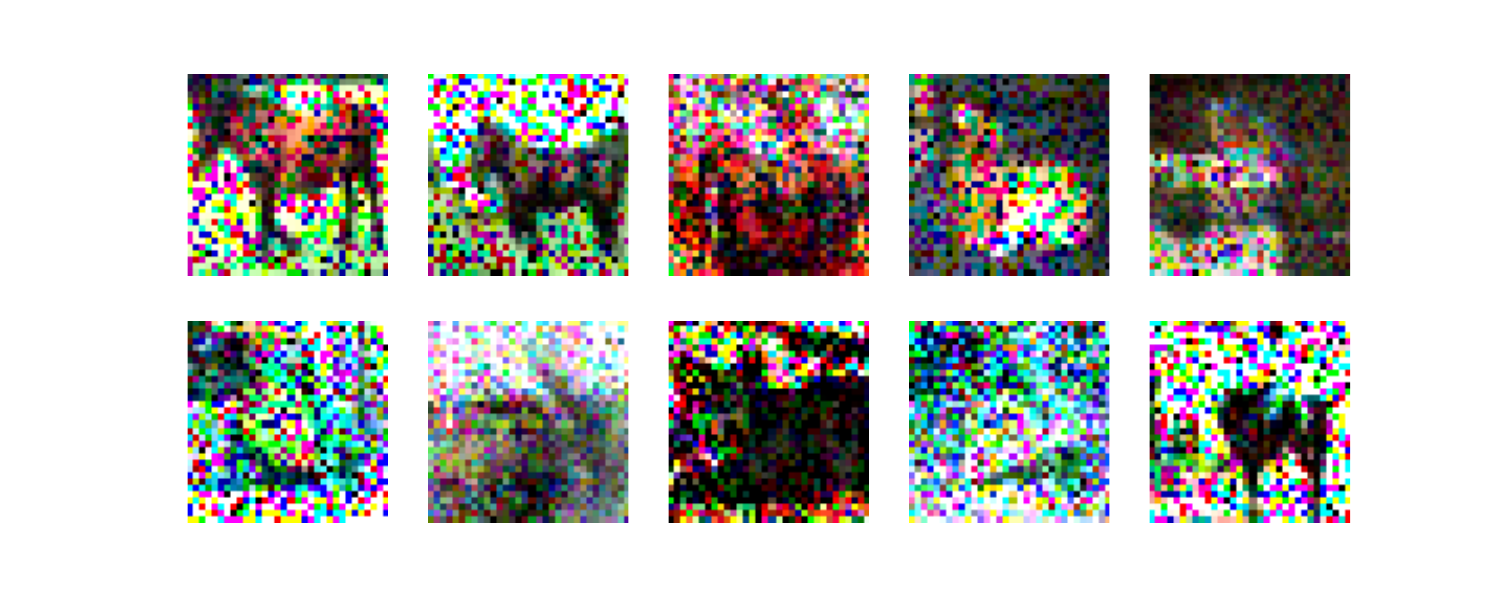}
\caption{Dropout with $p=0.3$.}
\end{subfigure}   
\begin{subfigure}[b]{0.43\textwidth}
\centering
\includegraphics[width=\linewidth]{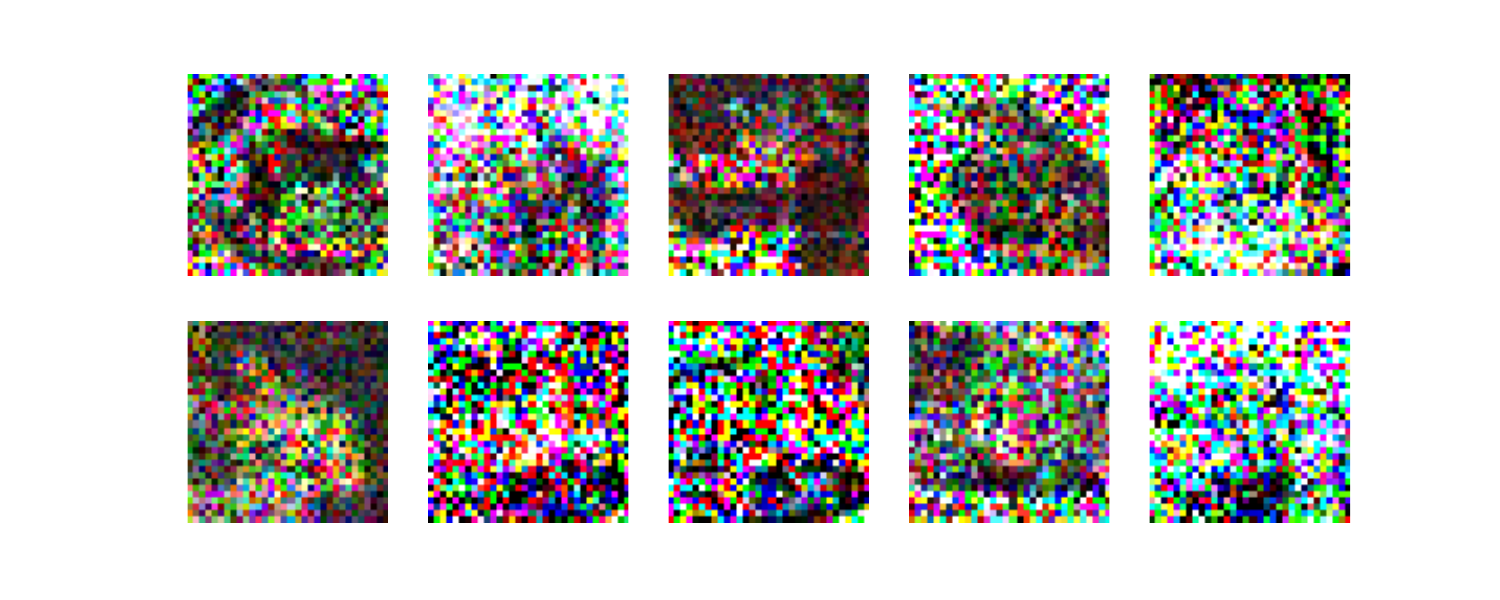}
\caption{Dropout with $p=0.5$.}
\end{subfigure}   
\begin{subfigure}[b]{0.43\textwidth}
\centering
\includegraphics[width=\linewidth]{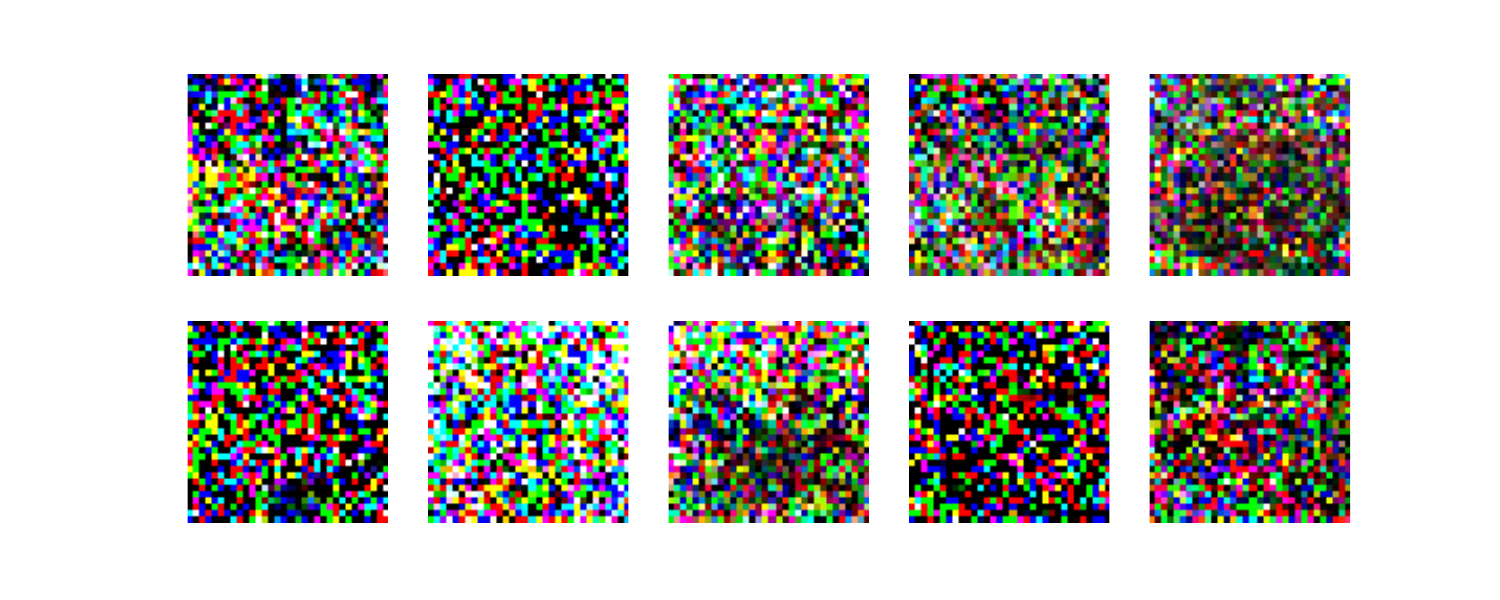}
\caption{Dropout with $p=0.7$.}
\end{subfigure}   
\begin{subfigure}[b]{0.43\textwidth}
\centering
\includegraphics[width=\linewidth]{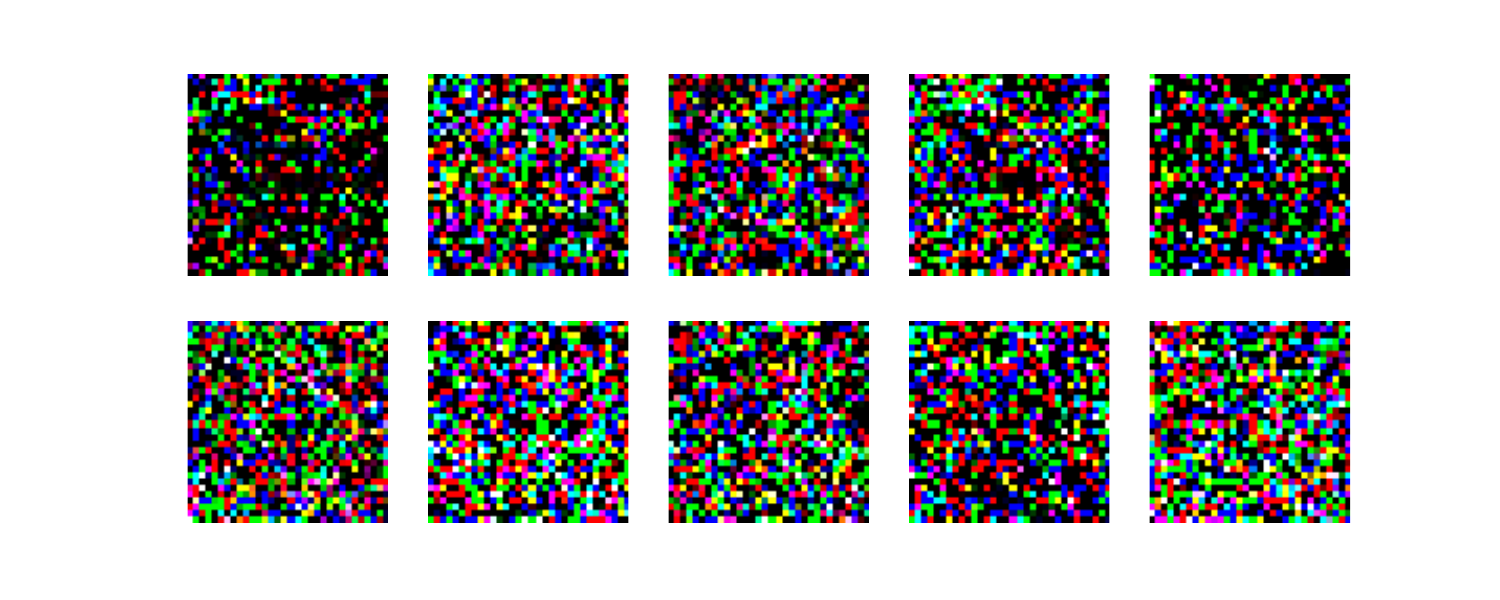}
\caption{Dropout with $p=0.9$.}
\end{subfigure}   
\caption{Batch size 20.}
\label{fig:dropouts_20}
\end{figure}

\begin{figure}[htb]
\centering
\begin{subfigure}[b]{0.43\textwidth}
\centering
\includegraphics[width=\linewidth]{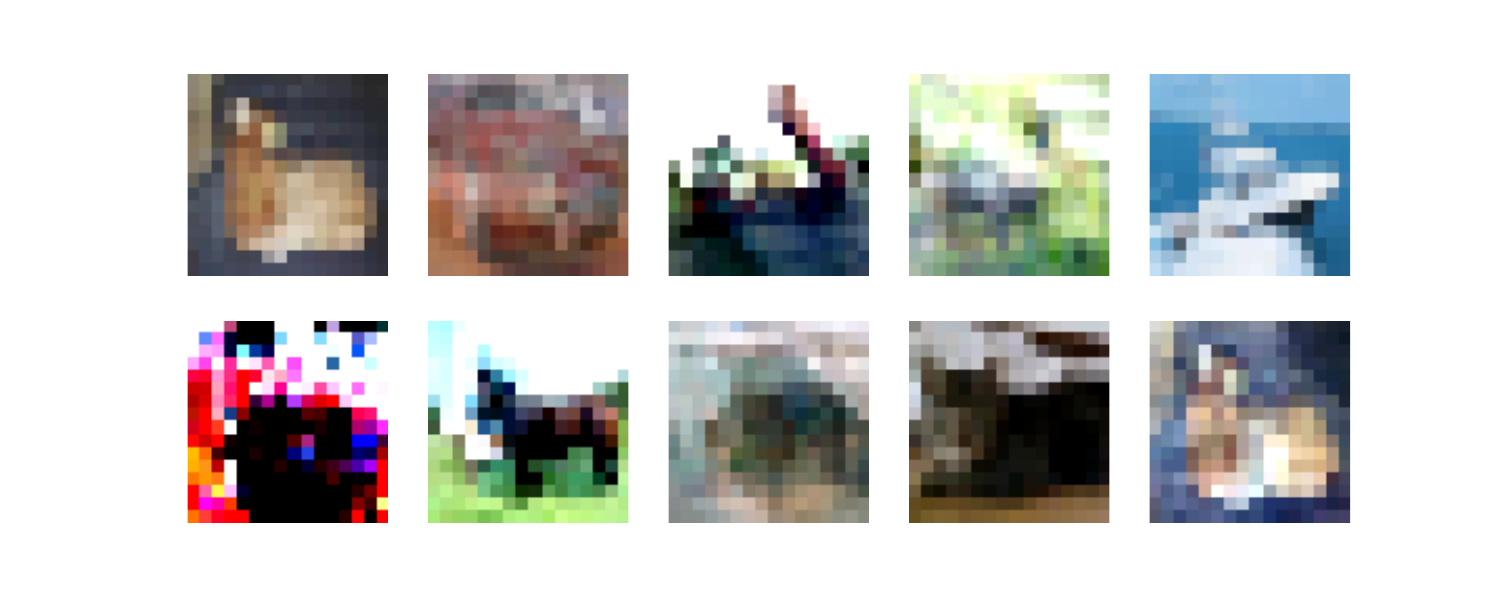}
\caption{Dropout with $p=0.0$ and pooling.}
\end{subfigure}     
\begin{subfigure}[b]{0.43\textwidth}
\centering
\includegraphics[width=\linewidth]{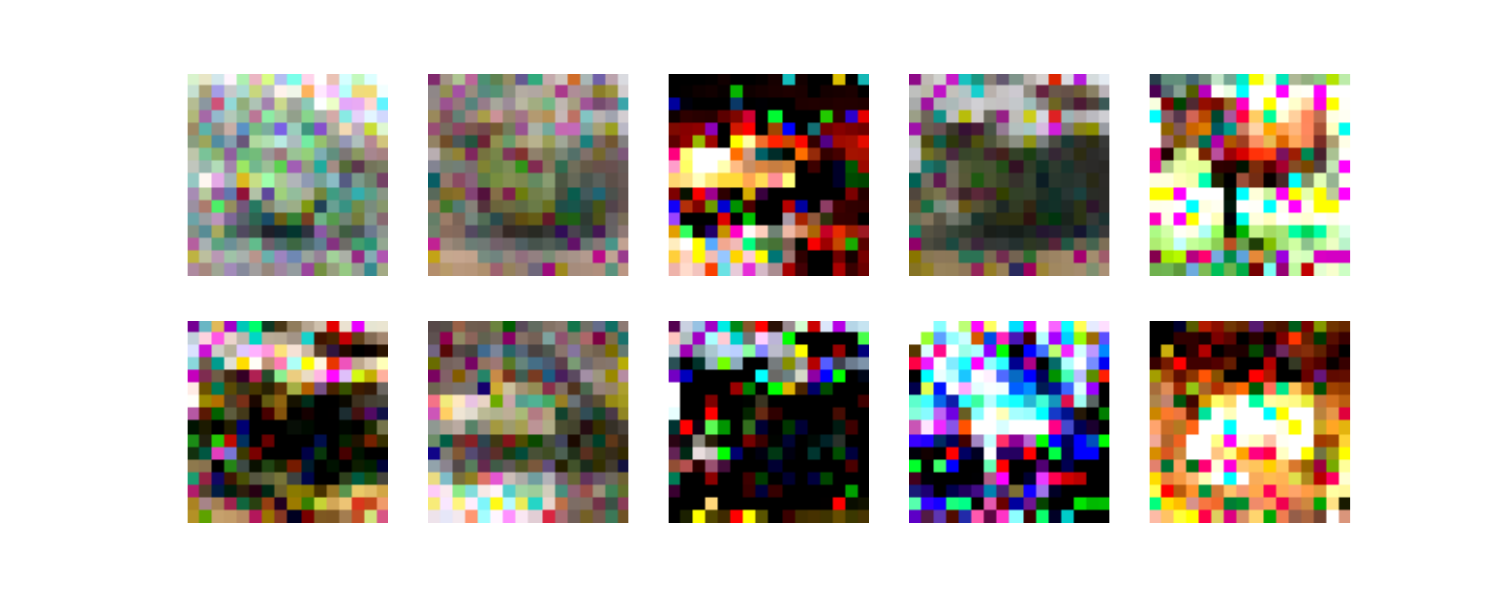}
\caption{Dropout with $p=0.1$ and pooling.}
\end{subfigure}   
\begin{subfigure}[b]{0.43\textwidth}
\centering
\includegraphics[width=\linewidth]{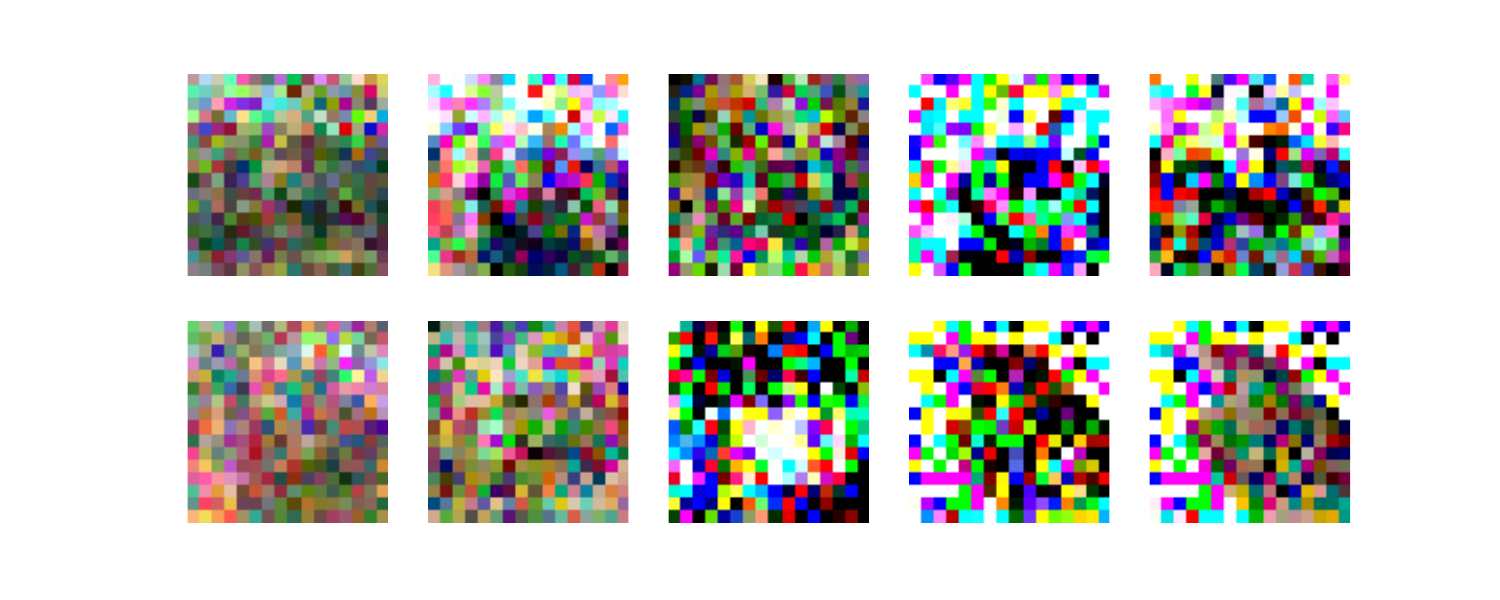}
\caption{Dropout with $p=0.3$ and pooling.}
\end{subfigure}   
\begin{subfigure}[b]{0.43\textwidth}
\centering
\includegraphics[width=\linewidth]{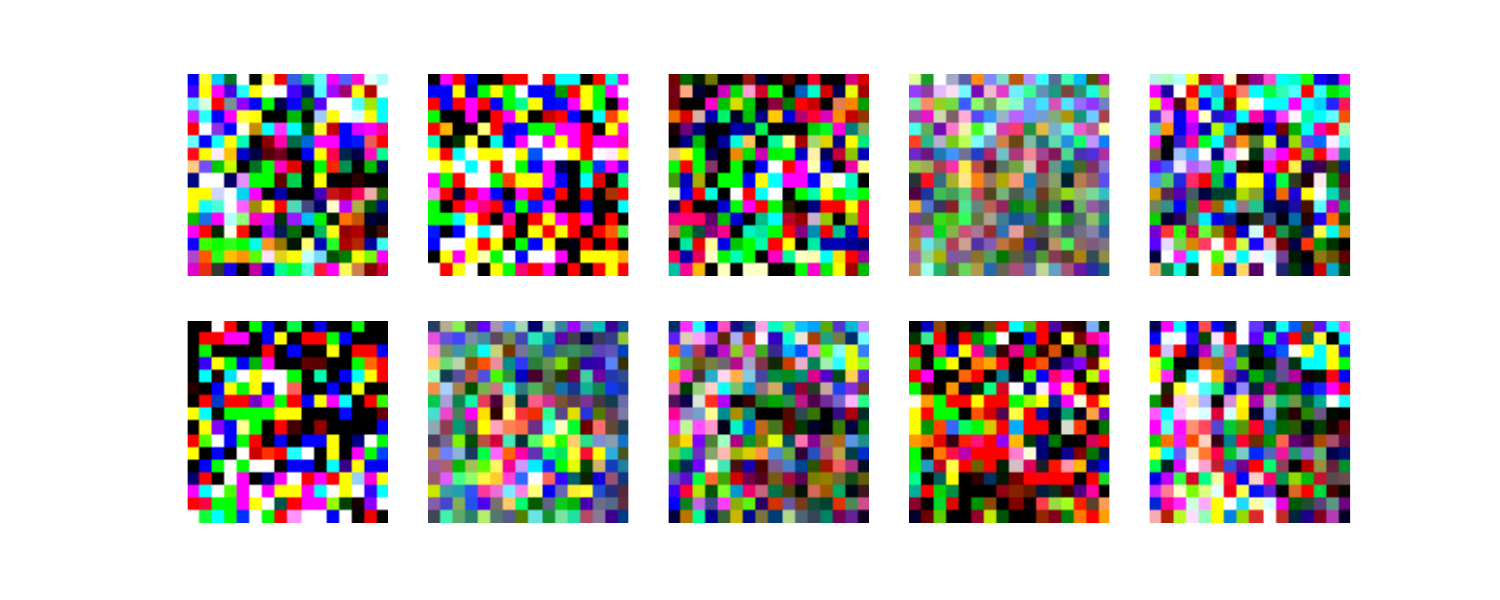}
\caption{Dropout with $p=0.5$ and pooling.}
\end{subfigure}   
\begin{subfigure}[b]{0.43\textwidth}
\centering
\includegraphics[width=\linewidth]{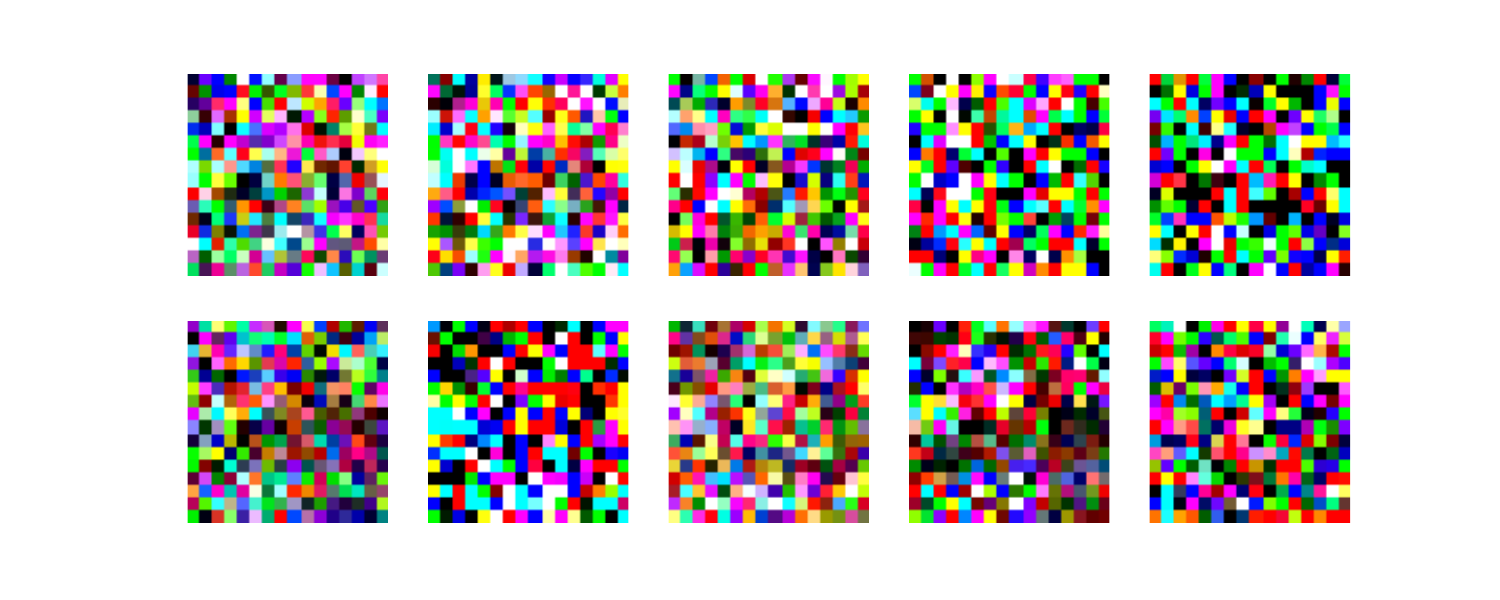}
\caption{Dropout with $p=0.7$ and pooling.}
\end{subfigure}   
\begin{subfigure}[b]{0.43\textwidth}
\centering
\includegraphics[width=\linewidth]{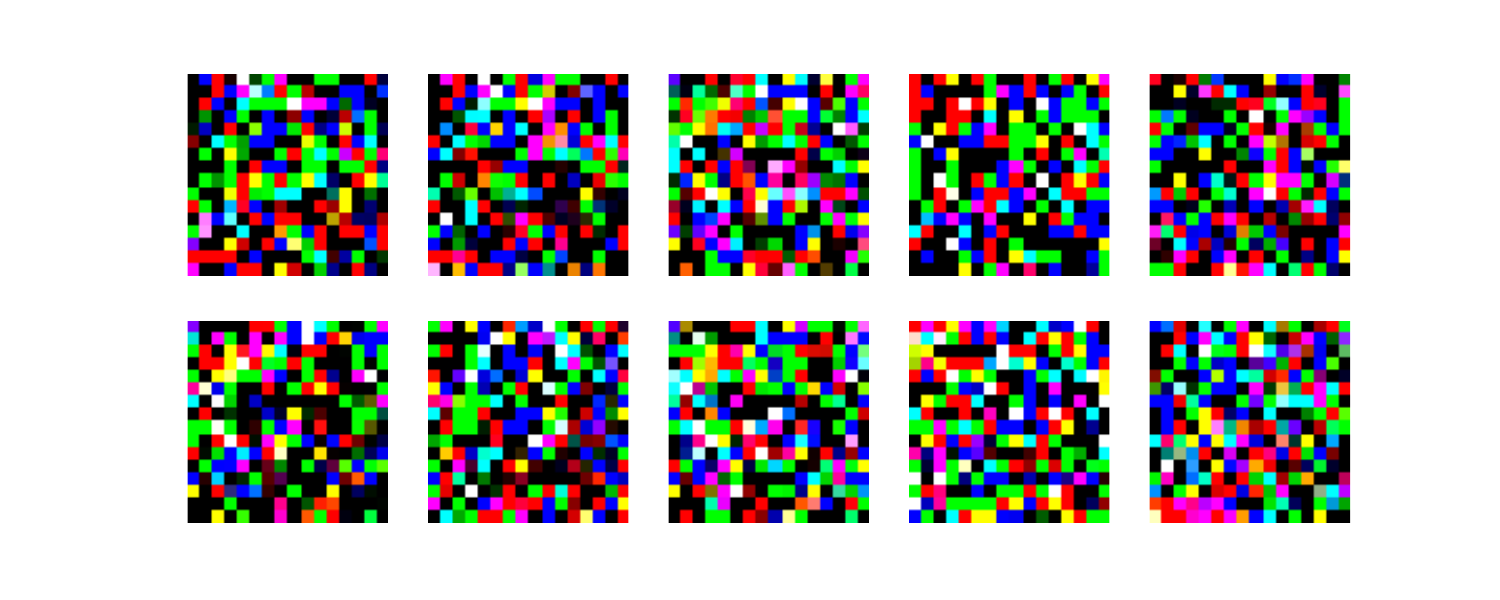}
\caption{Dropout with $p=0.9$ and pooling.}
\end{subfigure}   
\caption{Batch size 20.}
\label{fig:droppools_20}
\end{figure}

\Cref{fig:dropouts,fig:droppools} and \Cref{fig:dropouts_20,fig:droppools_20} show individual effects of dropout and pooling layers on a reconstructions for mini-batches of size 1 and 20 respectively. 
We evaluated different dropout rates $p \in \{0.0, 0.1, 0.3, 0.5, 0.7, 0.9\}$.
Note that the second dropout layer does not have a significant impact on the success of our reconstruction since we extract from the first fully-connected model layer before information can get lost due to the second dropout.
To evaluate dropout without pooling, we remove the MaxPool layer, and to evaluate pooling without dropout, we set the dropout rate to $p=0.1$.
Although existence of non-invertible components compromises overall reconstruction fidelity, we observe it is often possible to still recognise individual data points.

\subsection{TensorFlow Federated}
\label{app:tf-federated}
We adapted the implementation of FedAvg~\cite{tffedFedAvg} 
provided by the developers to pass each individual gradient update through our reconstruction function. Note that the whole change took only minutes of work and required minimal code changes, such that they could easily be implemented by a dishonest \cp. We pre-generated our adversarial initialized shared models with scaling factor $s=0.7$ and $s=0.99$ and 1000 neurons at the first fully-connected layer, and passed them to the \users. The aggregator then collects the gradients and performs reconstruction. 

We find that our attack works consistently well against commonly used FL benchmarks integrated into the library. Over 50 \users, for EMNIST~\cite{cohen_afshar_tapson_schaik_2017}, our \namenoformat yield $0.32\pm0.07$ \recall and $0.05\pm0.02$ \precision, versus $0.10\pm0.05$, and $0.02\pm0.01$ in the non-adversarial baseline.
For CIFAR100~\cite{Krizhevsky09learningmultiple} we get $0.44\pm0.05$ \recall and $0.22\pm0.06$ \precision, versus $0.20\pm0.04$, and $0.04\pm0.01$ in the baseline. These results are comparable to the ones reported in the previous experiments, and confirm that our attack is practical. 

\clearpage

\begin{figure*}[tbh]
\centering
\begin{subfigure}[]{0.24\textwidth}
\centering
\includegraphics[width=\linewidth]{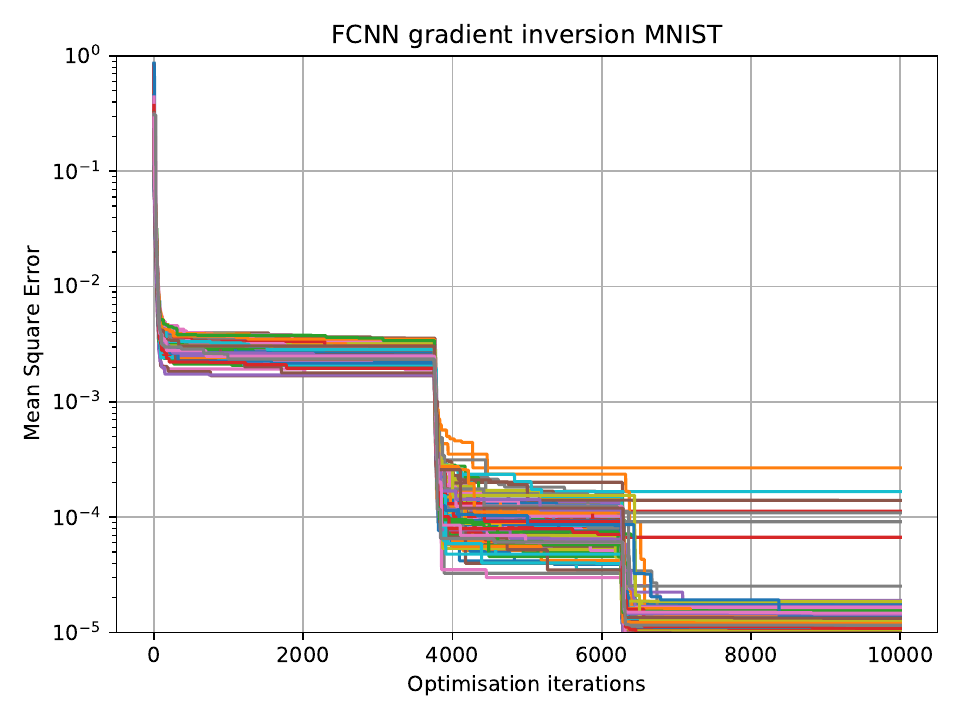}
\caption{FC-NN}
\label{subfig:mnist_fcnn_inv}
\end{subfigure}
\begin{subfigure}[]{0.24\textwidth}
\centering
\includegraphics[width=\linewidth]{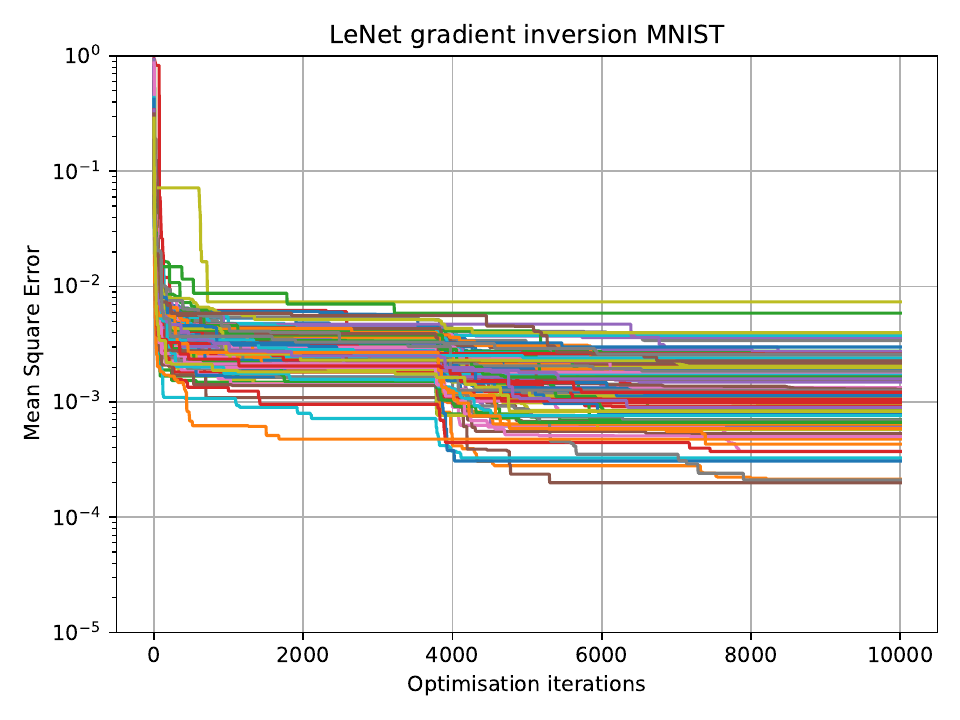}
\caption{LeNet-5}
\end{subfigure}
\begin{subfigure}[]{0.24\textwidth}
\centering
\includegraphics[width=\linewidth]{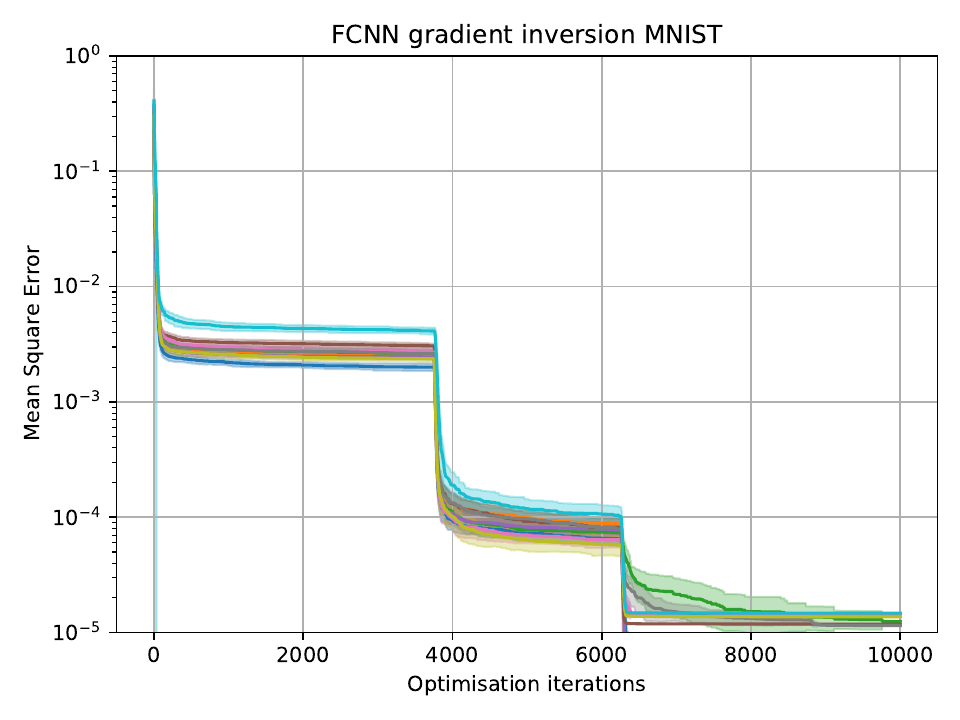}
\caption{FC-NN}
\label{subfig:restarts_mnist_fcnn_inv}
\end{subfigure}  
\begin{subfigure}[]{0.24\textwidth}
\centering
\includegraphics[width=\linewidth]{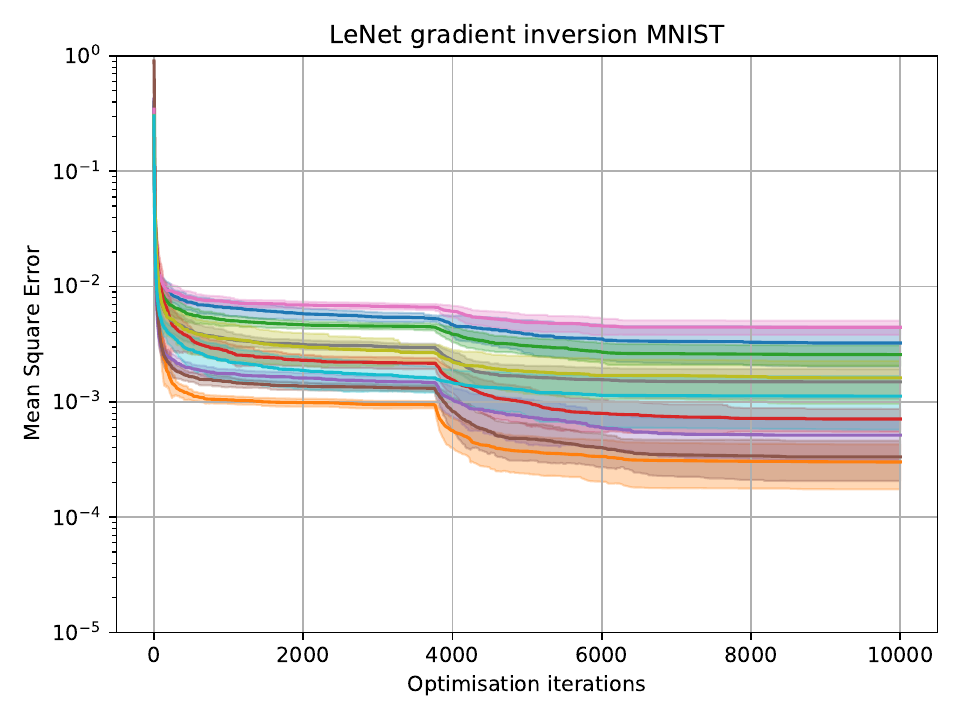}
\caption{LeNet-5}
\label{subfig:restarts_mnist_lenet_inv}
\end{subfigure}     
\label{subfig:mnist_lenet_inv}
\caption{\textbf{Baseline---Prior Work.} Single sample gradient inversion with untrained network using the inversion method proposed by \cite{Geiping.2020Inverting} for the first 100 MNIST datapoints. (a) and (b) shows fidelity of  individual datapoint reconstruction with no restarts, while (c) and (d) show 32 different optimisation start points. Error bars are a single standard deviation of individual restarts.}
\label{fig:mnist_recon_results}
\end{figure*}

\begin{figure*}[tbh]
\centering
\begin{subfigure}[]{0.24\textwidth}
\centering
\includegraphics[width=\linewidth]{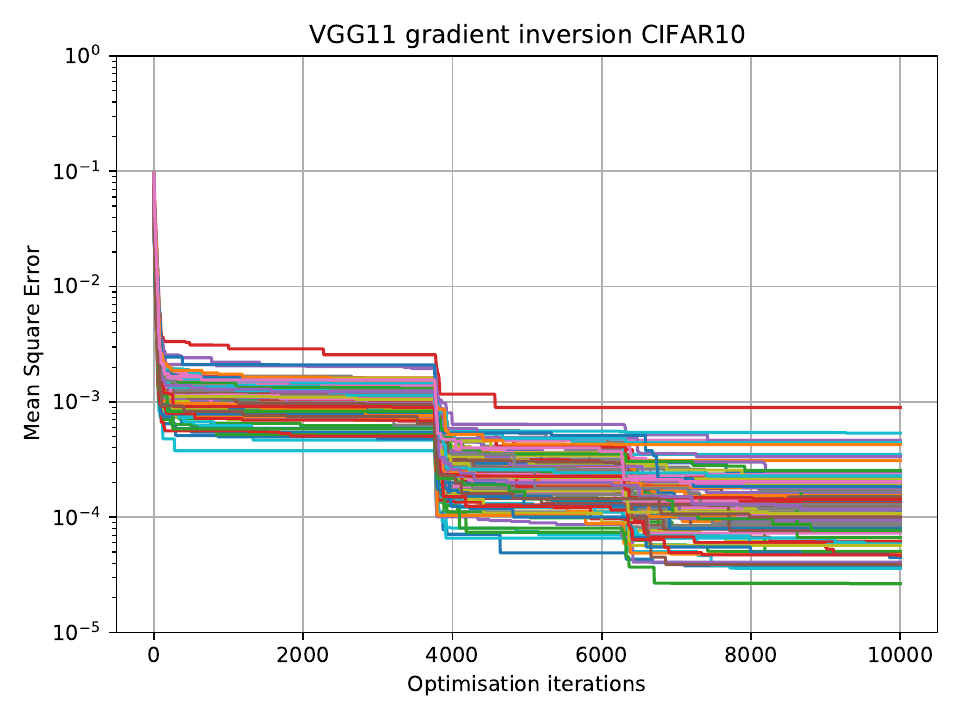}
\caption{VGG11-like}
\label{subfig:cifar10_vgg_inv}
\end{subfigure}  
\begin{subfigure}[]{0.24\textwidth}
\centering
\includegraphics[width=\linewidth]{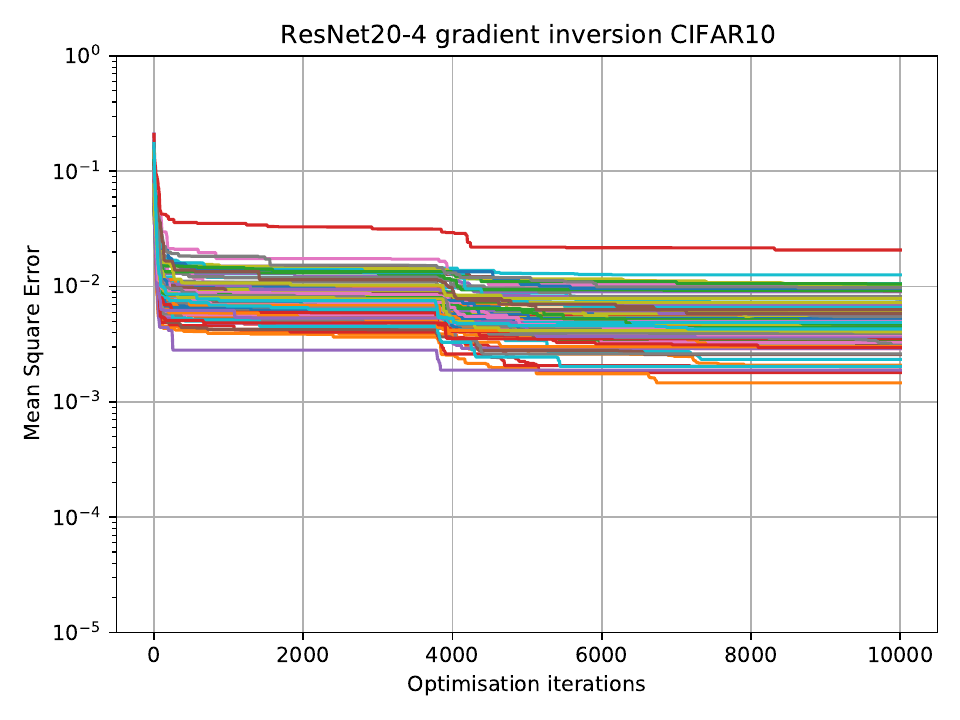}
\caption{ResNet20-4}
\label{subfig:cifar10_resnet_inv}
\end{subfigure}     
\begin{subfigure}[]{0.24\textwidth}
\centering
\includegraphics[width=\linewidth]{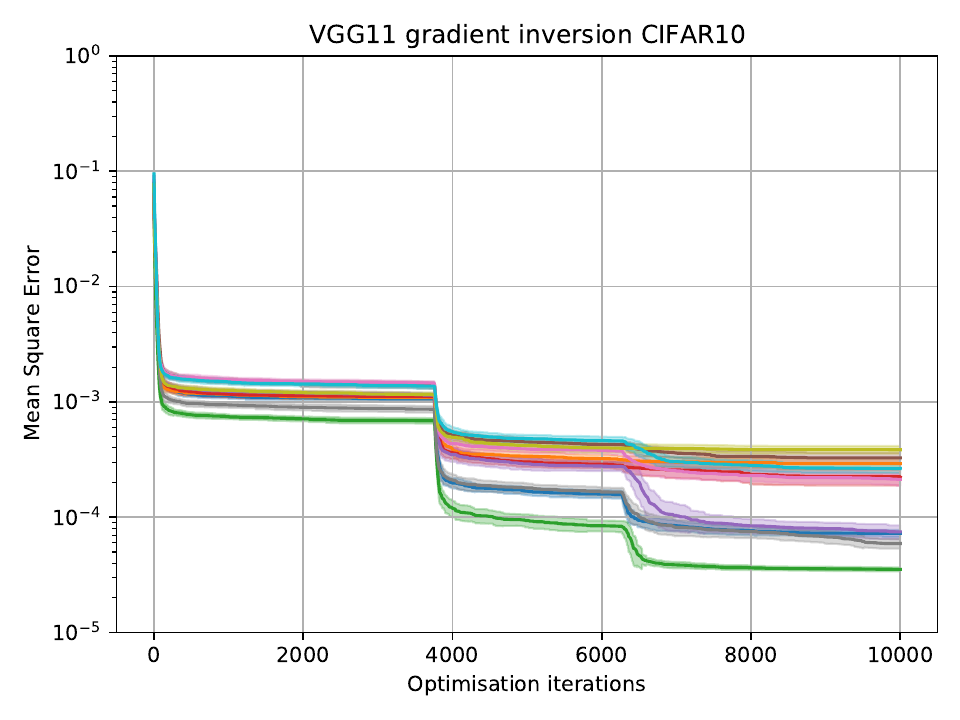}
\caption{VGG11-like}
\label{subfig:restarts_cifar10_vgg11_inv}
\end{subfigure}  
\begin{subfigure}[]{0.24\textwidth}
\centering
\includegraphics[width=\linewidth]{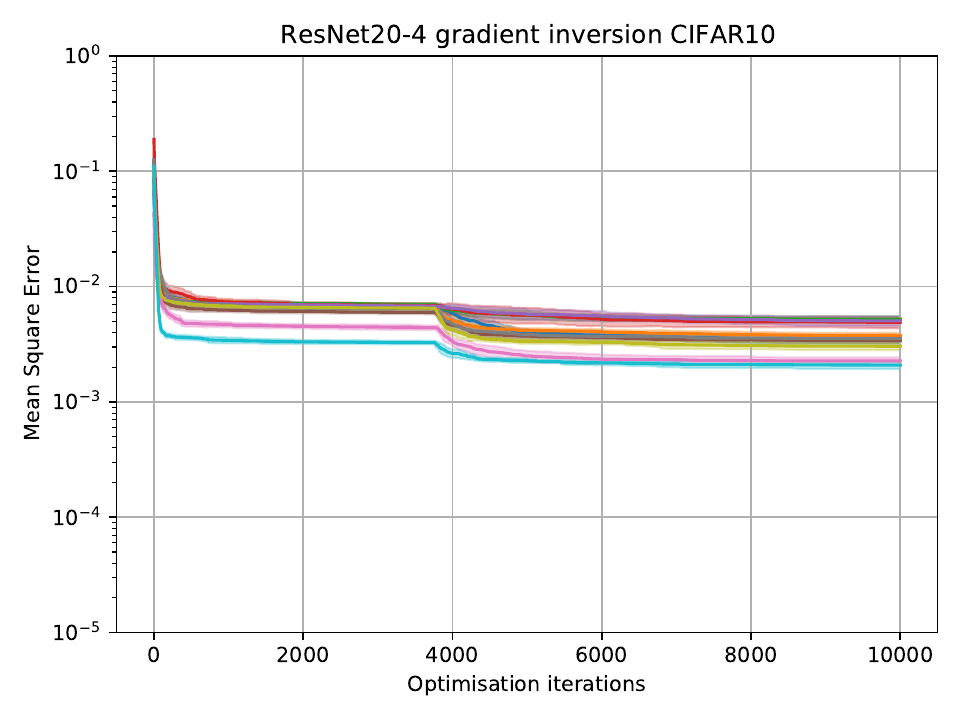}
\caption{ResNet20-4}
\label{subfig:restarts_cifar10_resnet_inv}
\end{subfigure}
\caption{\textbf{Baseline---Prior Work.} Single sample gradient inversion with untrained network using the inversion method proposed by \cite{Geiping.2020Inverting} for the first 100 CIFAR10 datapoints. (a) and (b) shows fidelity of  individual datapoint reconstruction with no restarts, while (c) and (d) show 32 different optimisation start points. Error bars are a single standard deviation of individual restarts.}
\label{fig:cifar10_recon_results}
\end{figure*}

\end{document}